\documentclass[11pt]{article}

\usepackage[final]{acl}

\usepackage{times}
\usepackage{latexsym}
\usepackage{amsmath}
\usepackage{amssymb}
\usepackage{mathtools}
\usepackage{amsthm}
\usepackage{xcolor}         
\usepackage{booktabs}       
\usepackage{tabularx}       
\usepackage{array}          
\usepackage{mathrsfs}
\usepackage{multirow}
\usepackage{booktabs}
\usepackage{tcolorbox}
\usepackage{multicol}
\usepackage{bbding}
\usepackage{pifont}
\usepackage{colortbl}
\usepackage{subcaption}

\usepackage[T1]{fontenc}

\usepackage[utf8]{inputenc}

\usepackage{microtype}

\usepackage{inconsolata}

\usepackage{graphicx}

\title{ICPO: Intrinsic Confidence-Driven Group Relative Preference Optimization for Efficient Reinforcement Learning}

\author{Jinpeng~Wang, Chao~Li, Ting~Ye, Mengyuan~Zhang, Wei~Liu, Jian~Luan \\
  MiLM Plus, Xiaomi Inc. \\
  \texttt{\{wangjinpeng1, lichao75, yeting1, zhangmengyuan7, liuwei40, luanjian\}@xiaomi.com}}

\begin{document}
\maketitle
\begin{abstract}
Reinforcement Learning with Verifiable Rewards (RLVR) demonstrates significant potential in enhancing the reasoning capabilities of Large Language Models (LLMs). However, existing RLVR methods are often constrained by issues such as sparse rewards, reward noise, and inefficient exploration, which lead to unstable training and entropy collapse. To address this challenge, we propose the \textbf{I}ntrinsic \textbf{C}onfidence-Driven Group Relative \textbf{P}reference \textbf{O}ptimization method (ICPO). The intuition behind it lies in the fact that the probabilities of an LLM generating different responses can inherently and directly reflect its self-assessment of the reasoning process. Inspired by the idea of preference modeling, ICPO calculates a preference advantage score for each response by comparing the relative generation probabilities of multiple responses under the same input prompt, and integrates this score with verifiable rewards to guide the exploration process. We have discovered that the preference advantage score not only alleviates the issues of sparse rewards and reward noise but also effectively curbs overconfident errors, enhances the relative superiority of undervalued high-quality responses, and prevents the model from overfitting to specific strategies. Comprehensive experiments across four general-domain benchmarks and three mathematical benchmarks demonstrate that ICPO steadily boosts reasoning compared to GRPO.
\end{abstract}

\section{Introduction}

Large-scale reinforcement learning with verifiable rewards (RLVR) has emerged as a prevailing paradigm for enhancing the reasoning capabilities of LLMs~\citep{Open-Reasoner-Zero-2025, Deepcoder-2025, DeepSeek-R1-2025}. Unlike reinforcement learning from human feedback (RLHF), RLVR eliminates reliance on subjective human judgments or complex learned reward models by directly employing rule-based reward functions to provide explicit feedback signals for model optimization. This paradigm has given rise to a series of efficient and scalable RLVR training algorithms, such as GRPO~\citep{DeepSeek-Coder-2024} and DAPO~\citep{DAPO-2025}. Despite these significant advances, such methods still face several challenging issues.

Firstly, relying solely on final answer to construct binary coarse-grained rewards~\citep{DeepSeek-R1-2025} results in locally sparse reward, which fails to effectively distinguish the quality of behaviors and may leads to zero advantage, thereby impeding the policy from achieving a stable optimization direction. Secondly, most RLVR methods are confined to domains such as mathematical problem-solving~\citep{Zichen-Liu-2025, SimpleRL-Zoo-2025} and code generation~\citep{Deepcoder-2025, Skywork-2025, Ganqu-CuiLifan-Yuan-2025}, rendering them incapable of designing rule-based verifiers for general-domain reasoning with free-form answers. Recent studies have attempted to address this issue by employing LLMs as verifiers~\citep{General-Reasoner-2025}. However, the unreliability and high variance of reward signals (reward noise) further undermine training stability. Finally, the model's policy distribution may rapidly skew toward a few high-reward output patterns due to the lack of fine-grained feedback, leading the policy to collapse into a repetitive action selection state—a phenomenon known as entropy collapse.

To address these challenges, we propose the \textbf{Intrinsic Confidence-Driven Group Relative Preference Optimization (ICPO)}, which leverages the model's inherent self-assessment capability to compensate for the deficiencies of verifiable rewards. The key intuition lies in the fact that the probability distribution generated by LLMs when producing different reasoning responses essentially represents an implicit self-assessment of the model's confidence in its own reasoning. A higher generation probability means the model deems the reasoning path as highly correct and possesses absolute confidence; however, this often corresponds to the model's path dependency on familiar patterns, potentially leading to inertial outputs for simple scenarios. Conversely, a lower generation probability may actually come from the model's attempts to reason through complex or rare samples, even when it lacks confidence. This probabilistic preference constitutes a fine-grained signal that can also reflect the effectiveness of policy optimization. Therefore, we can leverage this \textbf{intrinsic preference as an auxiliary signal} to guide policy learning.

Specifically, ICPO draws on the modeling approach of pairwise preferences in direct preference optimization algorithms. For multiple in-group responses generated in response to the same input prompt, it first sorts them in ascending order based on their generation probabilities. Subsequently, it forms pairwise response sets by combining the responses two by two within the group. Then, by comparing the probabilities of these pairwise responses, it calculates a \textbf{preference advantage score} for each response that reflects its relative superiority or inferiority within the group. During the optimization process, ICPO deeply \emph{integrates the preference advantage score with traditional verifiable external rewards}. This design effectively addresses two typical failure modes in reasoning-based reinforcement learning: (1) It can \emph{precisely identify and suppress overconfident errors} in the model—that is, the model generates seemingly plausible yet actually erroneous reasoning paths with high probability, while simultaneously \emph{enhancing the relative advantage of undervalued high-quality responses}; (2) By \emph{continuously providing fine-grained comparative signals regarding relative merits within groups}, it sustains the exploratory drive of the policy, thereby avoiding training instability and entropy collapse caused by ambiguous or extremely sparse external rewards.

Our core contributions are threefold:
\begin{itemize}
\item We innovatively propose a preference advantage score calculation mechanism that transforms the intrinsic probabilities of in-group responses into relative preference signals. This breakthrough overcomes the limitations of traditional verifiable rewards' singularity, providing stable guidance in scenarios where rewards are sparse or noisy. Additionally, it can suppress overconfident errors while enhancing the relative advantage of undervalued high-quality responses, thereby laying a reliable foundation for policy updates.
\item We propose the ICPO method, a simple yet highly extensible solution that deeply integrates preference advantage scores with external rewards through multi-stage weight adjustment, effectively balancing the complementary values of intrinsic self-assessment and extrinsic objective verification.
\item Comprehensive experiments on seven benchmarks demonstrate that ICPO consistently outperforms baselines (e.g., GRPO) across various model architectures like Qwen, Llama, and Gemma, demonstrating superior effectiveness and cross-model generalizability.
\end{itemize}

\section{Related Works}
\subsection{RLVR}

Reinforcement Learning with Verifiable Rewards (RLVR), as a robust alternative to Reinforcement Learning from Human Feedback (RLHF), has been demonstrated to effectively enhance model reasoning capabilities~\citep{Ganqu-CuiLifan-Yuan-2025, TÜLU3-2025, Deepscaler-2025}. While RLHF~\citep{Long-Ouyang-2022, Deborah-Cohen-2022, Leo-Gao-2023} leverages human preference data to train reward models—yielding notable improvements—it incurs substantial resource costs due to heavy reliance on manual annotation~\citep{Llama2-2023}. In contrast, RLVR employs rule-based verification functions (e.g., exact answer matching~\citep{Mimo-2025, OpenAI-o1-2024}) to provide reward signals, circumventing the complexity and potential pitfalls of learned reward models. However, this approach remains limited to domains with precise verifiers and suffers from sparse rewards. Recent studies have explored LLM-based verifiers (LLM-as-a-Judge)~\citep{Dawei-Li-2024, PandaLM-2024, JudgeLM-2025} to extend RLVR to open-ended question-answering scenarios. Yet, due to uncertainty in LLM outputs and hallucination issues, the reward signals exhibit low reliability, potentially misleading policy optimization.

\begin{figure*}[t]
  \includegraphics[width=\linewidth]{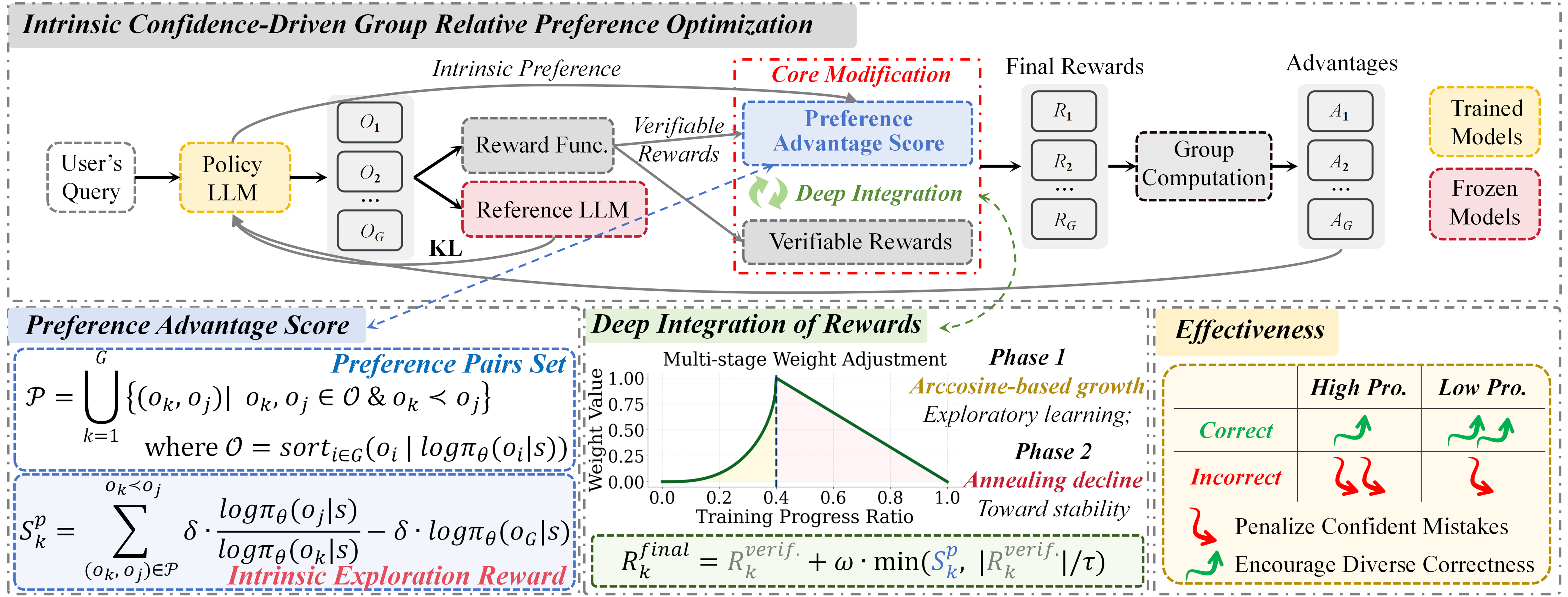}
  \caption {Illustration of ICPO. ICPO computes preference advantage scores for responses within a sampled group, integrates them with verifiable rewards, and stabilizes training via multi-stage weight adjustment.}
  \label{fig1}
  \vspace{-0.5em}
\end{figure*}

\subsection{Self-Reward Optimization}
Self-reward optimization via intrinsic signals has emerged as an effective approach to mitigate reliance on extrinsic rewards derived from manual annotation or specialized verification tools. These methods directly extract optimization guidance from the generative process or output features of policy models~\cite{Weizhe-Yuan-2024, TTRL-2025, Xuandong-Zhao-2025}, inspired by observations that LLMs exhibit lower confidence when handling complex problems—optimizing such confidence effectively enhances reasoning capabilities~\cite{Farquhar-Sebastian-2024, Semantic-Uncertainty-2023, Katie-Kang-2024, Zhewei-Kang-2025}. Early studies, such as SPIN~\cite{Zixiang-Chen-2024} and self-rewarding LLMs~\cite{Weizhe-Yuan-2024}, leveraged model-generated feedback for subsequent training iterations, while INTUITOR~\cite{Xuandong-Zhao-2025} employed self-confidence as an intrinsic confidence-based reward. However, such methods risk limiting exploration~\cite{Ganqu-Cui-2025, Andreas-Hochlehnert-2025}, as excessive reliance on model’s own epistemic state may trap the policy in self-reinforcing local optima. PLPR~\cite{RLPR-2025} introduced a stable probability-to-reward conversion method. CDE~\cite{CDE-2025} employs the Perplexity of responses as an auxiliary signal to guide policy optimization. However, relying on the absolute perplexity of individual responses rather than their relative quality within a group may lead to suboptimal performance.

\section{ICPO}
In this section, we first introduce the fundamental principles of ICPO. Next, we describe the method for calculating preference advantage scores for each response, and the multi-stage weight adjustment to ensure the stability of training process.

\subsection{RL with Intrinsic Confidence-Driven}
The core of RLVR lies in replacing traditional RLHF reward models with rule-based reward functions (or general LLM scoring). thereby eliminating the need to construct large-scale human preference datasets. Specifically, RLVR optimizes the policy by maximizing the following objective:
\begin{equation}\label{eq:1}
  \max_{\pi_\theta} \mathbb{E}_{o \sim \pi_\theta(q)} \left[ v(q, o) - \beta \mathrm{KL}[\pi_\theta(o|q) \| \pi_{\text{ref}}(o|q)] \right]
\end{equation}
where $q$ denotes the input query, $o$ represents the generated output, $\pi_{\text{ref}}$ is the initial reference policy, and $\beta$ controls the KL divergence coefficient to prevent excessive deviation from $\pi_{\text{ref}}$, $v(q, o)$ constitutes a verifiable reward function. Common RLVR algorithms include REINFORCE~\cite{Ronald-1992}, PPO~\cite{PPO-2017}, and GRPO~\cite{GRPO-2024}. GRPO significantly reduces GPU memory consumption and training computational costs by not needing the value model. Its efficacy has been validated in DeepSeekMath~\cite{GRPO-2024} and DeepSeek-R1~\cite{DeepSeek-R1-2025}, establishing its prominent position in reinforcement learning. Specifically, for each problem $q$, GRPO samples a set of outputs $\{o_1, o_2, …, o_G\}$ from the old policy $\pi_{\theta_{old}}$, then optimizes the policy model by maximizing the following objective:
\begin{equation}
\begin{split}
&\mathcal{J}(\theta)= \mathbb{E}_{q\sim D, \{o_k\}^G_{k=1} \sim \pi_{\theta_{old}}(O|q)} [\frac{1}{G} \sum^{G}_{k=1} \frac{1}{|o_k|}\sum^{|o_k|}_{t=1} \\
&\quad \min\left(r^t_kA^t_k, \mathrm{clip}(r^t_k, 1-\epsilon, 1+\epsilon)A^t_k\right) - \mathbb{KL}]
\end{split}
\end{equation}
\begin{equation}
r^t_k=\frac{\pi_\theta(o^t_k|q, o_k^{<t})}{\pi_{\theta_{old}}(o^t_k|q, o_k^{<t})}, \enspace  A^t_k=\frac{R_k - \mathrm{mean}(\{R_k\})}{\mathrm{std}(\{R_k\})}
\end{equation}
where $\epsilon$ and $\beta$ are hyperparameters, and $A^t_i$ represents the relative advantage estimation based on within-group rewards. However, GRPO suffers from training instability due to sparse rewards or reward noise, and may encounter entropy collapse, leading to suboptimal performance.

To address the above challenges, we propose Intrinsic Confidence-Driven Group Relative Preference Optimization (ICPO), as illustrated in Figure~\ref{fig1}. This method integrates external rewards with the model's self-evaluation of its reasoning processes. ICPO offers three key benefits: (1) it provides finer optimization guidance under sparse reward conditions; (2) it mitigates random noise in verifiable rewards and accentuates relative value differences between high- and low-quality responses; (3) it reduces prevalent overconfidence errors in reasoning while enhancing the relative superiority of undervalued high-quality responses. The optimization objective of ICPO is:
\vspace{-0.5em}
\begin{equation}
\begin{split}
&\mathcal{J}(\theta)= \mathbb{E}_{q\sim D, \{o_k\}^G_{k=1} \sim \pi_{\theta_{old}}(O|q)} [\frac{1}{G} \sum^{G}_{k=1} \frac{1}{|o_k|}\sum^{|o_k|}_{t=1} \\
&\quad \min\left(r^t_k\tilde{A}^t_k, \mathrm{clip}(r^t_k, 1-\epsilon, 1+\epsilon)\tilde{A}^t_k\right) - \mathbb{KL}]
\end{split}
\end{equation}
\begin{equation}
\mathbb{KL} = \beta \mathrm{KL}(\pi_\theta||\pi_{\theta_{ref}}), \enspace  \tilde{A}^t_k=\frac{\tilde{R}_k - \mathrm{mean}(\{\tilde{R}_k\})}{\mathrm{std}(\{\tilde{R}_k\})}
\end{equation}
where $\tilde{R}_k$ denotes the verifiable reward incorporating normalized preference advantage scores, which is the 
$R^{final}_k$ depicted in Figure~\ref{fig1}.


\subsection{Preference Advantage Scores}
Preference advantage scores essentially reflect the model's self-assessment of the relative merits of different responses within a group. This intuition stems from the observation that LLMs often exhibit lower probabilities when encountering unfamiliar tasks or lacking sufficient knowledge reserves ~\cite{Katie-Kang-2024}. By incorporating self-confidence as an additional reward, ICPO guides the policy to proactively learn from undervalued responses, thereby uncovering underutilized knowledge. Specifically, we first sort responses within a sampled group in ascending order based on their intrinsic preferences (i.e., responses with lower intrinsic probabilities are assigned earlier ranking positions). The sorting process is represented as:
\begin{equation}
\mathcal{O} = \text{sort}_{i \in G} \left( o_i \mid \left(log \pi_{\theta}(o_i|s) \right) \right)
\label{eq6}
\end{equation}
\begin{equation}
\pi_\theta(o_i|s) = \frac{1}{L_i} \sum_{t=1}^{L_i} \pi_\theta(o_t \mid o_{<t}, s)
\end{equation}
where $L_i$ represents the effective length of $o_i$ (i.e., the number of non-padding tokens), $\pi_\theta(o_t \mid o_{<t}, s)$ denotes the probability of model generating token $o_t$ at position $t$, and $\pi_\theta(o_i|s)$ denotes the sequence-level probability of response $o_i$. We normalize probabilities based on effective length to ensure fair comparison across responses and prevent bias toward shorter responses. This ranking structure, which is solely dominated by generation probabilities, is decoupled from external rewards, thereby avoiding noise interference from external rewards.

Based on the sorting results, we construct all valid preference pairs $(o_i, o_j)$ satisfying the partial order relation where $o_i$ is strictly ranked ahead of $o_j$. The preference pair set is defined as:
\begin{equation}
\mathcal{P} = \bigcup_{k=1}^{G} \left\{ (o_k, o_j) \mid o_k, o_j \in \mathcal{O} \ \& \ o_k \prec o_j \right\}
\label{eq8}
\end{equation}

Subsequently, we calculate a preference advantage score for each response. By effectively modeling the partial order relations, we quantify the relative superiority or inferiority of each response compared to other candidate responses within the same group. Formally, the preference advantage score for the $k$-th response is defined as:
\begin{equation}
S_k^p = \sum_{\substack{(o_k, o_j) \in \mathcal{P}}}^{o_k \prec o_j} \delta \cdot \frac{log\pi_\theta(o_j|s)} {log\pi_\theta(o_k|s)} - \delta \cdot log\pi_\theta(o_G|s)
\label{eq9}
\end{equation}
where $\delta$ represents the temperature scaling factor. The first term in the formula accumulates the potential learning value of the current response for subsequent responses, while the second term ensures that high-confidence responses are not entirely overlooked. The advantage score $S_k^p$ computed via this method emphasizes guiding the model to learn high-quality responses with low confidence, steering clear of the safe response pattern that merely pursues high probabilities but lacks substantive content.  Moreover, it effectively avoids excessive encouragement of extreme low-probability responses, which are of no practical learning value (as the first term in formula of these extremely low-probability responses is relatively small). Consequently, it facilitates the creation of fine-grained intrinsic reward signals, enabling precise capture and learning of high-quality responses underestimated by models.

\begin{figure}[t]
  \begin{minipage}{0.54\linewidth}
    \includegraphics[width=\textwidth]{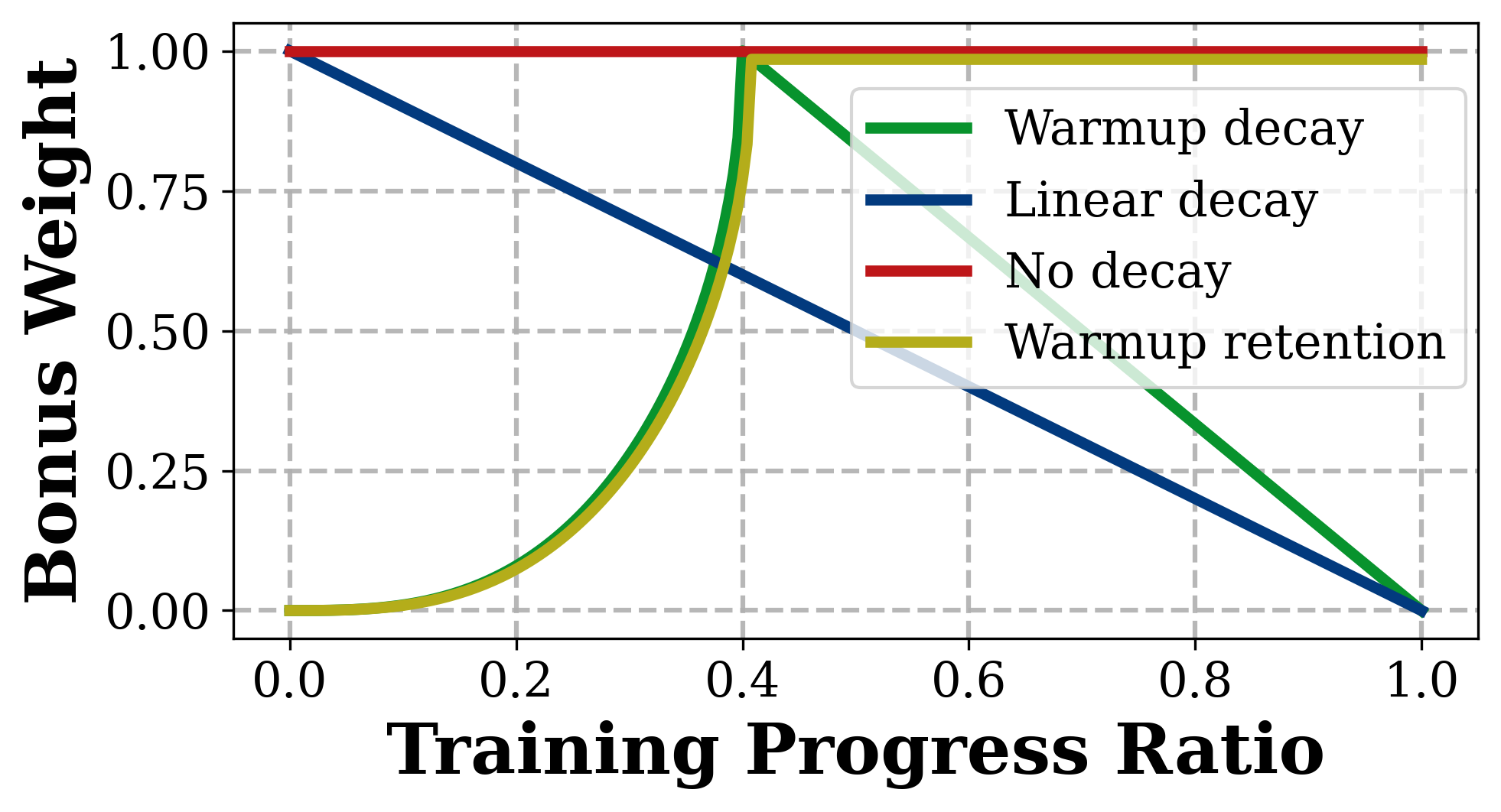}
    \label{fig2a}
  \end{minipage}
  \begin{minipage}{0.45\linewidth}
    \includegraphics[width=\textwidth]{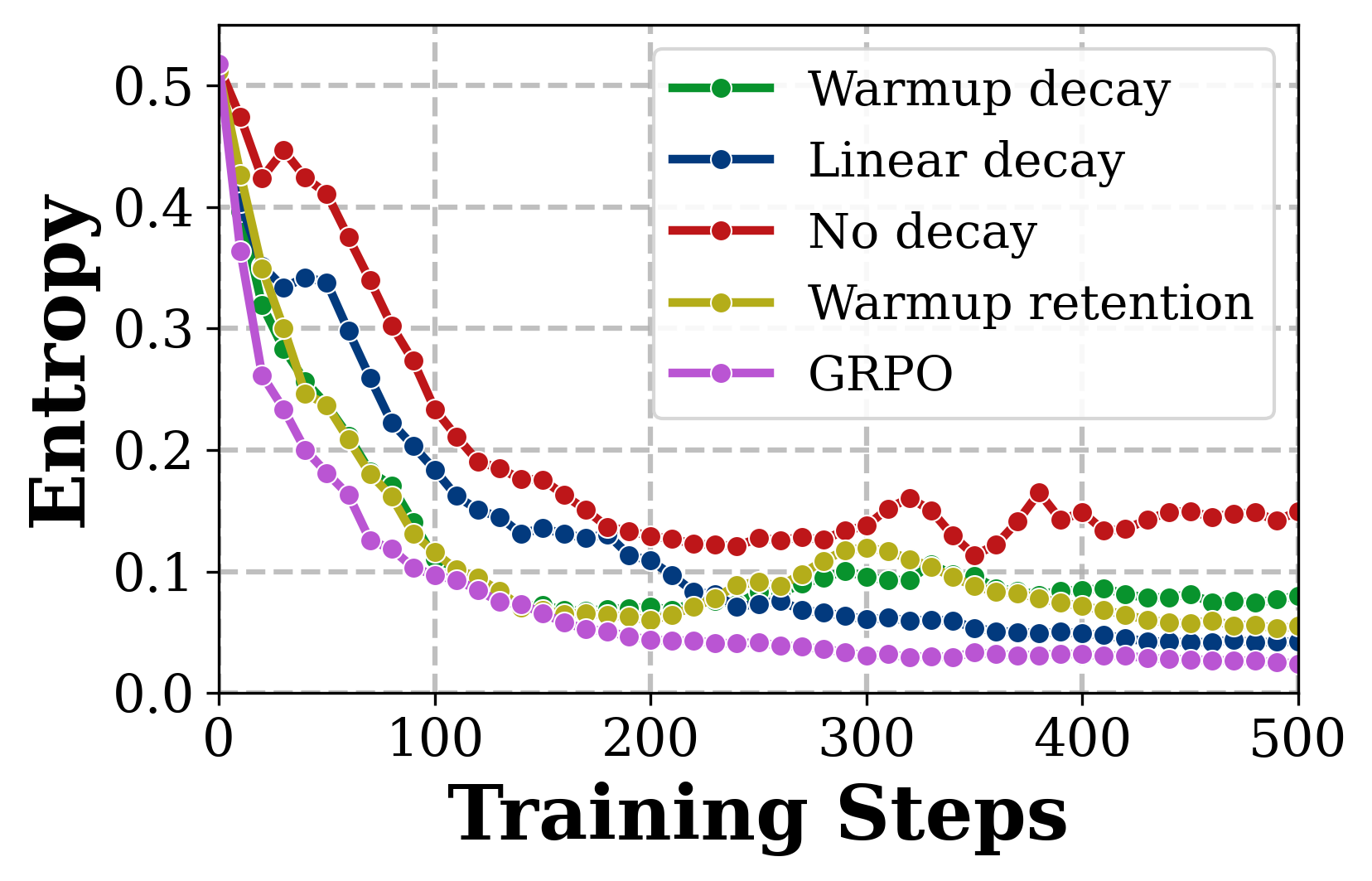}
    \label{fig2b}
  \end{minipage}
  \vspace{-1.5em}
  \caption{An illustration of different weight adjustment.}
  \label{fig2}
  \vspace{-1.0em}
\end{figure}

\subsection{Reward Design}
To more effectively guide the policy model in exploring and learning underutilized knowledge, we have devised a composite reward function that integrates verifiable rewards derived from external scoring strategies with advantage signals based on intrinsic generation preferences (i.e., preference advantage scores). This reward mechanism not only reflects the absolute performance of responses in the target task but also captures their relative learning value among candidate responses within the same group. The final reward is modeled as:
\begin{equation}
R^{verif.}_k = 0.9 \cdot R^{answer}_k+0.1 \cdot R^{format}_k
\end{equation}
\begin{equation}
R^{final}_k=R^{verif.}_k+ \omega \cdot \operatorname{min}\left(
S_k^p,\ {|R^{verif.}_k|} / {\tau} \right)
\label{eq11}
\end{equation}
where the weighting parameter $\omega$ is utilized to regulate the strength of intrinsic advantage signal injection. We adopt a multi-stage weight adjustment strategy to dynamically modify $\omega$ across different training stages, with specific settings detailed in Section~\ref{sec3.4}. Additionally, $\tilde{S}_k^p$ is clipped based on the verifiable reward of each individual response, thereby preventing intrinsic preferences from excessively dominating the composite reward. This reward mechanism ensures that encouragement is effectively applied to correct, low-probability responses that possess learning value (while incorrect low-probability responses receive minimal intrinsic rewards due to the truncation mechanism), and it also suppresses the model's overconfident errors.

\subsection{Multi-stage Weight Adjustment} \label{sec3.4}
To effectively balance the internal and external reward signals, we have designed a multi-stage weight adjustment strategy to dynamically modulate the value of $\omega$, as illustrated by the curve in Figure~\ref{fig1}. Specifically, during the initial training phase, we employ an inverse cosine growth pattern to alter the weight value, enabling the model to gradually assimilate fine-grained preference information driven by its own probability distribution. Once $\omega$ reaches its maximum value, we adopt a linear annealing to reduce the weight value to its minimum, thereby preventing the training process from being excessively guided by intrinsic rewards.

This design originates from the following insight: In early training, high entropy makes it challenging to determine whether low-probability responses are worth learning. As training continues and entropy drops, correct and novel low-probability responses emerge, at which point we can incrementally enhance the learning motivation for these responses. In later training, when most novel samples are learned, remaining low-probability responses may be noise or valueless signals; so reducing intrinsic signals' influence stabilizes policy learning. Figure~\ref{fig2} (left) displays various weight adjustment strategies, with a detailed analysis of their effects in the experimental section. The warmup turning point is set at 0.4 because, in GRPO training, the 20\%-40\% phase shows suitable entropy reduction with active exploration, making it ideal for introducing strong intrinsic preference rewards. An ablation study of this setting is in Appendix~\ref{appendixc3}.

\section{Experiments}
This section empirically validates the effectiveness of ICPO and explores its potential applications in both verifiable reward and noisy reward domains.

\begin{table*}[t]
\setlength{\tabcolsep}{4pt}        
\renewcommand{\arraystretch}{0.95}  
\begin{center}
\begin{small}
\begin{tabular}{m{0.95in}m{0.35in}<{\centering}|m{0.55in}<{\centering}m{0.35in}<{\centering}m{0.55in}<{\centering}m{0.41in}<{\centering}m{0.5in}<{\centering}m{0.38in}<{\centering}m{0.45in}<{\centering}|m{0.38in}<{\centering} m{0.25in}<{\centering}}
\toprule
\midrule
\multirow{2}{0.9in}{\textbf{Model}} & \multirow{2}{0.4in}{\textbf{Verifier}} & \textbf{MMLUPro} & \textbf{GPQA} & \textbf{TheoremQA} & \textbf{WebInst.} & \textbf{MATH500} & \textbf{Minerva} & \textbf{AIME24} & \textbf{General} & \textbf{Math}\\
& & Avg@2 & Avg@4 & Avg@2 & Avg@2 & Avg@2 & Avg@2 & Avg@16 & - & -\\
\midrule
\multicolumn{11}{c}{\textbf{Gemma Models}} \\
\midrule
Gemma2-2B-It & -- & 27.9 & 19.3 & 16.4 & 33.5 & 26.6 & 15.9 & 0.0 & 24.3 & 14.2\\
RLVR(GRPO)$^\dagger$ & Rule & 31.6 & 25.8 & 20.1 & 52.3 & \textbf{30.7} & 16.5 & \textbf{0.2} & 32.5 & 15.8\\
\rowcolor{cyan!20} \textbf{ICPO (Ours)} & Rule & \textbf{33.0} & \textbf{28.9} & \textbf{20.4} & \textbf{53.8} & 30.6 & \textbf{17.0} & \textbf{0.2} & \textbf{34.0} & \textbf{15.9}\\
\midrule
\multicolumn{11}{c}{\textbf{Llama Models}} \\
\midrule
Llama3.1-8B-Inst & -- & 46.4 & 31.6 & 31.3 & 54.7 & 50.1 & 32.7 & 4.2 & 40.5 & 29.0\\
RLVR(GRPO)$^\dagger$ & Rule & 50.3 & 36.0 & 33.0 & 63.2 & 51.9 & 35.2 & 6.5 & 45.6 & 31.2\\
\rowcolor{cyan!20} \textbf{ICPO (Ours)} & Rule & \textbf{52.7} & \textbf{37.2} & \textbf{34.7} & \textbf{68.9} & \textbf{53.8} & \textbf{39.0} & \textbf{8.8} & \textbf{48.4} & \textbf{33.9}\\
\midrule
\multicolumn{11}{c}{\textbf{Qwen Models}} \\
\midrule
Qwen2.5-7B & -- & 45.3 & 32.4 & 41.4 & 60.4 & 51.0 & 37.6 & 6.5 & 44.9 & 31.7\\
Qwen2.5-7B-Inst & -- & 54.5 & 34.2 & 47.3 & 72.6 & 75.4 & 49.4 & 9.4 & 52.2 & 44.7\\
\midrule
Orat-Zero(M) & Rule & 45.8 & 38.8 & 53.3 & 71.5 & \underline{80.8} & 52.1 & \textbf{29.8} & 52.4 & \textbf{54.2}\\
PRIME(M) & Rule & 39.5 & 32.1 & 47.7 & 54.5 & 76.4 & 45.5 & 20.4 & 43.4 & 47.4\\
SimpleRL-Zoo(M) & Rule & 46.9 & 38.4 & 51.1 & 70.3 & 77.1 & 51.0 & 26.5 & 51.7 & 51.5\\
\midrule
TTRL & Rule & 51.1 & 34.1 & 48.8 & 68.0 & \textbf{82.1} & 52.8 & 15.8 & 50.5 & 50.2\\
SimpleRL-Zoo & Rule & 54.1 & 36.2 & 49.5 & 70.7 & 76.3 & 49.2 & 14.8 & 52.6 & 46.8\\
General Reasoner & Model & 55.4 & 37.4 & 52.1 & 74.5 & 77.0 & 51.7 & 16.0 & 54.8 & 48.2\\
RLPR & \color{green}{\ding{55}} & 56.0 & 37.6 & \textbf{55.4} & 75.5 & 78.0 & \textbf{56.5} & 16.3 & \underline{56.1} & 50.3\\
INTUITOR$^\dagger$ & \color{green}{\ding{55}} & 54.9 & 37.3 & 53.0 & \underline{76.1} & 76.5 & 53.6 & 16.0 & 55.3 & 48.7\\
CDE$^\dagger$ & Rule & \underline{57.4} & \textbf{40.1} & \underline{54.0} & 72.3 & 76.0 & 53.7 & \underline{23.3} & 55.9 & 51.0\\
RLVR(GRPO)$^\dagger$ & Rule & 55.1 & 36.2 & 52.2 & 75.3 & 76.5 & 54.9 & 17.7 & 54.7 & 49.7 \\
\rowcolor{cyan!20} \textbf{ICPO (Ours)} & Rule &  \textbf{57.6} & \underline{39.4} & \underline{54.0} & \textbf{77.8} & 76.2 & \underline{56.2} & \underline{23.3} & \textbf{57.2} & \underline{51.9} \\
\bottomrule
\end{tabular}
\caption{Overall performance on seven reasoning benchmarks. General: Average of MMLU-Pro, GPQA, TheoremQA and WebInst. Math: Average of MATH-500, Minerva and AIME24. The best and second results are marked in \textbf{bold} and \underline{underlined}, respectively. {\color{green}{\ding{55}}}: Method does not require verifiers. (M): Method is trained based on Qwen2.5-Math-7B. $^\dagger$: Official code/methods-based reproduced results.}
\label{tab1}
\vspace{-1em}
\end{small}
\end{center}
\end{table*}

\subsection{Baselines and Setup}
We conduct experiments on Gemma2~\cite{Gemma2-2024}, Llama3.1 ~\cite{llama3-2024} and Qwen2.5~\cite{Qwen2.5-2024} series models for fair comparison with existing methods. Unless otherwise specified, experiments are conducted on Qwen2.5-7B-Base. A comprehensive performance evaluation of ICPO method was carried out using three mathematical reasoning benchmarks, namely MATH-500~\cite{MATH-500-2024}, Minerva~\cite{Minerva-2022}, and AIME24, as well as four general-domain benchmarks, including MMLU-Pro~\cite{Mmlu-pro-2024}, GPQA~\cite{GPQA-2023}, TheoremQA~\cite{TheoremQA-2023}, and WebInstruct. The baseline models include: (1) Base and Instruct models; (2) PRIME~\cite{Ganqu-CuiLifan-Yuan-2025}; (3) SimpleRL-Zoo~\cite{SimpleRL-Zoo-2025}; (4) Oat-Zero~\cite{Zichen-Liu-2025}; (5) TTRL~\cite{TTRL-2025}; (6) General Reasoner~\cite{General-Reasoner-2025}; (7) RLPR~\cite{RLPR-2025}; (8) INTUITOR~\cite{Xuandong-Zhao-2025}; and (9) CDE~\cite{CDE-2025}. Detailed descriptions of datasets and experimental setups are elaborated upon in Appendix~\ref{sec:appendix1} and~\ref{sec:appendix2}.

\subsection{Main Results}

\begin{figure}[t]
  \begin{minipage}{0.49\linewidth}
    \includegraphics[width=\textwidth]{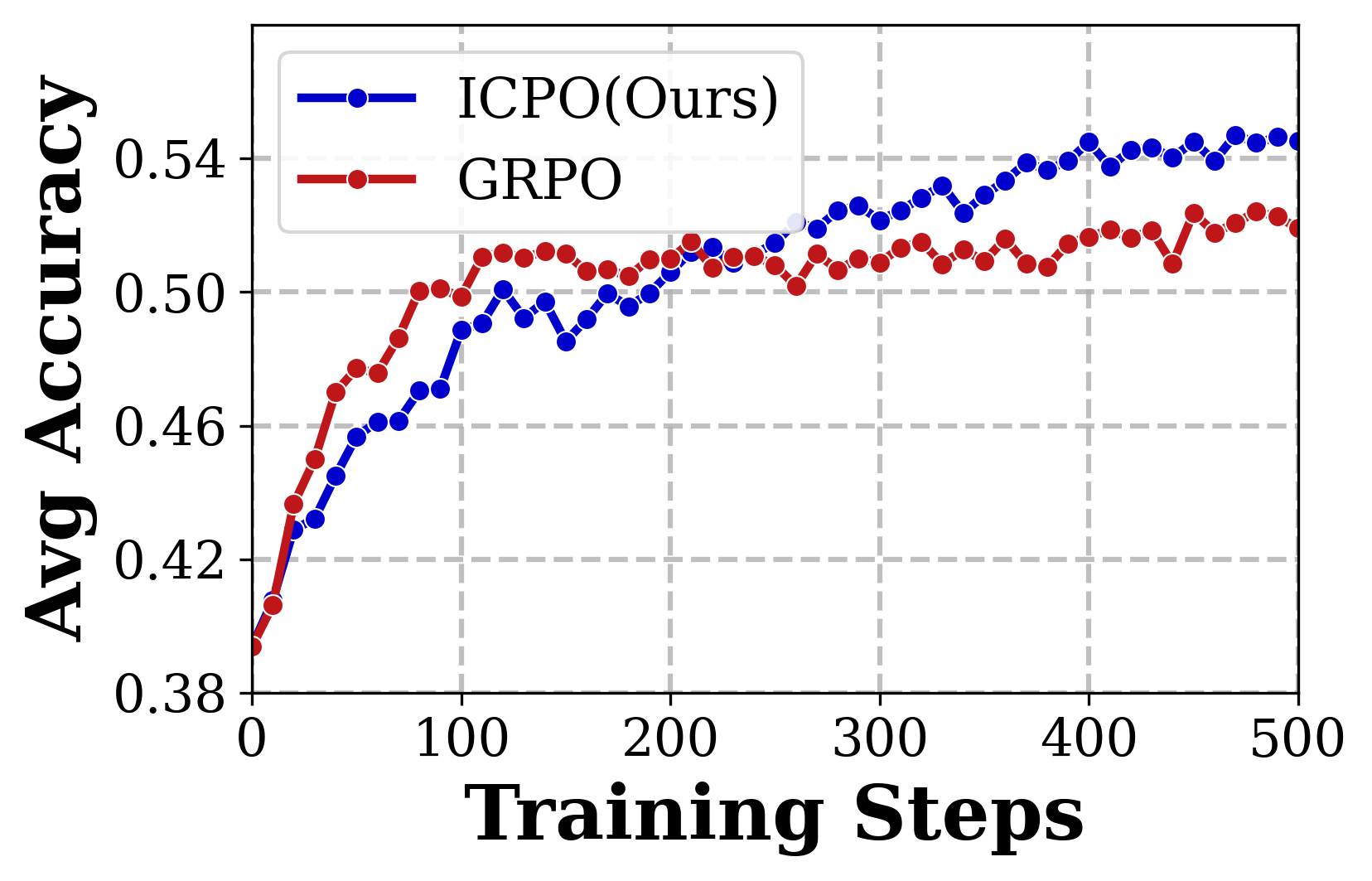}
    \label{fig3a}
  \end{minipage}
  \begin{minipage}{0.49\linewidth}
    \includegraphics[width=\textwidth]{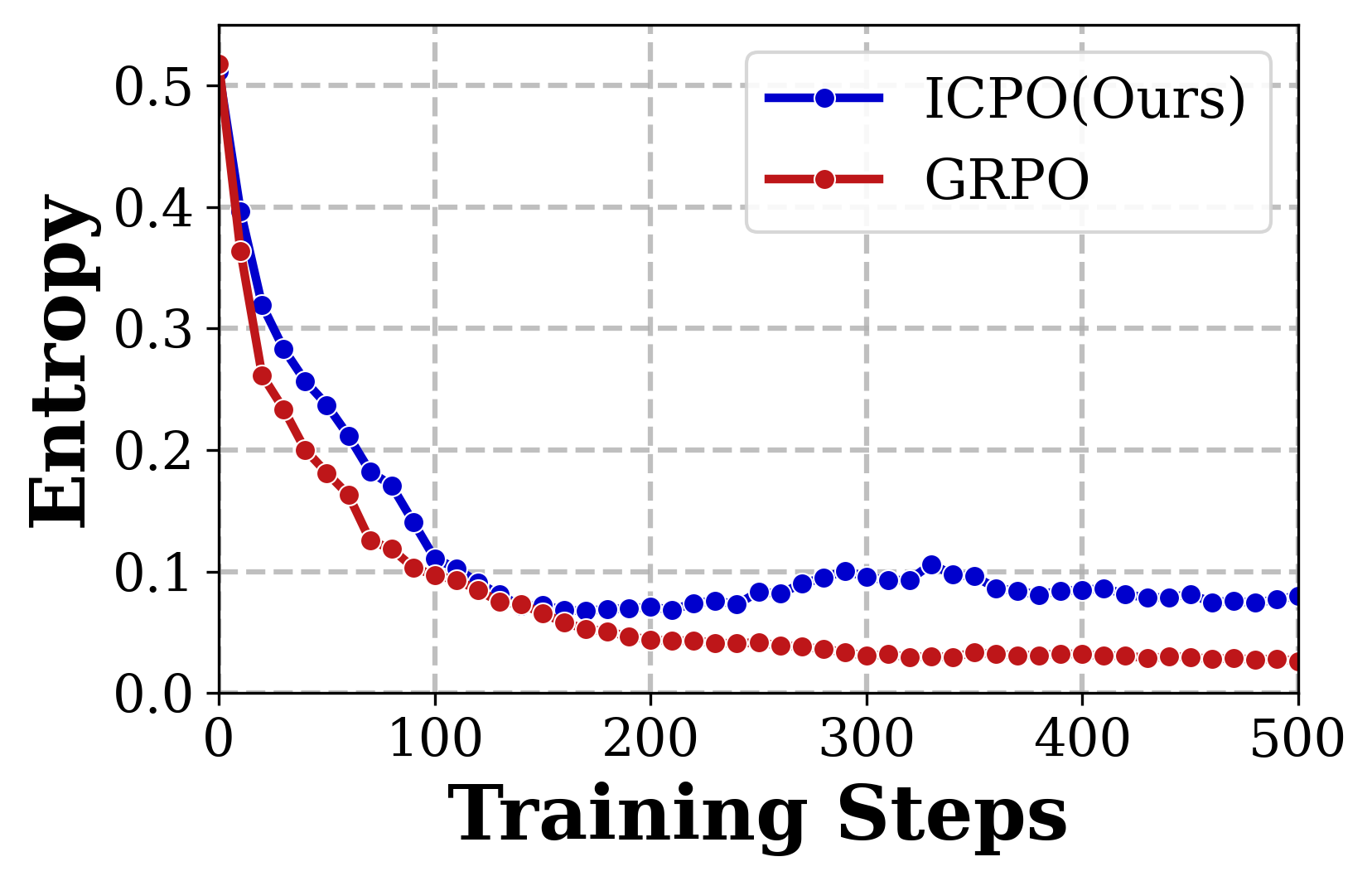}
    \label{fig3b}
  \end{minipage}
  \vspace{-1.5em}
  \caption{Comparison of Avg accuracy on seven benchmarks and entropy of GRPO (Baseline) and ICPO (Our).}
  \label{fig3}
  \vspace{-1.0em}
\end{figure}

Main results and training dynamic are presented in Table~\ref{tab1} and Figure~\ref{fig3}. Key findings are as follows:

(1) Intrinsic confidence-driven mechanism has substantially enhanced reasoning performance in general domains. Compared to RLVR trained using the vanilla GRPO, we observed greater improvements in general reasoning performance across the Gemma, Llama, and Qwen models, with average increases of 1.5, 2.8, and 2.5 points, respectively. Furthermore, ICPO also demonstrated a significant advantage over other baseline methods.

(2) ICPO demonstrates exceptional performance in mathematical reasoning scenarios, surpassing both PRIME and SimpleRL-Zoo across three mathematical benchmarks. Notably, even without training on mathematical models, ICPO still demonstrates a significant enhancement in mathematical reasoning capability, comparable to methods specifically trained on mathematical models.

(3) During the training process, ICPO exhibits a slower rate of improvement in test set accuracy compared to the vanilla GRPO in the early training stages. However, it gradually catches up and ultimately surpasses the performance of GRPO. Moreover, the policy entropy is maintained at a stable exploratory state throughout the entire training process. This aligns with intrinsic confidence-driven exploration: ICPO prevents premature convergence to false high-reward paths, encouraging thorough exploration of high-quality actions in low-confidence regions. This helps the model shift smoothly toward goal-directed exploration, reducing entropy collapse and boosting accuracy.

\begin{table}
    \centering
	\renewcommand\arraystretch{0.95}
    \footnotesize
	\begin{tabular}{m{0.43in}m{0.53in}<{\centering}m{0.38in}<{\centering}m{0.51in}<{\centering}m{0.38in}<{\centering}}
		\toprule
		\midrule
		\multirow{2}{0.43in}{\textbf{Methods}} & {\cellcolor{gray!30} \textbf{MMLUPro}} & {\cellcolor{gray!30}\textbf{GPQA}}  & {\cellcolor{gray!30}\textbf{MATH500}} & {\cellcolor{gray!30}\textbf{Minerve}} \\ 
            \cmidrule{2-5}
			& {\cellcolor{gray!30} Avg@2} & {\cellcolor{gray!30} Avg@4} & {\cellcolor{gray!30} Avg@2} & {\cellcolor{gray!30} Avg@16} \\
		\midrule
            Baseline & 45.3 & 32.4 & 51.0 & 37.6  \\
            + GRPO & 46.9 & 31.8 & 56.2 & 38.9  \\
            \rowcolor{cyan!20} \textbf{+ ICPO} & \textbf{52.3} & \textbf{35.0} & \textbf{68.5} & \textbf{45.5} \\
		\bottomrule
	\end{tabular}
    \caption{Performance on sparse reward settings.}
    \label{tab2}
    \vspace{-1.0em}
\end{table}

\begin{table*}[t]
\setlength{\tabcolsep}{4pt}        
\renewcommand{\arraystretch}{0.95}  
\begin{center}
\begin{small}
\begin{tabular}{m{0.9in}m{0.37in}<{\centering}|m{0.55in}<{\centering}m{0.35in}<{\centering}m{0.55in}<{\centering}m{0.41in}<{\centering}m{0.5in}<{\centering}m{0.38in}<{\centering}m{0.45in}<{\centering}|m{0.38in}<{\centering} m{0.25in}<{\centering}}
\toprule
\midrule
\multirow{2}{0.9in}{\textbf{Model}} & \multirow{2}{0.4in}{\textbf{Verifier}} & \textbf{MMLUPro} & \textbf{GPQA} & \textbf{TheoremQA} & \textbf{WebInst.} & \textbf{MATH500} & \textbf{Minerva} & \textbf{AIME24} & \textbf{General} & \textbf{Math}\\
& & Avg@2 & Avg@4 & Avg@2 & Avg@2 & Avg@2 & Avg@2 & Avg@16 & - & -\\
\midrule
\multicolumn{11}{c}{\textbf{Qwen2.5-Base Models}} \\
\midrule
Qwen2.5-3B & -- & 34.6 & 26.3 & 27.4 & 44.5 & 42.6 & 19.4 & 0.0 & 33.2 & 21.3\\
+ GRPO & Rule & 50.1 & 30.4 & 37.6 & 61.1 & \textbf{60.8} & 35.4 & \textbf{10.0} & 44.8 & \textbf{35.4}\\
\rowcolor{cyan!20} \textbf{+ ICPO} & Rule & \textbf{51.7} & \textbf{32.9} & \textbf{39.1} & \textbf{63.2} & 60.4 & \textbf{36.5} & 9.4 & \textbf{46.7} & \textbf{35.4}\\
\midrule
Qwen2.5-7B & -- & 45.3 & 32.4 & 41.4 & 60.4 & 51.0 & 37.6 & 6.5 & 44.9 & 31.7\\
+ GRPO & Rule & 55.1 & 36.2 & 52.2 & 75.3 & \textbf{76.5} & 54.9 & 17.7 & 54.7 & 49.7 \\
\rowcolor{cyan!20} \textbf{+ ICPO} & Rule & \textbf{57.6} & \textbf{39.4} & \textbf{54.0} & \textbf{77.8} & 76.2 & \textbf{56.2} & \textbf{23.3} & \textbf{57.2} & \textbf{51.9} \\
\midrule
Qwen2.5-14B & -- & 51.2 & 32.8 & 43.0 & 66.1 & 55.6 & 41.7 & 10.0 & 48.3 & 35.8\\
+ GRPO & Rule & \textbf{63.2} & 43.6 & \textbf{58.8} & 79.0 & 80.4 & 59.2 & \textbf{20.0} & 61.1 & 53.2\\
\rowcolor{cyan!20} \textbf{+ ICPO} & Rule & 63.0 & \textbf{45.1} & \textbf{58.8} & \textbf{80.3} & \textbf{81.2} & \textbf{60.5} & \textbf{20.0} & \textbf{61.8} & \textbf{53.9}\\
\midrule
\multicolumn{11}{c}{\textbf{Qwen2.5-Instruct Models}} \\
\midrule
Qwen2.5-7B-Inst & -- & 54.5 & 34.2 & 47.3 & 72.6 & 75.4 & 49.4 & 9.4 & 52.2 & 44.7\\
+ GRPO & Rule & 52.0 & 40.1 & \textbf{56.3} & 63.2 & 77.0 & \textbf{55.8} & 17.7 & 52.9 & 50.2\\
\rowcolor{cyan!20} \textbf{+ ICPO} & Rule & \textbf{58.1} & \textbf{42.6} & 55.4 & \textbf{74.9} & \textbf{77.8} & \textbf{55.8} & \textbf{26.7} & \textbf{57.8} & \textbf{53.4}\\
\midrule
\multicolumn{11}{c}{\textbf{Qwen3-Base Models}} \\
\midrule
Qwen3-4B & -- & 50.5 & 36.8 & 33.7 & 55.7 & 54.1 & 29.4 & 3.33 & 44.2 & 28.9\\
+ GRPO & Rule & 57.3 & 41.5 & 53.3 & 65.8 & 78.5 & 54.7 & 16.7 & 54.5 & 49.9\\
\rowcolor{cyan!20} \textbf{+ ICPO} & Rule & \textbf{59.5} & \textbf{43.9} & \textbf{56.8} & \textbf{73.1} & \textbf{81.7} & \textbf{57.8} & \textbf{20.0} & \textbf{58.3} & \textbf{53.2}\\
\bottomrule
\end{tabular}
\vspace{-0.5em}
\caption{Overall performance comparison between ICPO (Our Method) and vanilla GRPO (Baseline) across models of different scales (3B, 7B and 14B), types (Base and Instruct) and versions (Qwen2.5 and Qwen3).}
\label{tab3}
\vspace{-1.3em}
\end{small}
\end{center}
\end{table*}

\subsection{Analysis of ICPO} \label{sec4.3}

\textbf{(1) Beyond Binary Rewards: Providing Continuous Optimization Signals}. We filtered the training data, keeping only samples where all instances were entirely correct or entirely incorrect, and used them as extremely sparse reward training set. As shown in Table~\ref{tab2}, compared to GRPO, ICPO still effectively explores under complete reward sparsity: it progressively identifies high-potential response paths and ultimately achieves stable performance gains. In contrast, GRPO, lacking fine-grained feedback, fails to pinpoint accurate policy update directions, resulting in stagnant or even deteriorating performance. These findings confirm that ICPO, through implicit contrastive modeling, effectively mitigates the exploration-exploitation dilemma in extremely sparse reward settings, offering a viable optimization approach for scenarios without explicit preference labels.

\begin{figure}[t]
  \begin{minipage}{0.49\linewidth}
    \includegraphics[width=\textwidth]{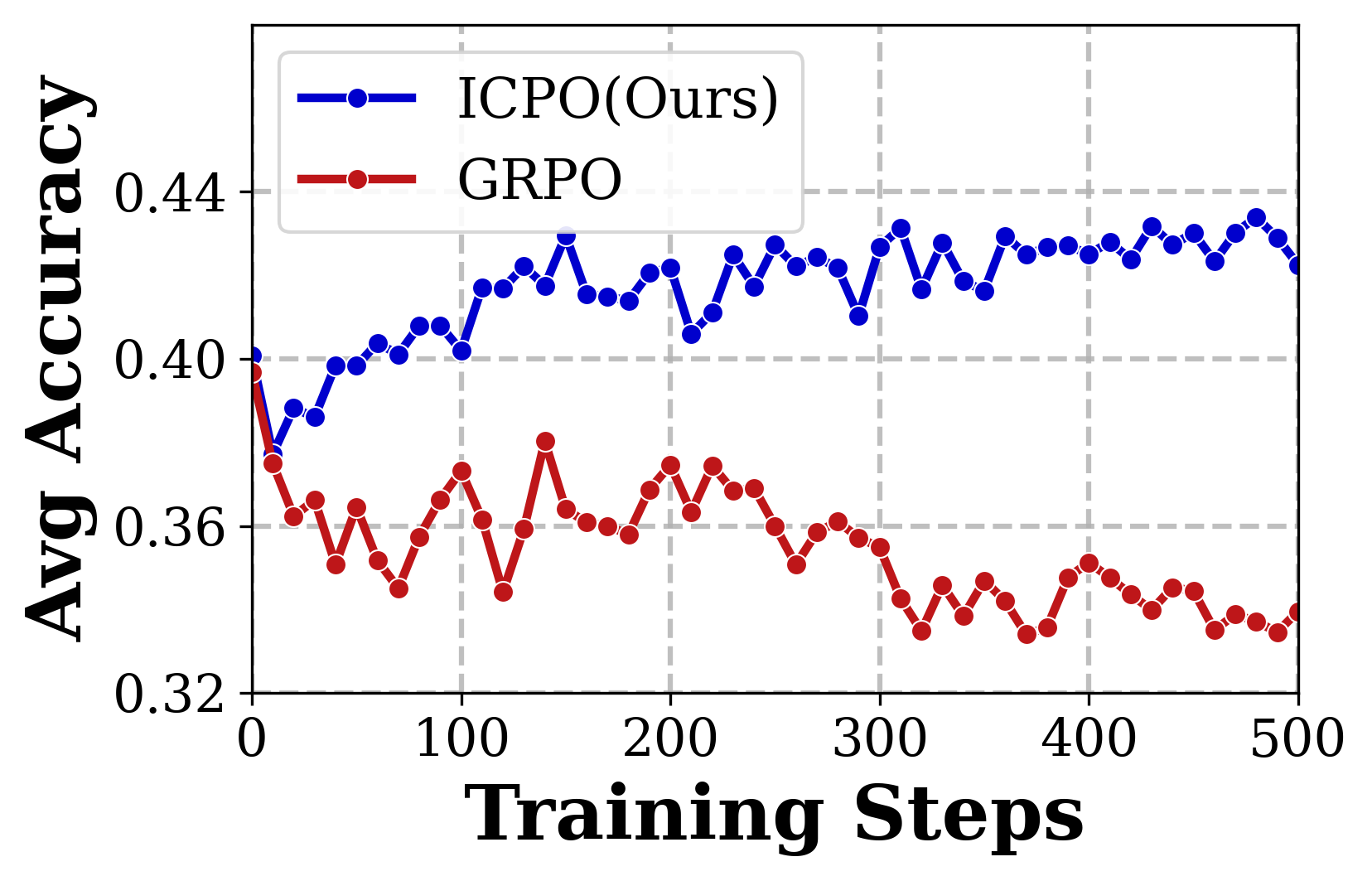}
    \label{fig4a}
  \end{minipage}
  \begin{minipage}{0.49\linewidth}
    \includegraphics[width=\textwidth]{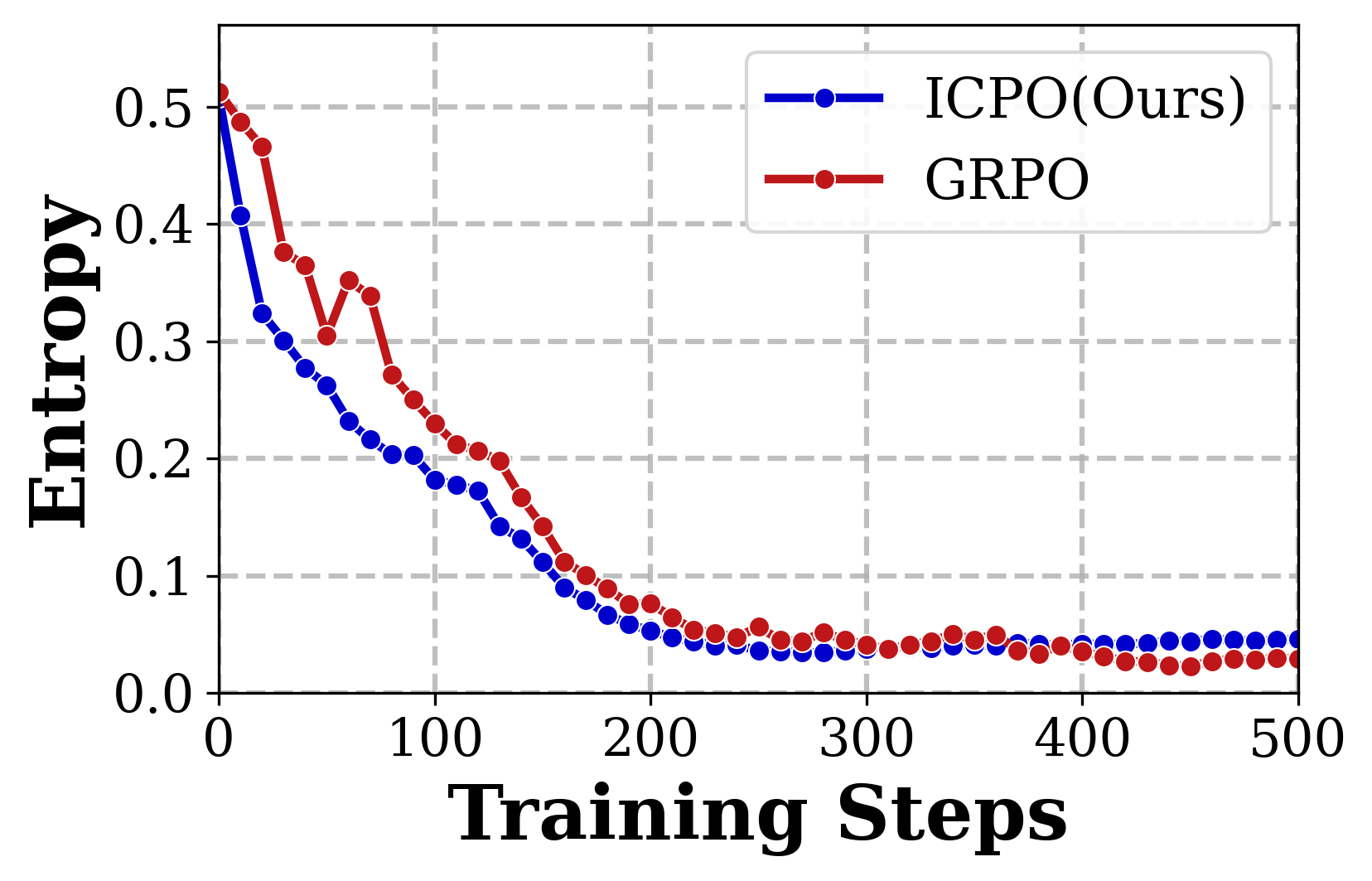}
    \label{fig4b}
  \end{minipage}
  \vspace{-1.5em}
  \caption{Comparison of Avg accuracy and entropy between vanilla GRPO (baseline) and ICPO (Our method) on seven benchmarks under noisy reward scenarios.}
  \label{fig4}
  \vspace{-1em}
\end{figure}

\begin{table*}[t]
\setlength{\tabcolsep}{4pt}        
\renewcommand{\arraystretch}{0.95}  
\begin{center}
    \begin{small}
    \begin{tabular}{m{1.2in}|m{0.55in}<{\centering}m{0.35in}<{\centering}m{0.55in}<{\centering}m{0.41in}<{\centering}m{0.5in}<{\centering}m{0.38in}<{\centering}m{0.45in}<{\centering}|m{0.38in}<{\centering} m{0.25in}<{\centering}}
    \toprule
        \midrule
        \multirow{2}{1.2in}{\textbf{Model}} & \textbf{MMLUPro} & \textbf{GPQA} & \textbf{TheoremQA} & \textbf{WebInst.} & \textbf{MATH500} & \textbf{Minerva} & \textbf{AIME24} & \textbf{General} & \textbf{Math}\\
        & Avg@2 & Avg@4 & Avg@2 & Avg@2 & Avg@2 & Avg@2 & Avg@16 & - & -\\
        \midrule
        Qwen2.5-7B-GRPO &  55.1 & 36.2 & 52.2 & 75.3 & \textbf{76.5} & 54.9 & 17.7 & 54.7 & 49.7 \\
        $\Rightarrow$ $\omega$ No decay & 54.8 & 35.8 & \textbf{54.0} & \underline{77.0} & 76.0 & \underline{55.9} & 16.7 & 55.4 & 49.5\\
        $\Rightarrow$ $\omega$ Linear decay & 56.0 & \underline{39.0} & \underline{53.5} & 75.8 & \underline{76.2} & 55.4 & \underline{20.0} & 56.1 & 50.5\\
        $\Rightarrow$ $\omega$ Warmup retention & \underline{57.0} & 38.7 & \underline{53.5} & 76.2 & \textbf{76.5} & 55.4 & \underline{20.0} & \underline{56.4} & \underline{50.6}\\
        \rowcolor{cyan!20} \textbf{$\Rightarrow$ $\omega$ Warmup decay} & \textbf{57.6} & \textbf{39.4} & \textbf{54.0} & \textbf{77.8} & \underline{76.2} & \textbf{56.2} & \textbf{23.3} & \textbf{57.2} & \textbf{51.9} \\
    \bottomrule
    \end{tabular}
    \vspace{-0.5em}
    \caption{Performance comparison of ICPO under different weight adjustment schemes for intrinsic confidence reward. The weight adjustment schemes adhere to those illustrated in Figure~\ref{fig2}.}
    \label{tab4}
    \vspace{-1.0em}
    \end{small}
\end{center}
\end{table*}

\textbf{(2) Correcting Reward Bias: Guiding Toward the Right Optimization Direction}. Figure~\ref{fig4} compares the performance of ICPO and vanilla GRPO on noisy reward scenarios. In scenarios without precise verification tools, LLMs automatically generate reward signals. However, due to LLMs' inherent uncertainty or hallucination issues, these rewards are often noisy and may even unfairly assign low scores to high-quality outputs due to assessment model bias. To validate the effectiveness of ICPO in such contexts, we randomly selected 40\% of the training data in each update round and injected random noise into their rewards to simulate this scenario. The experimental results demonstrate that ICPO, leveraging its intrinsic confidence-driven mechanism, effectively decouples the strong dependency between policy confidence and external rewards, maintaining positive reinforcement for high-potential responses and enabling effective learning even when external rewards are distorted. In contrast, GRPO, lacking accurate advantage signals, experiences a sharp increase in estimation variance, leading to rapid misalignment of policy update directions and severe performance degradation. These findings further corroborate the effectiveness of ICPO in mitigating reward noise.

\begin{figure}[t]
  \begin{minipage}{0.49\linewidth}
    \includegraphics[width=\textwidth]{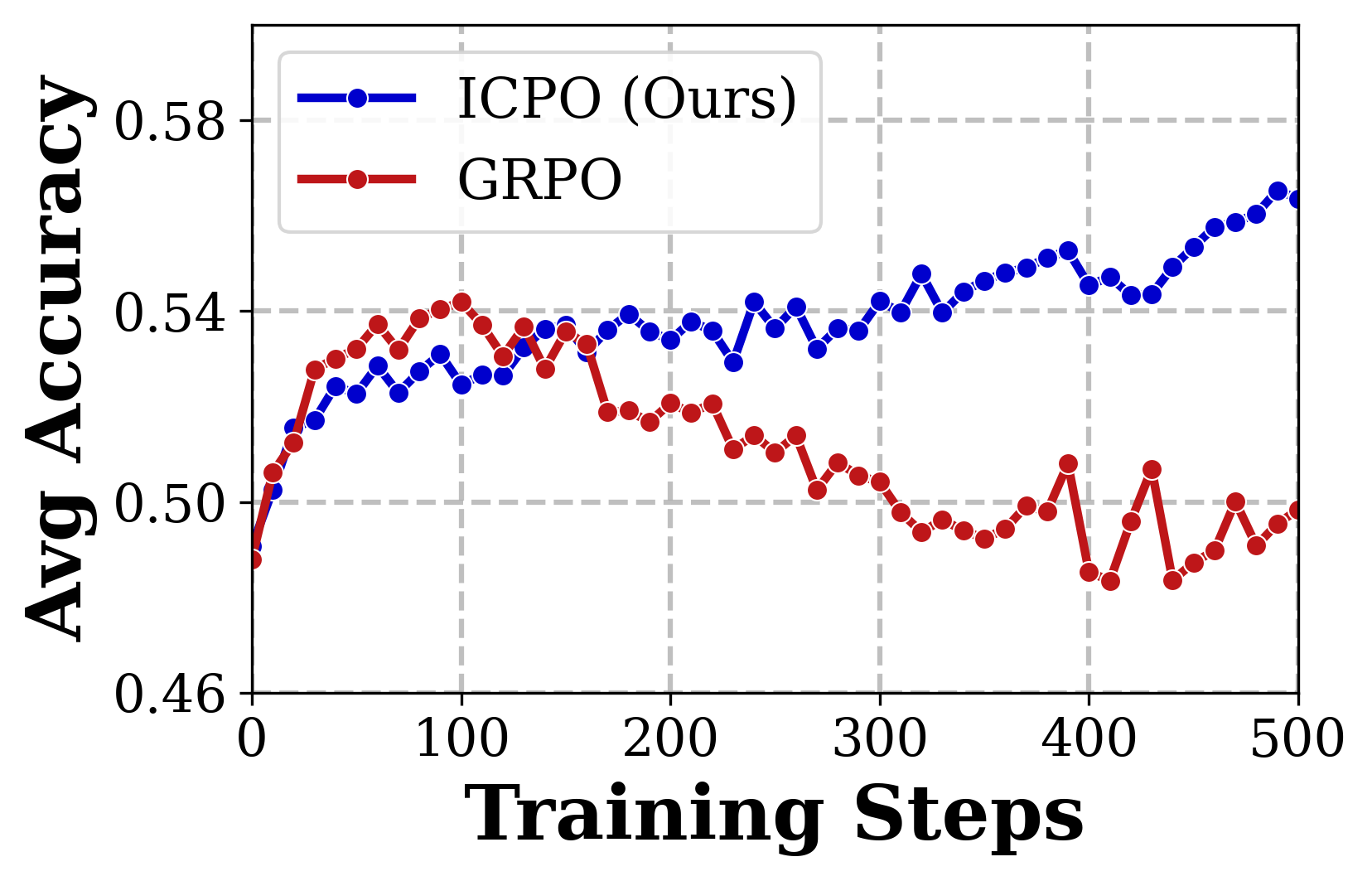}
    \label{fig5a}
  \end{minipage}
  \begin{minipage}{0.49\linewidth}
    \includegraphics[width=\textwidth]{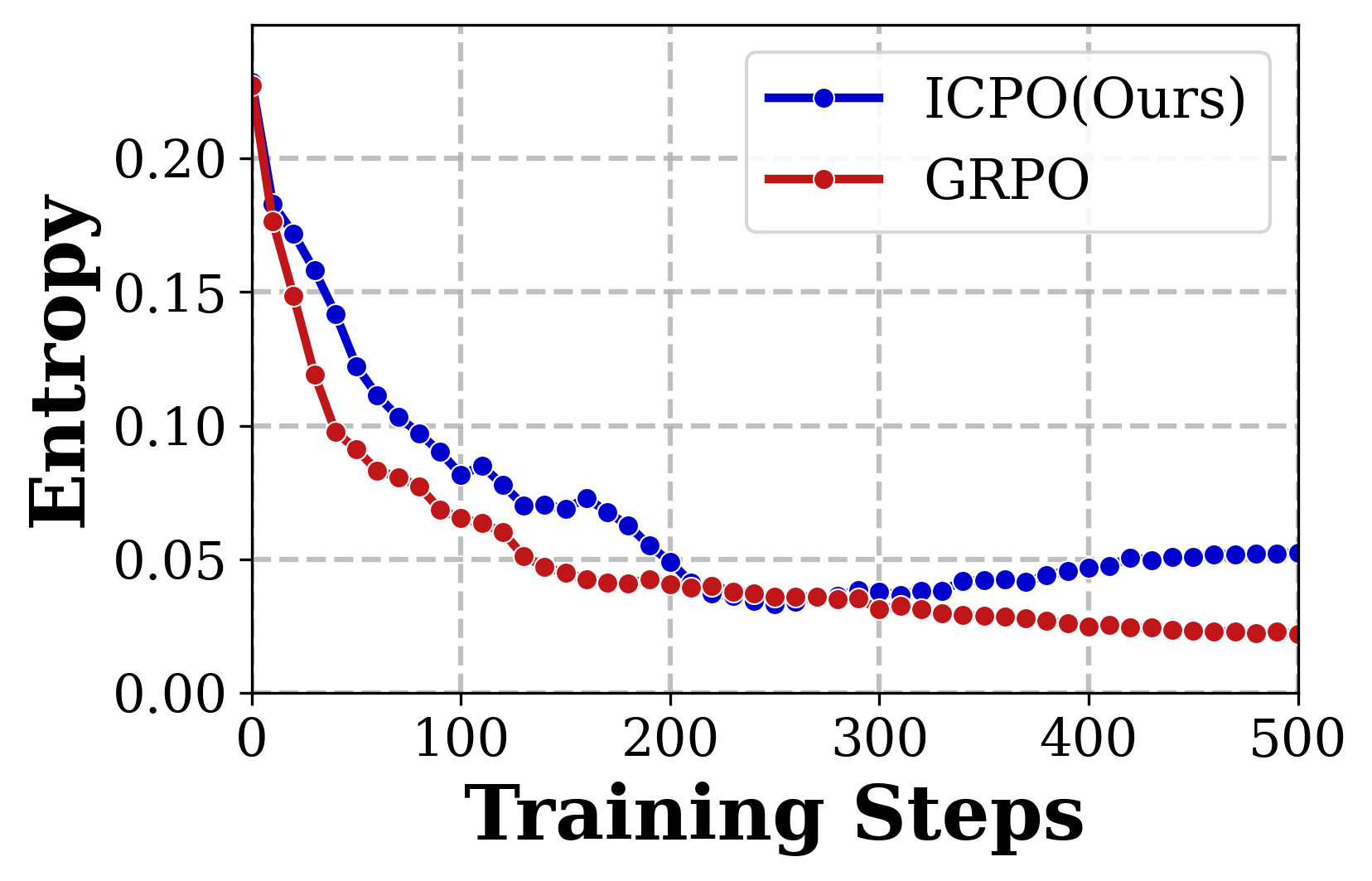}
    \label{fig5b}
  \end{minipage}
  \vspace{-1.5em}
  \caption{Comparison of Avg accuracy and entropy between vanilla GRPO (baseline) and ICPO (Our method) across seven benchmarks using Qwen2.5-7B-Instruct.}
  \label{fig5}
  \vspace{-1.0em}
\end{figure}

\subsection{Natural Scalability}
ICPO exhibits scalability and can effectively adapt to models of varying scales. As shown in Table~\ref{tab3}, with the growth of model parameters (from 3B to 7B, then to 14B), its performance keeps improving, fully reflecting ICPO's inherent scalability. An interesting finding is that ICPO shows the greatest performance gain on 7B-scale model, with smaller improvements on 3B and 14B-scale models. We posit that this may stem from the non-linear relationship between model scale and confidence calibration: small models lack confidence, providing uncertain probability signals for many questions, while large models tend to be overconfident~\citep{ConfRAG-2025}, assigning high confidence even to wrong answers. Consequently, both types of models exhibit relatively flat probability distributions, resulting in small in-group confidence differences.

Furthermore, ICPO continues to demonstrate its versatility and robustness on both the Qwen2.5-7B-Instruct model and the Qwen3-4B-base model. Whether applied to interactive models emphasizing instruction-following capabilities or foundational reasoning models focusing on thinking abilities, ICPO can seamlessly adapt and effectively enhance their performance in core tasks.

Notably, GRPO training on Qwen2.5-7B-Inst. shows sustained performance decline in later stages (as shown in Figure~\ref{fig5}), whereas ICPO maintains stable exploration. We attribute this to the substantial performance improvement of the instruction-tuned model, which results in smaller or even zero relative advantages among responses within groups. This further validating the robustness and anti-degradation benefits of intrinsic confidence-driven mechanism in extremely sparse reward scenarios.

\subsection{Investigate the Effect of Intrinsic Bonus}
Reward weight adjustment is crucial. We compared four adjustment schemes of $\omega$: No decay, linear decay, warmup retention, and warmup decay (as shown in Figure~\ref{fig2} (left)). Table~\ref{tab4} presents the performance under each scheme and uncovers two key findings: (1) Weight decay is essential since all decay schemes outperform the no decay baseline, facilitating a smooth shift from exploration to exploitation. (2) Early-stage warmup exploration matters. Warmup decay scheme expands the state-action coverage by gradually increasing the exploration intensity in early phase and ensures stable convergence through linear decay in later phase. In contrast, schemes without warmup exhibit an excessively high exploration intensity at the initial stage, leading to an imbalanced policy update direction and a consequent reduction in performance.

Additionally, as previously indicated by research, entropy offers a crucial perspective for understanding exploration capabilities~\cite{Ganqu-Cui-2025}. A sharp decline or high-level fluctuation in entropy typically indicates insufficient model exploration or policy failure. Figure~\ref{fig2} (right) presents a comparative analysis of entropy dynamics between vanilla GRPO and ICPO. Firstly, the intrinsic confidence-driven mechanism effectively mitigates the phenomenon of entropy collapse, fully demonstrating its role in promoting exploration. Secondly, when comparing different decay schemes, the warmup decay approach yields a more stable entropy trajectory. This finding aligns with our understanding: prioritizing warmup and appropriately decaying reward weights ensures stable convergence while effectively supporting the exploration process.

\section{Conclusion}
We propose an Intrinsic Confidence-Driven Group Relative Preference Optimization (ICPO), an efficient technique that enhances LLMs learning by modeling preferences based on the generation probabilities of responses within a group. ICPO is lightweight, requiring only minor adjustments to the original GRPO training framework. Its efficacy is demonstrated by consistent accuracy improvements across general and mathematical benchmarks, outperforming strong baselines. These results strongly support the intuition that intrinsic confidence contains rich preference information to effectively guide policy optimization.



\section*{Limitations}
Although ICPO can achieve effective learning in sparse reward settings and mitigate the problem of noisy rewards, it still encounters limitations and challenges when rewards are completely inaccurate. The fundamental limitation is that ICPO inherently promotes in-depth learning of correct but low-probability samples while curbing overconfidence in incorrect yet high-probability ones. When rewards completely deviate from the true task objectives, this \textbf{encouragement-suppression} mechanism experiences a total reversal. Specifically, genuinely valid low-probability behaviors are incorrectly marked as erroneous samples by the inaccurate rewards, while truly ineffective or even detrimental behaviors are falsely portrayed as correct but low-probability high-quality samples and become the focal point of ICPO's optimization, ultimately causing policy failure. We are currently actively exploring solutions to these issues and will present them in our future research.




\bibliography{custom}

\appendix

\begin{table*}[h]
	\centering
    \normalsize
	\renewcommand\arraystretch{1.0}
		\begin{tabular}{m{6.0in}}
			\toprule
			\rowcolor{gray!30} \textbf{ICPO training prompt} \\
			\midrule
            \textcolor{black!60}{\texttt{<|im\_start|>system}} \\
            \textcolor{black!60}{\texttt{A conversation between User and Assistant. The user asks a question,  and the Assistant solves it. The assistant first thinks about the reasoning process in the mind and then provides the user with the answer. The reasoning process and answer are enclosed within <think> </think> and <answer> </answer> tags, respectively, i.e., <think> reasoning process here </think> <answer> answer here </answer>.}} \\
            \textcolor{black!60}{\texttt{<|im\_end|>}} \\
            \textcolor{black!60}{\texttt{<|im\_start|>user}} \\
            \textcolor{black!60}{\texttt{\{\{question\}\}<|im\_end|>}} \\
            \textcolor{black!60}{\texttt{<|im\_start|>assistant}} \\
			\bottomrule
		\end{tabular}
    \caption{We adopt the training prompt of R1~\cite{DeepSeek-R1-2025} for ICPO.\label{tab5}}
    \vspace{-0.5em}
\end{table*}

\section{Experimental data}\label{sec:appendix1}

\subsection{Training data}

Our training data is identical to that employed in RLPR~\cite{RLPR-2025}, utilizing the high-quality reasoning question bank released by Ma et al.~\cite{General-Reasoner-2025}, which encompasses high-caliber reasoning problems across multiple complex domains. This training dataset excludes mathematics-related prompts to focus on general-domain reasoning and adopts a multi-stage filtering strategy to ensure its difficulty level: initially, history-related questions and those at primary/junior high school levels are removed to avoid overly simplistic or common-sense content; subsequently, only high-difficulty samples are retained based on reasoning scores assigned by GPT-4.1-mini. Ultimately, this process yields 77,687 focused and high-quality complex non-mathematical reasoning data entries.

\subsection{Evaluation data}
We evaluate reasoning capabilities through multiple general reasoning benchmarks and mathematical benchmarks. For mathematical reasoning, we have selected the following three benchmarks:
\begin{itemize}
\item MATH-500~\cite{MATH-500-2024} is a challenging benchmark specifically designed for evaluating the mathematical reasoning capabilities of LLMs. It comprises 500 high-quality, challenging mathematical problems, aiming to comprehensively assess a model's ability to solve complex mathematical issues.
\item Minerva~\cite{Minerva-2022}, developed by Google DeepMind, is a high-quality dataset for training mathematical reasoning models. It aims to narrow the capability gap of LLMs in quantitative reasoning, empowering models with mathematical thinking and symbolic reasoning skills from basic calculations to advanced scientific research levels.
\item AIME24 centers on intricate problems at the AIME (American Invitational Mathematics Examination) level, covering key areas like algebra, geometry, and combinatorics. It offers structured data, including problem descriptions, solution steps, and standard answers, setting a new standard for assessing AI models' logical reasoning skills.
\end{itemize}

For general domains, we adopt four benchmarks:
\begin{itemize}
\item MMLU-Pro~\cite{Mmlu-pro-2024} is a widely-adopted multi-task language understanding benchmark, encompassing challenging reasoning-intensive questions across multiple domains. To strike a balance between evaluation efficiency and data diversity, we randomly sampled 1,000 prompts from this benchmark for our experiments.
\item GPQA~\cite{GPQA-2023} encompasses graduate-level questions across multiple disciplines such as physics and chemistry. We employ the highest-quality GPQA-diamond subset for our evaluation.
\item TheoremQA~\cite{TheoremQA-2023} is designed to evaluate a model's ability to solve complex scientific problems by applying theorems. This benchmark comprises 800 high-quality questions, covering 350 theorems across fields such as mathematics and physics. We remove the 53 multimodal instructions.
\item WebInstruct: We derived a validation subset from WebInstruct~\cite{General-Reasoner-2025} to serve as an easily evaluable benchmark tailored for medium-sized models. Despite its relatively lower difficulty, the dataset can still evaluate multidisciplinary reasoning skills. Adhering to the filtering settings introduced in RLPR~\cite{RLPR-2025}, we conducted our evaluation using 638 unique questions.
\end{itemize}

\begin{table*}[t]
    \centering
    \label{tab:configs}
    \begin{subtable}{.45\textwidth}
      \centering
        \begin{tabular}{lcc}
            \toprule
            \rowcolor{gray!30} \textbf{Config} & \textbf{GRPO} & \textbf{ICPO} \\
            \midrule
            actor-lr & 1e-6 & 1e-6 \\
            kl\_coef & 0.001 & 0.001 \\
            max\_prompt\_length & 2K & 2K \\
            max\_response\_length & 3K & 3K \\
            train\_batch\_size & 768 & 768 \\
            ppo\_mini\_batch\_size & 192 & 192 \\
            clip\_ratio & 0.20 & 0.20 \\
            sample\_temperature & 1.0 & 1.0 \\
            rollout.n & 5 & 5 \\
            total\_training\_steps & 500 & 500 \\
            \bottomrule
        \end{tabular}
        \caption{}
        \label{tab6:subtab1}
    \end{subtable}%
    \begin{subtable}{.54\textwidth}
      \centering
        \begin{tabular}{lccc}
            \toprule
            \rowcolor{gray!30} \textbf{Config} & \textbf{3B} & \textbf{7B} & \textbf{14B} \\
            \midrule
            $\delta$ & 0.5 & 0.4 & 0.3 \\
            $\tau$ & 2.0 & 2.0 & 2.0 \\
            $\omega$ & \small{Warmup decay} & \small{Warmup decay} & \small{Warmup decay} \\
            \bottomrule
        \end{tabular}
        \caption{}
        \label{tab6:subtab2}
    \end{subtable}
    \caption{(a) General Training Configurations for the Qwen2.5 Series Models: Training parameters shared by both ICPO and GRPO. These settings are uniformly applied across all methods based on GRPO and ICPO (e.g., "Qwen2.5-3B (7B, 14B)-GRPO" and "Qwen2.5-3B (7B, 14B)-ICPO" in Table~\ref{tab3}). (b) ICPO-Specific Training Configurations for the Qwen2.5 Series Models: Unique training parameters exclusive to ICPO.}
    \label{tab6}
    \vspace{-0.5em}
\end{table*}

\section{Experimental Setup}\label{sec:appendix2}

\subsection{Training Setup}
Unless otherwise specified, all our experiments were conducted on the Qwen2.5-7B model. Following the majority of RLVR practices, we omitted the supervised fine-tuning process and directly performed post-training on the base model. In the main experiment, we controlled the output structure to extract parsable reasoning chains and answers by adjusting the prompt templates during both the training and validation phases. The specific prompt templates are detailed in Table~\ref{tab5}. All experimental results presented in this paper are averaged over multiple trials.

In the experiments targeting the Gemma and Llama models, we adjusted the temperature parameters for both training and evaluation to 0.6. Following RLPR~\citep{RLPR-2025} guidelines, we removed \verb|<think>| segments from templates to prevent a decline in generation quality. We observed that rule-based scoring scripts introduced errors in benchmark tests containing non-multiple-choice formatted questions. To address this issue, we deployed a Qwen2.5-7B-Instruct model server for evaluation and additionally utilized Qwen2.5-72B-Instruct to handle more complex benchmarks, such as TheoremQA and Minerva.

In the experiment focusing on sparse reward scenarios, we employed Qwen2.5-7B model to conduct a single-epoch training session on the training set. We retained the original data where all responses within each group were either uniformly correct or uniformly incorrect, while filtering out those data instances where responses within the same group exhibited discrepancies in rewards. Subsequently, we utilized the retained data as extremely sparse reward training set to further train Qwen2.5-7B model, thereby validating the efficacy of ICPO in extremely sparse reward settings.

In the experiment concerning noisy reward scenarios, we randomly selected 40\% of the responses generated at each training step and injected random noise into their rewards. Specifically, we randomly added or subtracted 0.3 points from the original reward values to simulate the hallucinations and unstable fluctuations that may occur when LLMs assign scores. Subsequently, we utilized the noise-injected rewards to calculate the intra-group advantage for guiding policy updates, thereby validating the comparative performance of ICPO and vanilla GRPO in noisy reward environments.

\begin{figure*}[t]
  \begin{minipage}{0.33\textwidth}
    \includegraphics[width=\textwidth]{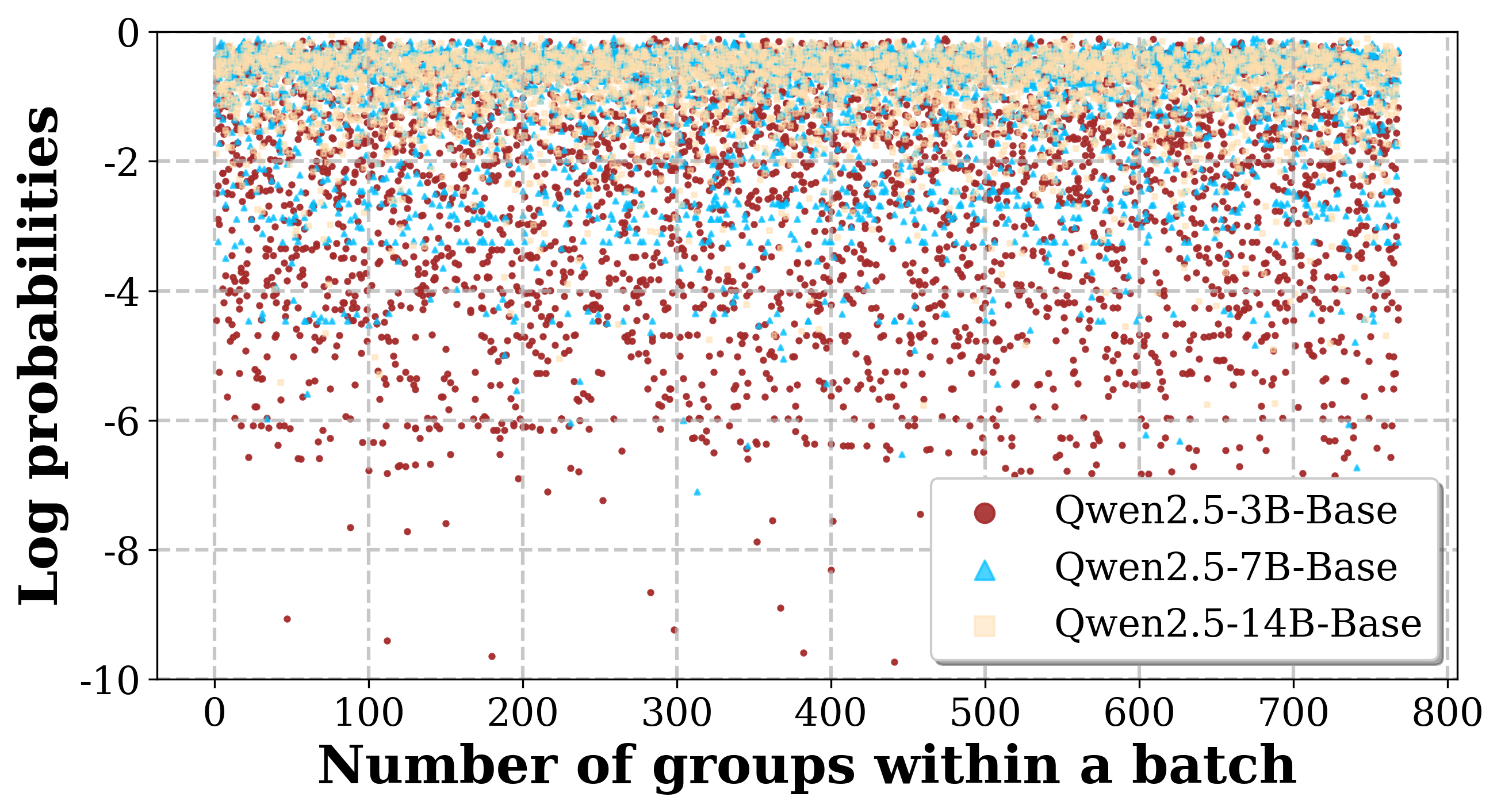}
    \label{fig3a}
  \end{minipage}
  \begin{minipage}{0.33\textwidth}
    \includegraphics[width=\textwidth]{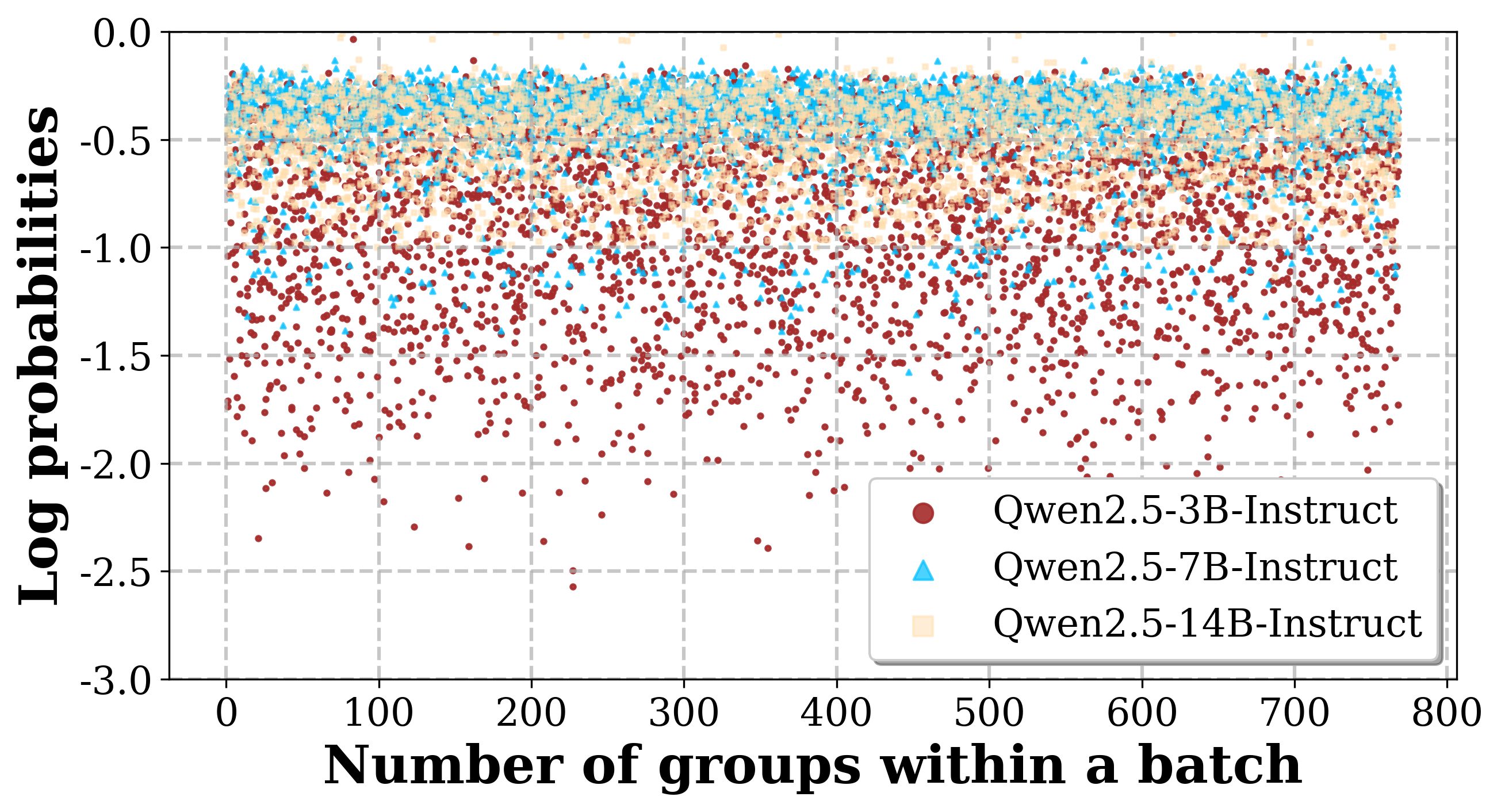}
    \label{fig3b}
  \end{minipage}
  \begin{minipage}{0.33\textwidth}
    \includegraphics[width=\textwidth]{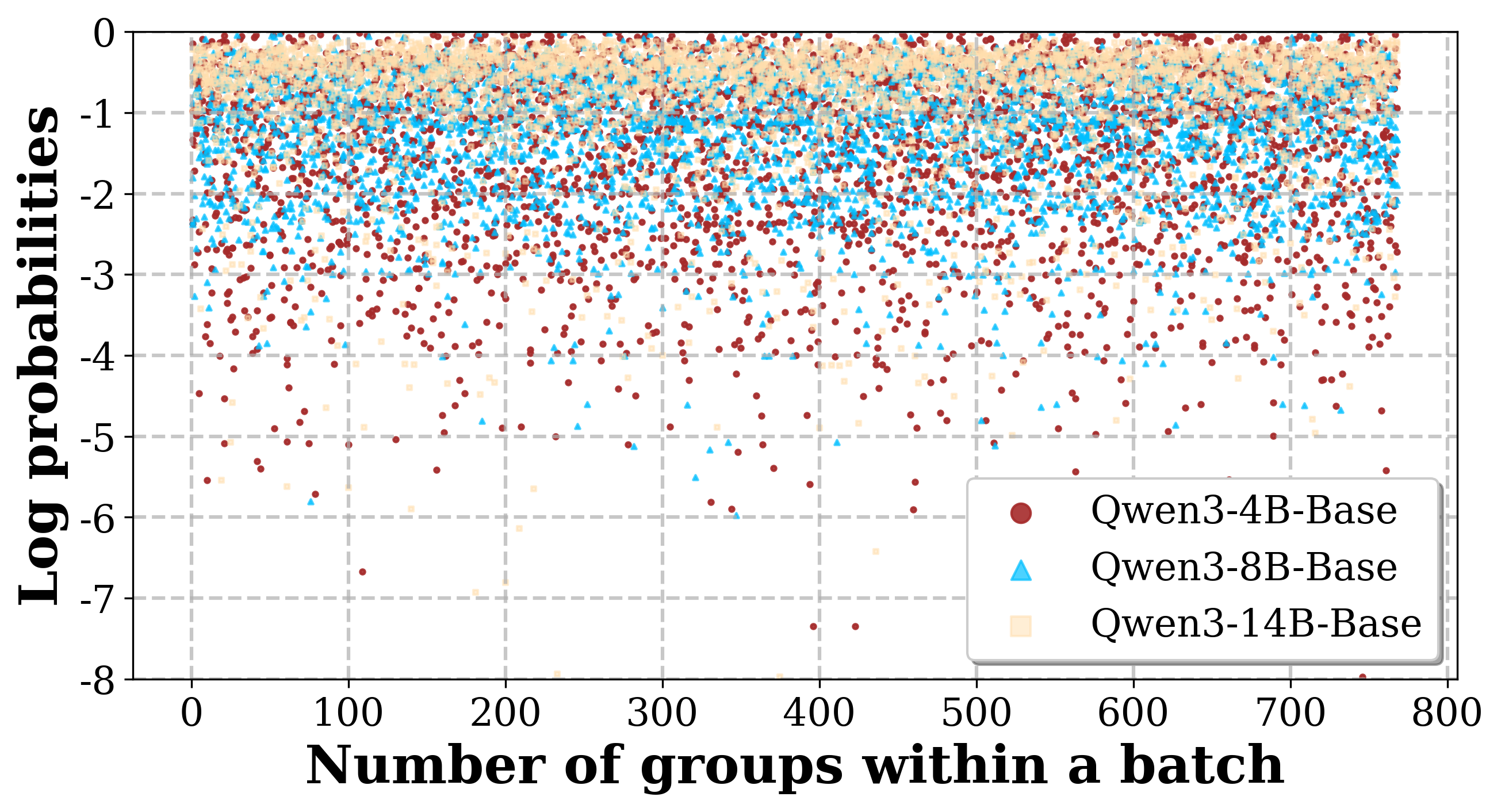}
    \label{fig3a}
  \end{minipage}
  \vspace{-1.5em}
  \caption{Log-probability distribution of output responses from models of different scales. From left to right are the Qwen2.5-Base series, Qwen2.5-Instruct series, and Qwen3-Base series.}
  \label{fig6}
  \vspace{-0.5em}
\end{figure*}

\subsection{Parameters Setup}
We employed Verl as our training framework\footnote{\url{https://github.com/volcengine/verl}}. Table~\ref{tab6} presents the configuration schemes adopted for training both ICPO and vanilla GRPO based on Qwen2.5 series models. All experiments were conducted on a single node, with each node equipped with 8 NVIDIA H200 141GB GPUs.

\section{Hyperparameter Search Experiment}
\label{sec:appendix3}
We conducted additional hyperparameter search experiments focusing on the following hyperparameters: (1) Selection method of parameter $\delta$; (2) Impact of $\tau$ on model performance; and (3) Analysis of the weight adjustment approach for $\omega$.

\subsection{Selection method of parameter $\delta$}
Table~\ref{tab7} presents a performance comparison of $\delta$ across models of different types and scales. Through analysis of the experimental results, we observe that the optimal value of $\delta$ varies across models of different scales. However, for models of different types (Qwen2.5-Base, Qwen2.5-Instruct, and Qwen3-Base), the optimal values of $\delta$ are approximately the same under the same scale.

\begin{table*}[t]
\setlength{\tabcolsep}{4pt}        
\renewcommand{\arraystretch}{0.98}  
\begin{center}
\begin{small}
\begin{tabular}{m{1.3in}|m{0.6in}<{\centering}|m{0.6in}<{\centering}m{0.6in}<{\centering}m{0.6in}<{\centering}m{0.6in}<{\centering}m{0.6in}<{\centering}}
\toprule
\midrule
\multirow{2}{1.3in}{\textbf{Model}} & \multirow{2}{0.4in}{\textbf{GRPO}} & \multicolumn{5}{c}{\cellcolor{gray!30}\textbf{ICPO}}\\
& & \cellcolor{gray!30} $\delta=0.2$ & \cellcolor{gray!30} $\delta=0.3$ & \cellcolor{gray!30} $\delta=0.4$ & \cellcolor{gray!30} $\delta=0.5$ & \cellcolor{gray!30} $\delta=0.6$ \\
\midrule
\multicolumn{7}{c}{\textbf{Qwen2.5-Base Models}} \\
\midrule
\cellcolor{gray!30} Qwen2.5-3B-Base & 40.10 & 40.30 & 40.88 & 40.70 & \cellcolor{cyan!20} \textbf{41.10} & 40.85 \\
\cellcolor{gray!30} Qwen2.5-7B-Base & 52.20 & 52.03 & 54.23 & \cellcolor{cyan!20}  \textbf{54.90} & 53.65 & 52.90 \\
\cellcolor{gray!30} Qwen2.5-14B-Base & 57.05 & 57.63 & \cellcolor{cyan!20}  \textbf{57.85} & 57.24 & 57.30 & 56.45 \\
\midrule
\multicolumn{7}{c}{\textbf{Qwen2.5-Instruct Models}} \\
\midrule
\cellcolor{gray!30} Qwen2.5-3B-Instruct & 42.40 & 42.15 & 43.33 & 43.67 & \cellcolor{cyan!20}  \textbf{44.10} & 43.90 \\
\cellcolor{gray!30} Qwen2.5-7B-Instruct & 51.55 & 53.65 & 55.47 & \cellcolor{cyan!20} \textbf{55.60} & 54.70 & 51.90 \\
\cellcolor{gray!30} Qwen2.5-14B-Instruct & 60.01 & 61.05 & \cellcolor{cyan!20} \textbf{61.40} & 60.38 & 59.70 & 60.13 \\
\midrule
\multicolumn{7}{c}{\textbf{Qwen3-Base Models}} \\
\midrule
\cellcolor{gray!30} Qwen3-4B-Base & 49.55 & 52.20 & 53.35 & \cellcolor{cyan!20} \textbf{54.05} & 53.90 & 52.67 \\
\cellcolor{gray!30} Qwen3-8B-Base & 57.30 & 59.27 & 59.15 & \cellcolor{cyan!20} \textbf{59.93} & 58.80 & 57.33 \\
\cellcolor{gray!30} Qwen3-14B-Base & 58.93 & 59.70 & \cellcolor{cyan!20} \textbf{60.57} & 59.48 & 59.20 & 58.75 \\
\bottomrule
\end{tabular}
\vspace{-0.5em}
\caption{The impact of parameter $\delta$ on performance across models of different scales (3B, 7B and 14B), types (Base and Instruct) and versions (Qwen2.5 and Qwen3), where the performance is the average across seven benchmarks.}
\label{tab7}
\vspace{-1em}
\end{small}
\end{center}
\end{table*}

We analyze that the possible reason lies in the fact that the probability distributions of model outputs under the same scale are roughly similar, even when the model types differ. To verify this hypothesis, we conducted a statistical analysis of the logarithmic probability distributions output by different models and plotted scatter diagrams, as illustrated in Figure~\ref{fig6}. The experimental results robustly validate our hypothesis and reveal a key insight: Smaller models exhibit low confidence, providing uncertain probabilistic signals for many queries, whereas larger models tend toward overconfidence, assigning high probabilities even to incorrect answers. Models of moderate scale (4B-8B parameters) demonstrate optimal consistency in response probability distributions. Moreover, for models of comparable scales, their output probability distributions exhibit substantial similarity. Consequently, the configuration of parameter $\delta$ demonstrates general applicability across different model types. Specifically, we recommend setting $\delta=0.5$ for models with fewer than 4B parameters, $\delta=0.4$ for models ranging from 4B to 8B parameters, and $\delta=0.3$ for models exceeding 8B parameters.

\subsection{Impact of parameter $\tau$ on performance}
\begin{figure}[t]
  \begin{minipage}{0.49\linewidth}
    \includegraphics[width=\textwidth]{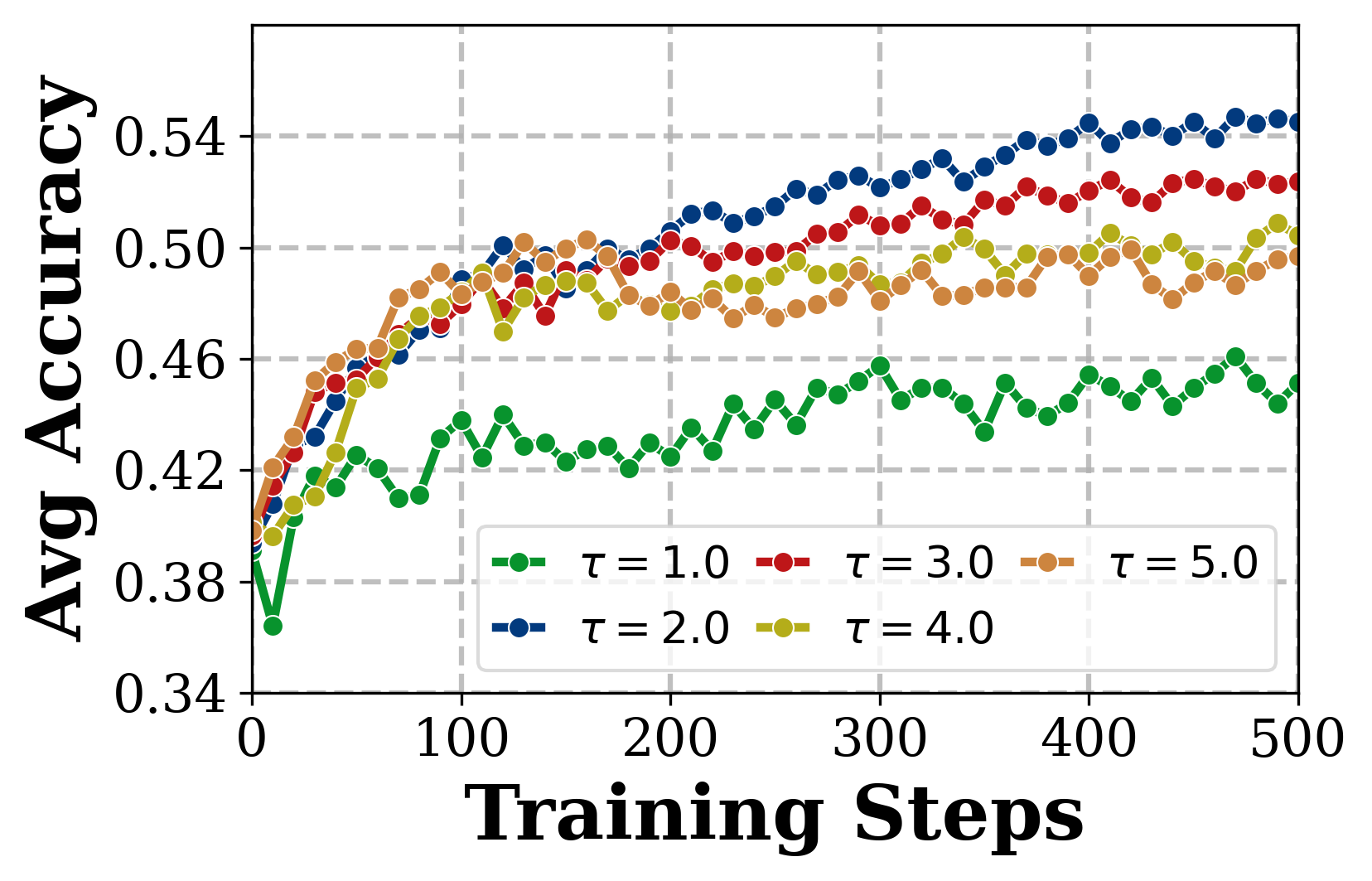}
    \label{fig2a}
  \end{minipage}
  \begin{minipage}{0.49\linewidth}
    \includegraphics[width=\textwidth]{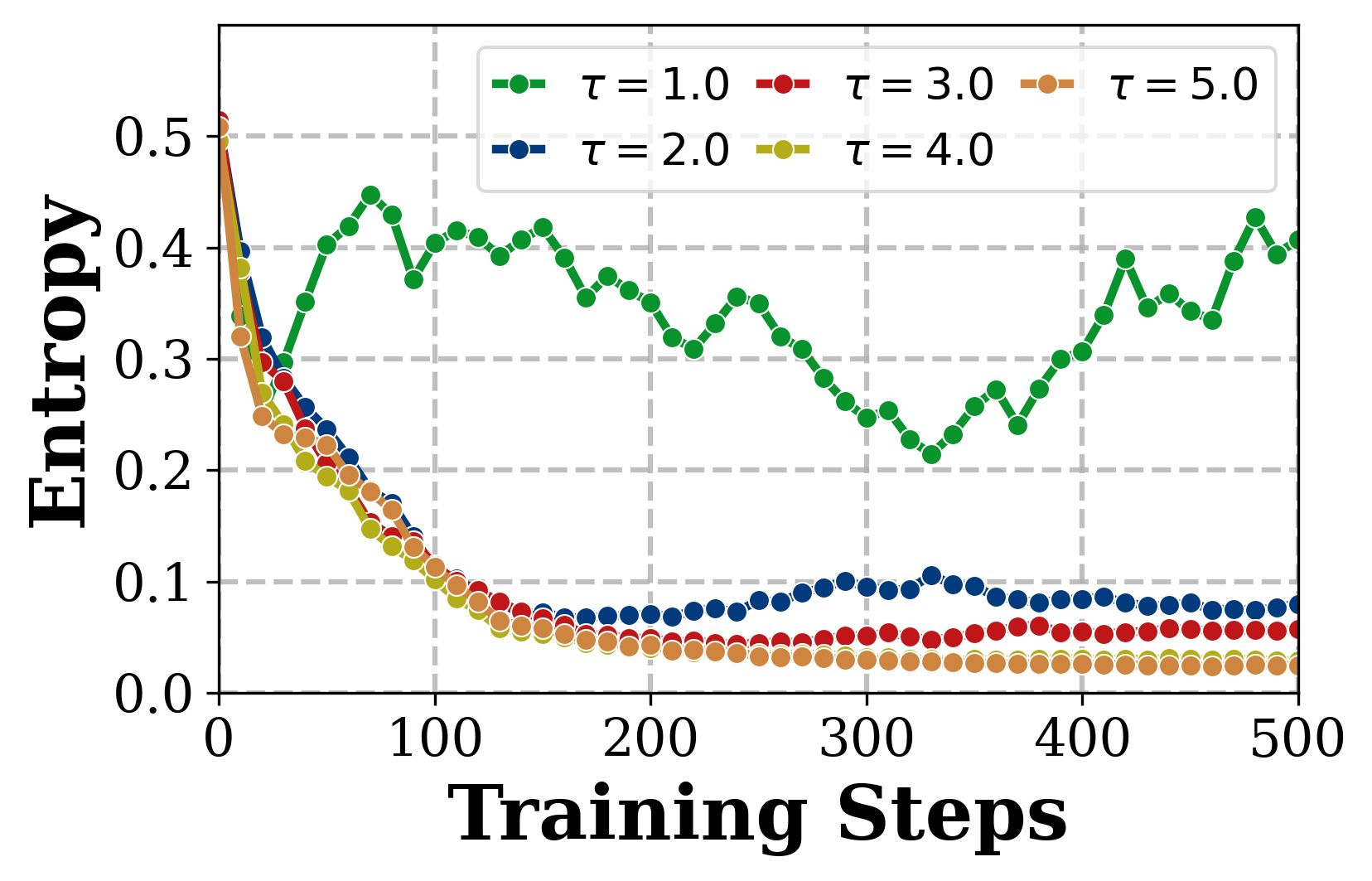}
    \label{fig2b}
  \end{minipage}
  \vspace{-1.5em}
  \caption{Impact of parameter $\tau$ on performance.}
  \label{fig7}
  \vspace{-1.0em}
\end{figure}

The core function of parameter $\tau$ is to regulate the weight allocation of intrinsic confidence rewards within the overall reward, thereby determining its proportion of influence on the policy update process. If $\tau$ is set too large, the policy update will be excessively dominated by intrinsic preferences, making it difficult to ensure the performance effectiveness of the policy in real-world task scenarios. Conversely, if $\tau$ is set too small, the guiding role of intrinsic preferences in the optimization process will be weakened, failing to effectively calibrate the direction of policy updates. We set the values of parameter $\tau$ to 1.0, 2.0, 3.0, 4.0, and 5.0 respectively, and observed the changes in the performance of ICPO under different settings. The experimental results are shown in Figure~\ref{fig7}.

From the results, we observe that the model achieves optimal performance when the parameter $\tau=2.0$, with $\tau=3.0$ yielding the second-best yet still competitive performance. In contrast, setting $\tau=1.0$ leads to a significant degradation in both policy entropy and overall performance, indicating that over-reliance on intrinsic preferences for guiding policy updates destabilizes training. When $\tau$ is further increased to 4.0 or 5.0, the model exhibits rapid performance gains in the early stages of training but subsequently plateaus, accompanied by a sharp decline in policy entropy. This suggests that excessively downweighting the influence of intrinsic preferences during policy updates undermines the agent’s exploratory capacity. Collectively, these findings demonstrate that appropriately leveraging intrinsic preferences to guide policy updates helps mitigate entropy collapse and effectively preserves policy exploration.

\subsection{Analysis of Weight Adjustment Strategies}\label{appendixc3}
\begin{figure}[t]
  \begin{minipage}{0.49\linewidth}
    \includegraphics[width=\textwidth]{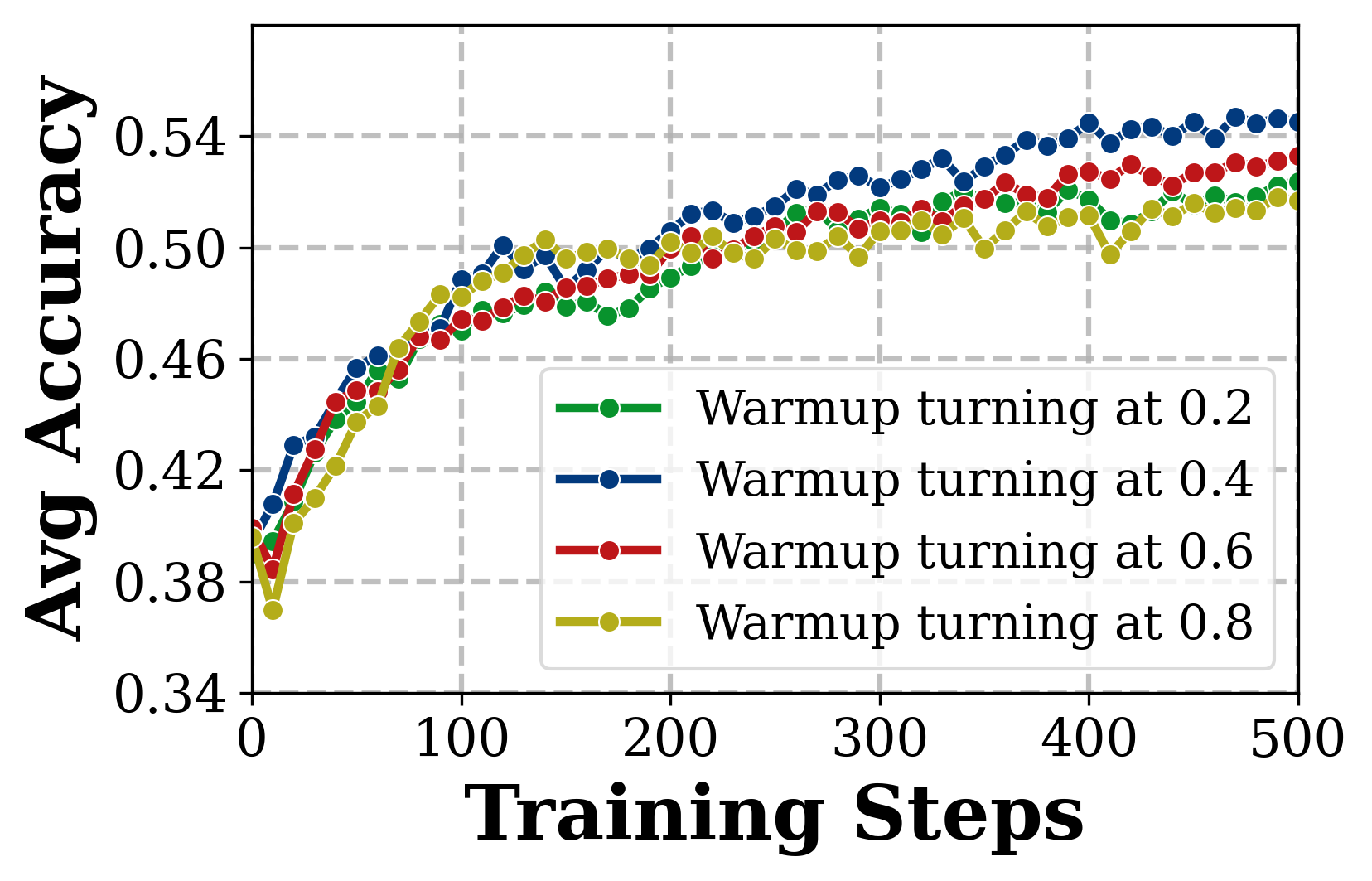}
    \label{fig2a}
  \end{minipage}
  \begin{minipage}{0.49\linewidth}
    \includegraphics[width=\textwidth]{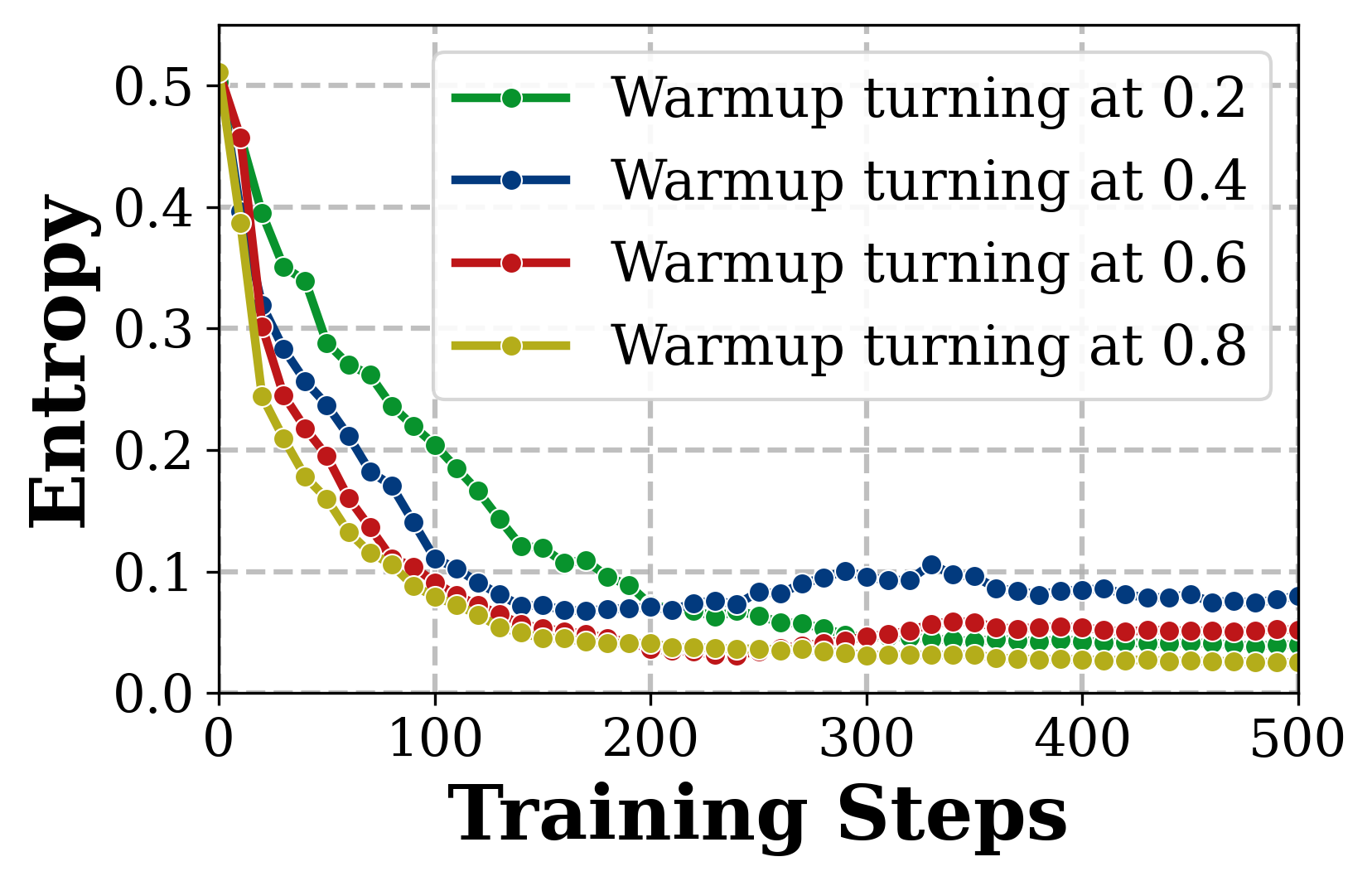}
    \label{fig2b}
  \end{minipage}
  \vspace{-1.5em}
  \caption{Parameter search for warm-up turning point.}
  \label{fig8}
  \vspace{-1.0em}
\end{figure}

Table~\ref{tab8} presents the hyperparameter search experiments for the weight adjustment strategies across different models. Figure~\ref{fig8} illustrates the parameter search experiment for the warm-up turning point. From these, we observe that: (1) Weight decay mechanism is of critical importance. For Qwen2.5-7B-Base, Qwen2.5-7B-Instruct, and Qwen3-4B-Base, all decay schemes outperform the no-decay baseline, enabling a smooth transition from exploration to exploitation. (2) For the Qwen2.5-7B-Base and Qwen3-4B-Base models, warm-up exploration in the early stages is crucial. However, for the Qwen2.5-7B-Instruct model, the impact of warm-up exploration on performance improvement is not significant. (3) For the Qwen2.5-7B-Base model, the warm-up decay weight adjustment method achieves optimal performance when the warm-up turning point is set to 0.4. Both arriving at the warm-up turning time point earlier or later will exert a certain impact on the exploration-exploitation effectiveness of the strategy.

\begin{table*}[t]
\setlength{\tabcolsep}{4pt}        
\renewcommand{\arraystretch}{0.95}  
\begin{center}
    \begin{small}
    \begin{tabular}{m{1.5in}|m{0.55in}<{\centering}m{0.35in}<{\centering}m{0.55in}<{\centering}m{0.41in}<{\centering}m{0.5in}<{\centering}m{0.38in}<{\centering}m{0.45in}<{\centering}|m{0.38in}<{\centering}}
    \toprule
        \midrule
        \multirow{2}{1.5in}{\textbf{Model}} & \textbf{MMLUPro} & \textbf{GPQA} & \textbf{TheoremQA} & \textbf{WebInst.} & \textbf{MATH500} & \textbf{Minerva} & \textbf{AIME24} & \textbf{All}\\
        & Avg@2 & Avg@4 & Avg@2 & Avg@2 & Avg@2 & Avg@2 & Avg@16 & - \\
        \midrule
        \multicolumn{9}{c}{\textbf{Bonus Weight Adjustment Schedules}} \\
        \midrule
        Qwen2.5-7B-Base-GRPO &  55.1 & 36.2 & 52.2 & 75.3 & \textbf{76.5} & 54.9 & 17.7 & 52.20 \\
        $\Rightarrow$ $\omega$ No decay & 54.8 & 35.8 & \textbf{54.0} & \underline{77.0} & 76.0 & \underline{55.9} & 16.7 & 52.89 \\
        $\Rightarrow$ $\omega$ Linear decay & 56.0 & \underline{39.0} & \underline{53.5} & 75.8 & \underline{76.2} & 55.4 & \underline{20.0} & 53.70 \\
        $\Rightarrow$ $\omega$ Warmup retention & \underline{57.0} & 38.7 & \underline{53.5} & 76.2 & \textbf{76.5} & 55.4 & \underline{20.0} & \underline{53.90} \\
        \rowcolor{cyan!20} \textbf{$\Rightarrow$ $\omega$ Warmup decay} & \textbf{57.6} & \textbf{39.4} & \textbf{54.0} & \textbf{77.8} & \underline{76.2} & \textbf{56.2} & \textbf{23.3} & \textbf{54.90} \\
        \midrule
        Qwen2.5-7B-Inst.-GRPO &  52.0 & 40.1 & \textbf{56.3} & 63.2 & 77.0 & \textbf{55.8} & 17.7 & 51.73 \\
        $\Rightarrow$ $\omega$ No decay & 57.7 & 42.2 & 52.1 & \underline{76.8} & 74.5 & 52.6 & 20.0 & 53.70 \\
        $\Rightarrow$ $\omega$ Linear decay & \textbf{58.3} & \textbf{43.8} & 54.7 & \textbf{77.3} & \underline{77.0} & \underline{54.2} & \underline{23.3} & \underline{55.51} \\
        $\Rightarrow$ $\omega$ Warmup retention & 57.4 & 42.0 & 54.3 & 75.2 & 75.7 & 53.5 & 20.0 & 54.01 \\
        \rowcolor{cyan!20} \textbf{$\Rightarrow$ $\omega$ Warmup decay} & \underline{58.1} & \underline{42.6} & \underline{55.4} & 74.9 & \textbf{77.8} & \textbf{55.8} & \textbf{26.7} & \textbf{55.90} \\
        \midrule
        Qwen3-4B-Base-GRPO &  57.3 & 41.5 & 53.3 & 65.8 & 78.5 & 54.7 & \underline{16.7} & 52.54 \\
        $\Rightarrow$ $\omega$ No decay & 53.6 & 40.8 & 54.5 & 65.6 & 79.2 & 54.8 & \underline{16.7} & 52.17 \\
        $\Rightarrow$ $\omega$ Linear decay & \underline{58.1} & 43.2 & \underline{56.1} & \underline{71.4} & \textbf{82.2} & 55.7 & \textbf{20.0} & \underline{55.24} \\
        $\Rightarrow$ $\omega$ Warmup retention & 57.5 & \underline{43.5} & 55.7 & 68.2 & 80.8 & \underline{56.9} & \textbf{20.0} & 54.66 \\
        \rowcolor{cyan!20} \textbf{$\Rightarrow$ $\omega$ Warmup decay} & \textbf{59.5} & \textbf{43.9} & \textbf{56.8} & \textbf{73.1} & \underline{81.7} & \textbf{57.8} & \textbf{20.0} & \textbf{56.11} \\
    \bottomrule
    \end{tabular}
    \vspace{-0.5em}
    \caption{Performance comparison of ICPO under different weight adjustment schemes for intrinsic confidence reward across different models. The weight adjustment schemes adhere to those illustrated in Figure~\ref{fig2}.}
    \label{tab8}
    \vspace{-0.5em}
    \end{small}
\end{center}
\end{table*}

\begin{figure*}[t]
  \begin{minipage}{0.33\textwidth}
    \includegraphics[width=\textwidth]{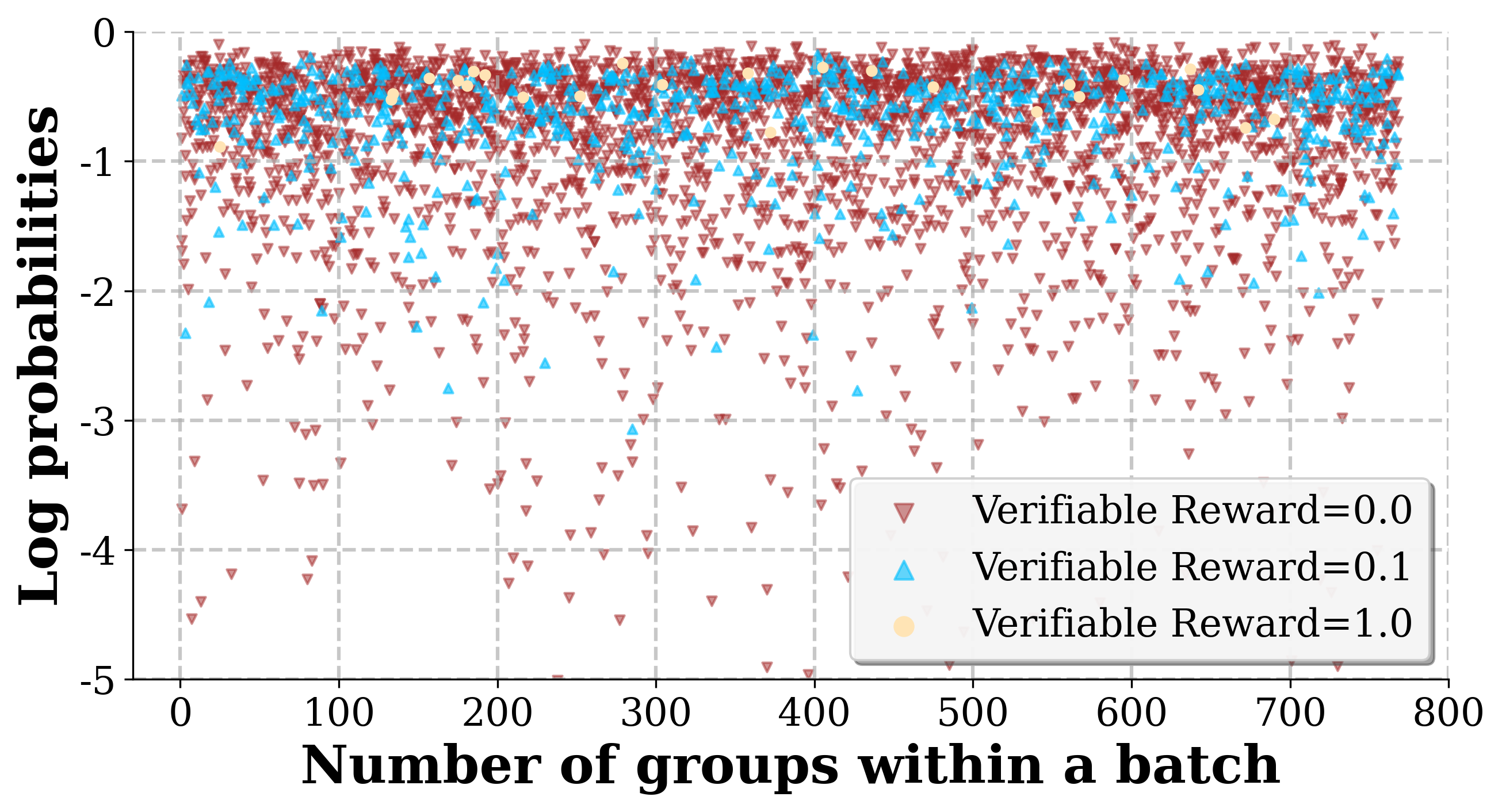}
    \subcaption{0\% of entire training process}
    \label{fig3a}
  \end{minipage}
  \begin{minipage}{0.33\textwidth}
    \includegraphics[width=\textwidth]{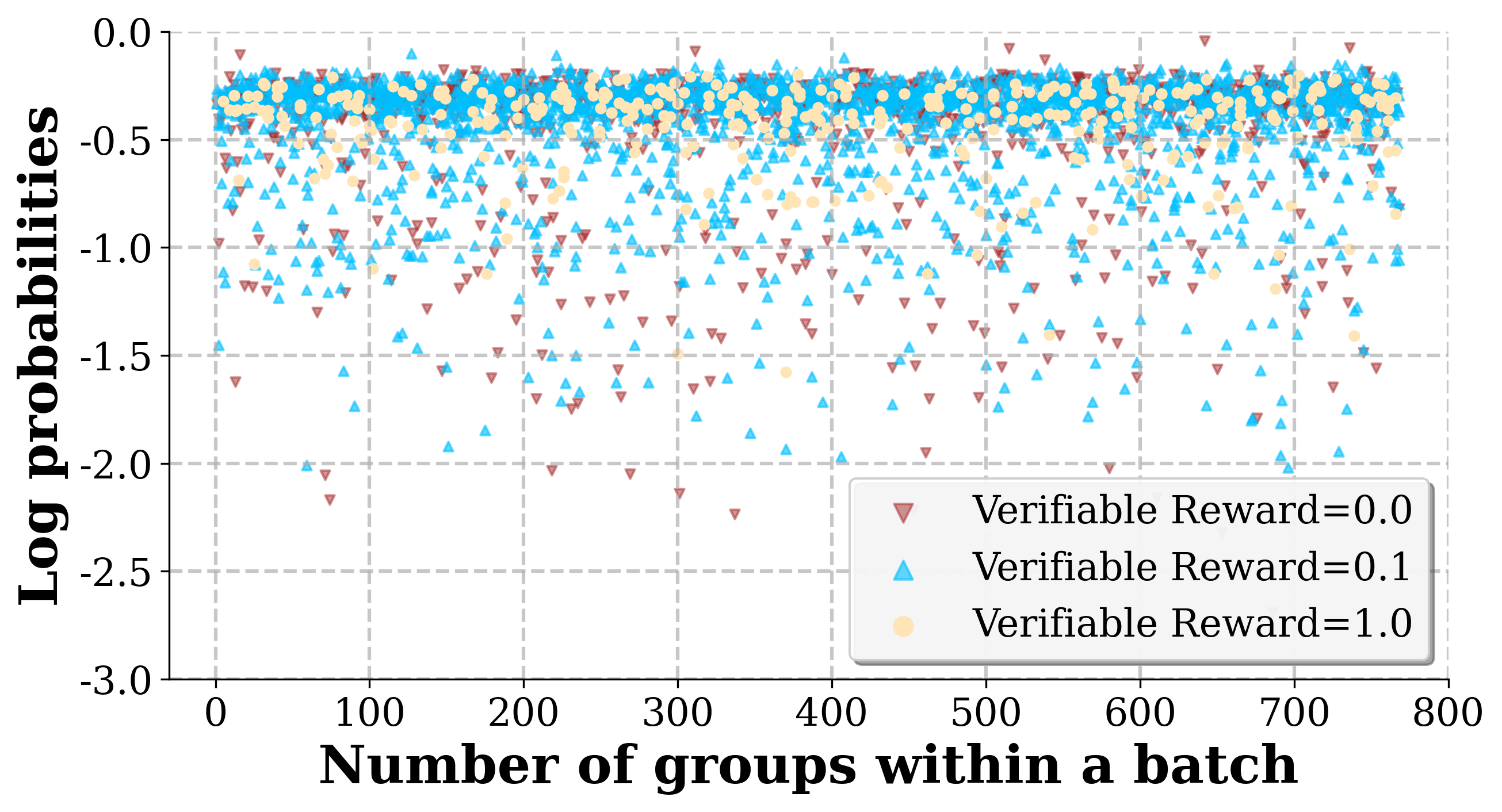}
    \subcaption{20\% of entire training process}
    \label{fig3b}
  \end{minipage}
  \begin{minipage}{0.33\textwidth}
    \includegraphics[width=\textwidth]{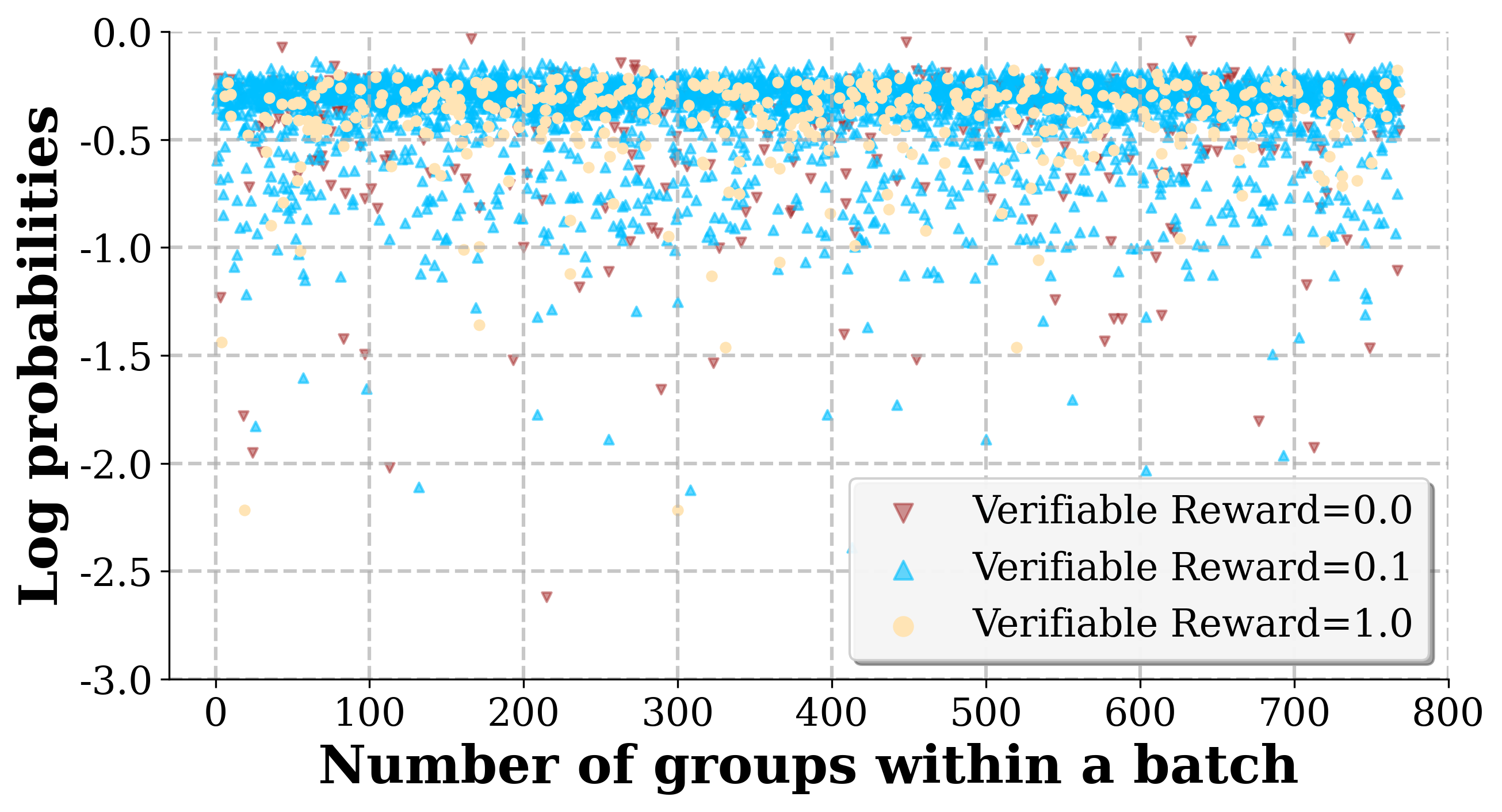}
    \subcaption{40\% of entire training process}
    \label{fig3a}
  \end{minipage}
  \begin{minipage}{0.33\textwidth}
    \includegraphics[width=\textwidth]{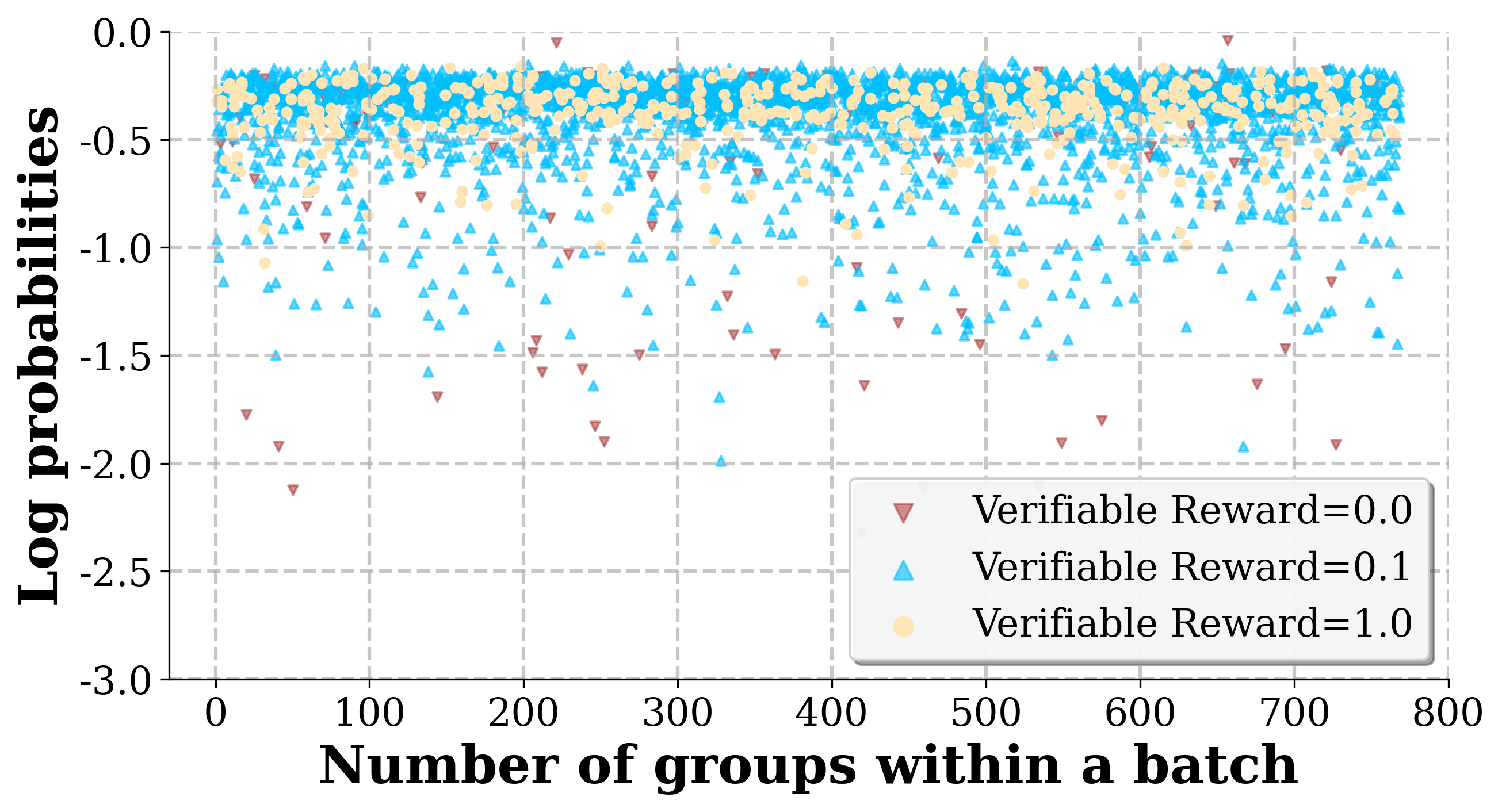}
    \subcaption{60\% of entire training process}
    \label{fig3a}
  \end{minipage}
  \begin{minipage}{0.33\textwidth}
    \includegraphics[width=\textwidth]{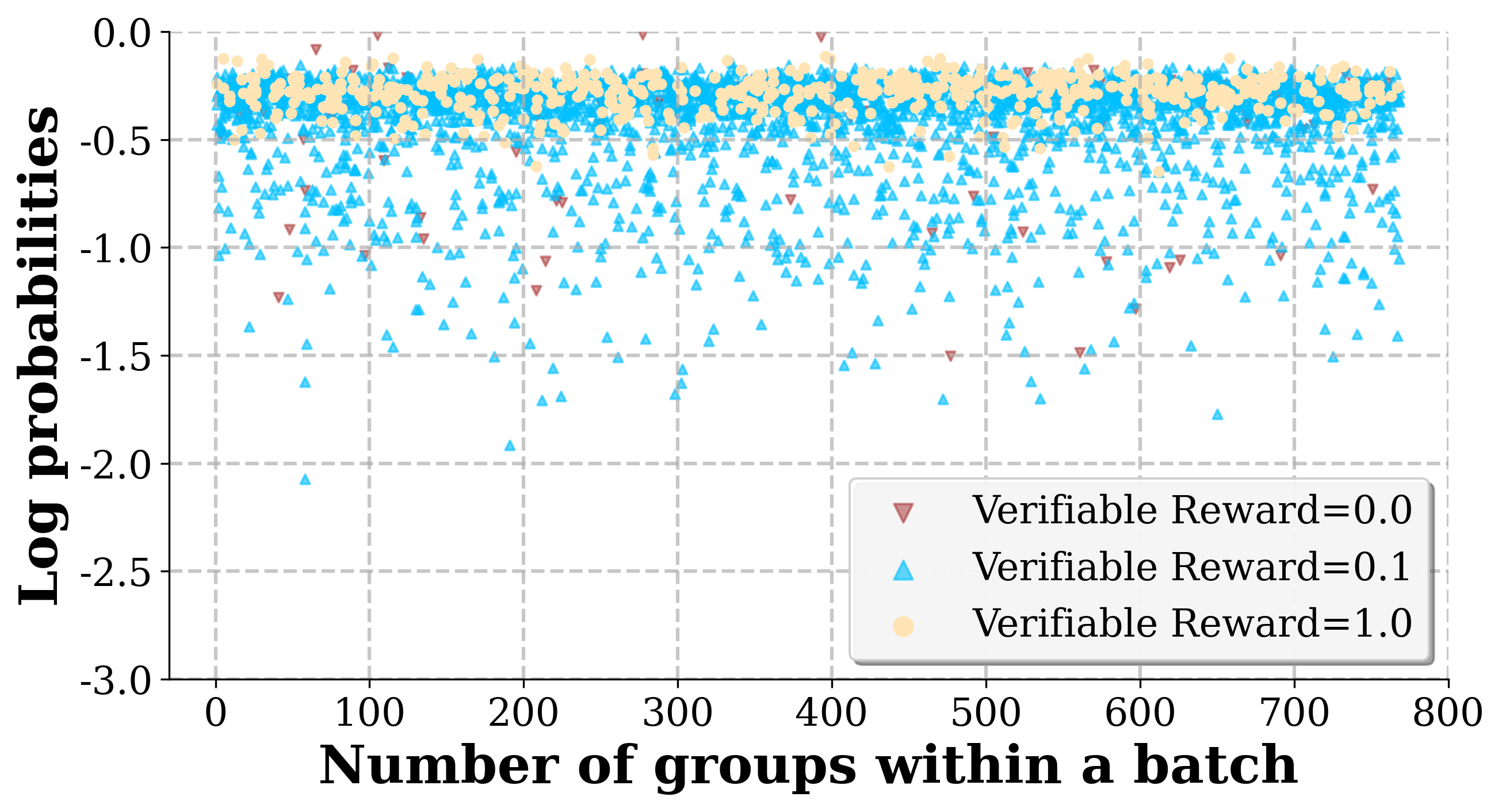}
    \subcaption{80\% of entire training process}
    \label{fig3b}
  \end{minipage}
  \begin{minipage}{0.33\textwidth}
    \includegraphics[width=\textwidth]{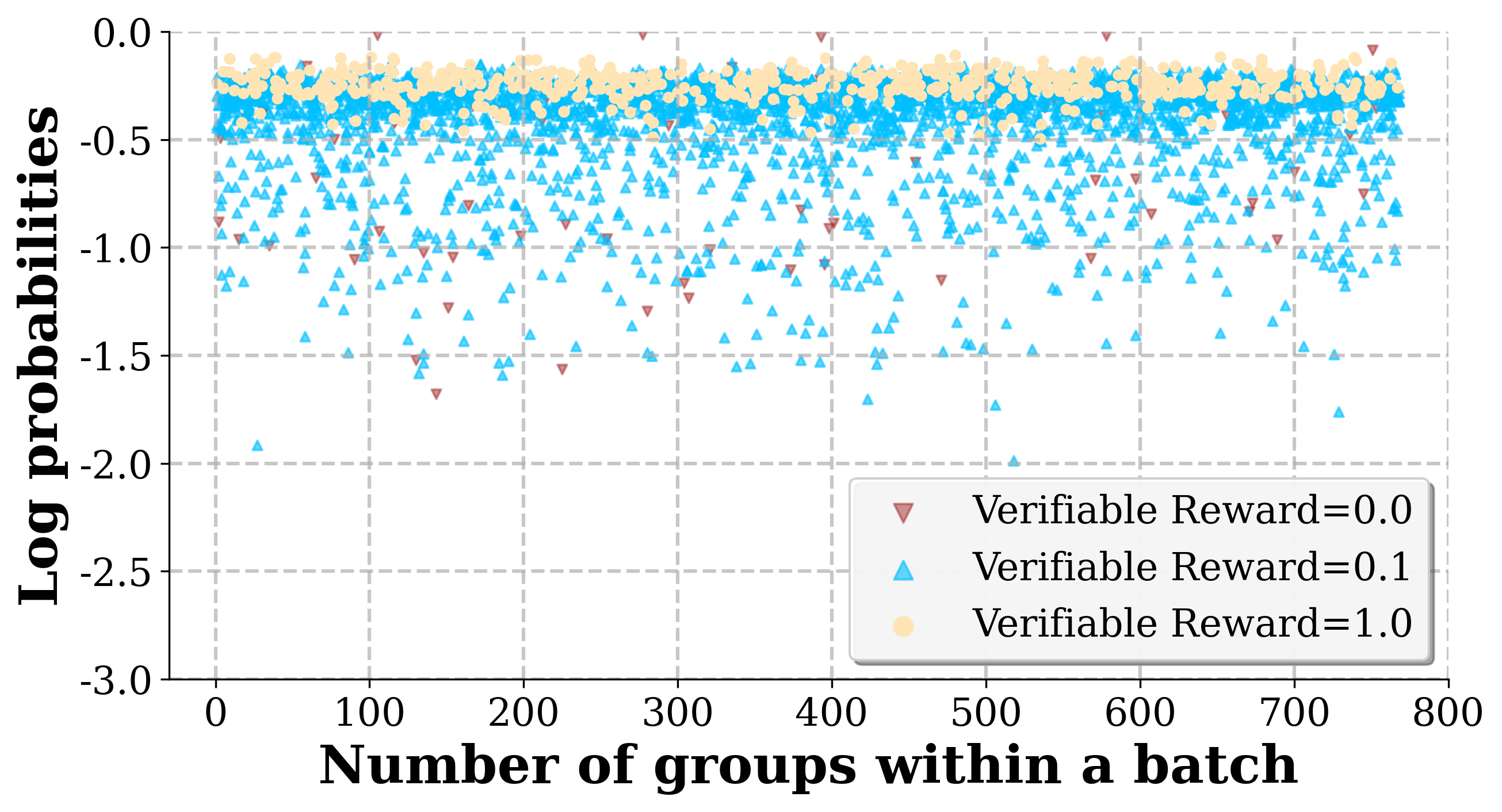}
    \subcaption{100\% of entire training process}
    \label{fig3a}
  \end{minipage}
  \vspace{-0.5em}
  \caption{The distribution of generation probabilities for correct responses and incorrect responses across different training phases throughout the entire training process on the Qwen2.5-7B-Base model.}
  \label{fig9}
  \vspace{-0.5em}
\end{figure*}

These experimental results are precisely consistent with the key insights we proposed in Section~\ref{sec3.4}: (1) In the early stages of training, a high-entropy environment makes it difficult to determine whether low-probability responses are worth learning. As training progresses, the entropy gradually decreases, and correct yet novel low-probability responses begin to emerge, thereby gradually enhancing the motivation to learn these responses. (2) In the later stages of training, when most novel samples have already been mastered, the remaining low-probability responses may merely constitute noise or valueless signals; thus, reducing the influence of intrinsic signals helps stabilize policy learning. (3) The period during which the warm-up phase sharply rises and reaches its peak turning point should align with the period when the entropy gradually decreases to an appropriate level while still retaining active exploration behavior.

For the Qwen2.5-7B-Instruct model, there is little difference in performance between warm-up exploration and directly applying weight decay. The primary reason lies in the fact that this model, having undergone instruction fine-tuning training on top of a pre-trained model, starts with a relatively low entropy level in the initial training phase (absent of a high-entropy environment), thus allowing for selective warm-up exploration. In contrast, for models like Qwen2.5-7B-Base and Qwen3-4B-Base, which are solely pre-trained, the presence of a high-entropy environment in the early training stages makes it more effective to gradually enhance their learning motivation for correct yet low-probability responses.

\begin{figure*}[t]
  \begin{minipage}{0.245\textwidth}
    \includegraphics[width=\textwidth]{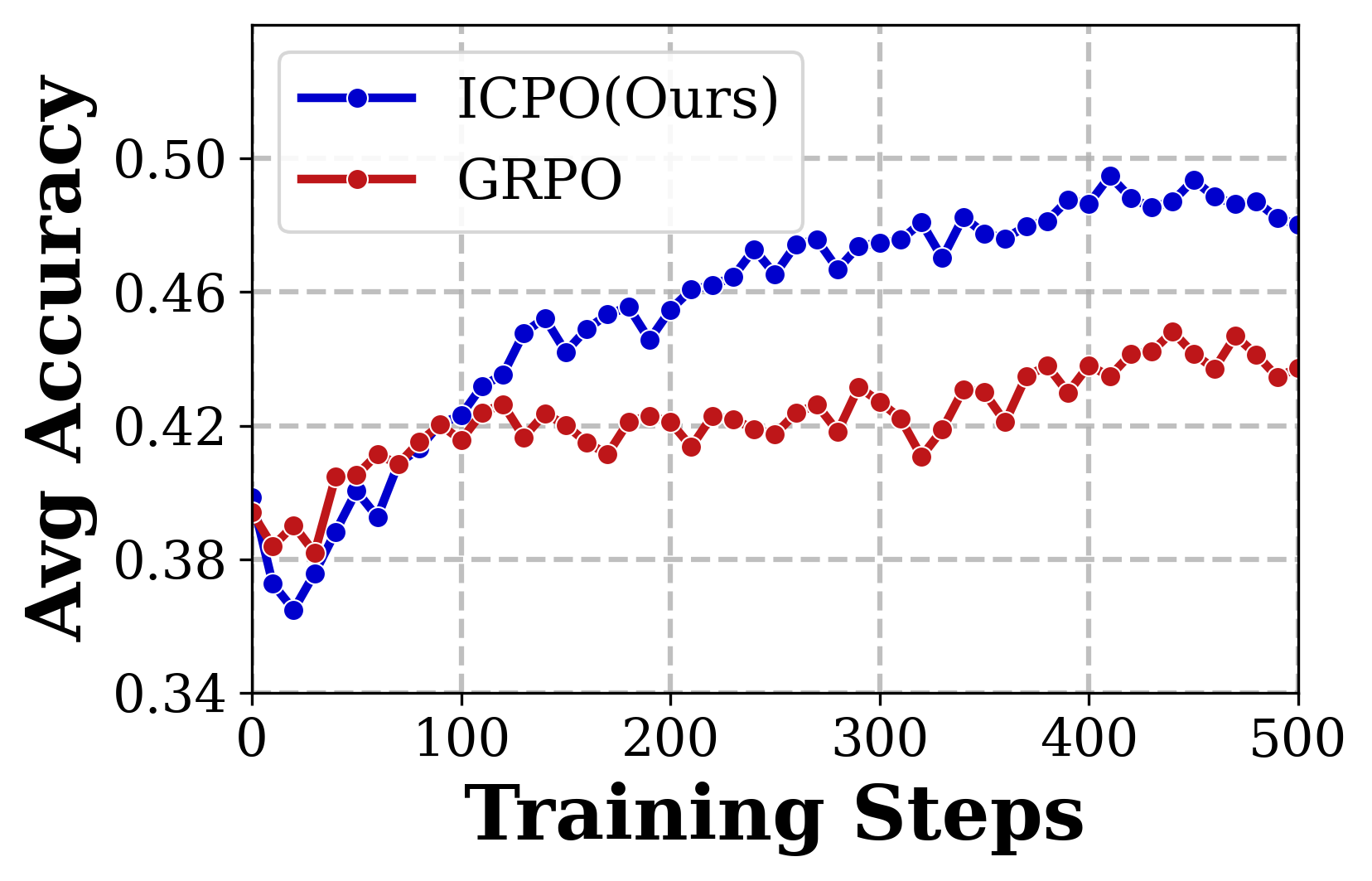}
    \subcaption{0.1 random reward noise}
    \label{fig3a}
  \end{minipage}
  \begin{minipage}{0.245\textwidth}
    \includegraphics[width=\textwidth]{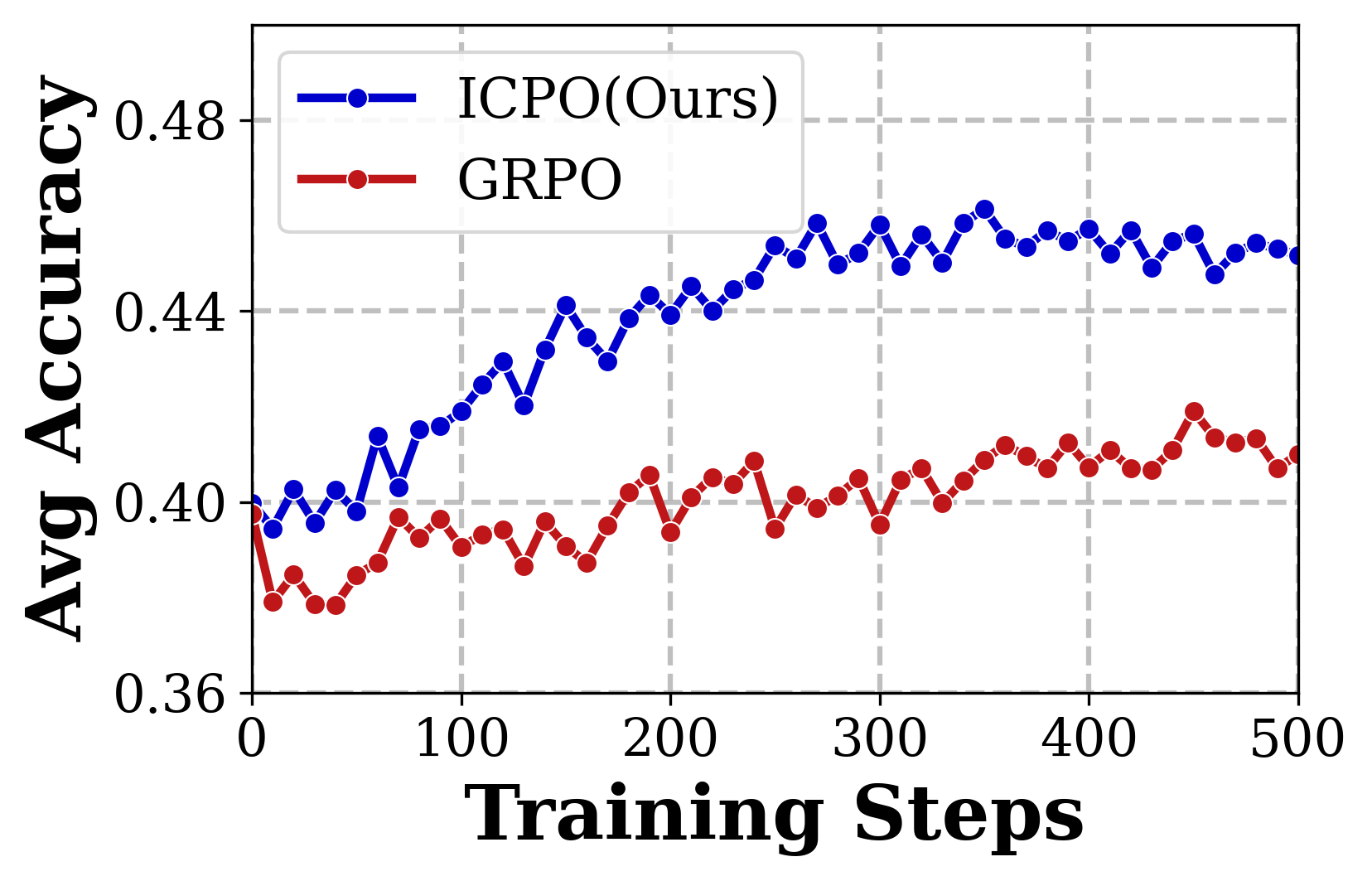}
    \subcaption{0.2 random reward noise}
    \label{fig3b}
  \end{minipage}
  \begin{minipage}{0.245\textwidth}
    \includegraphics[width=\textwidth]{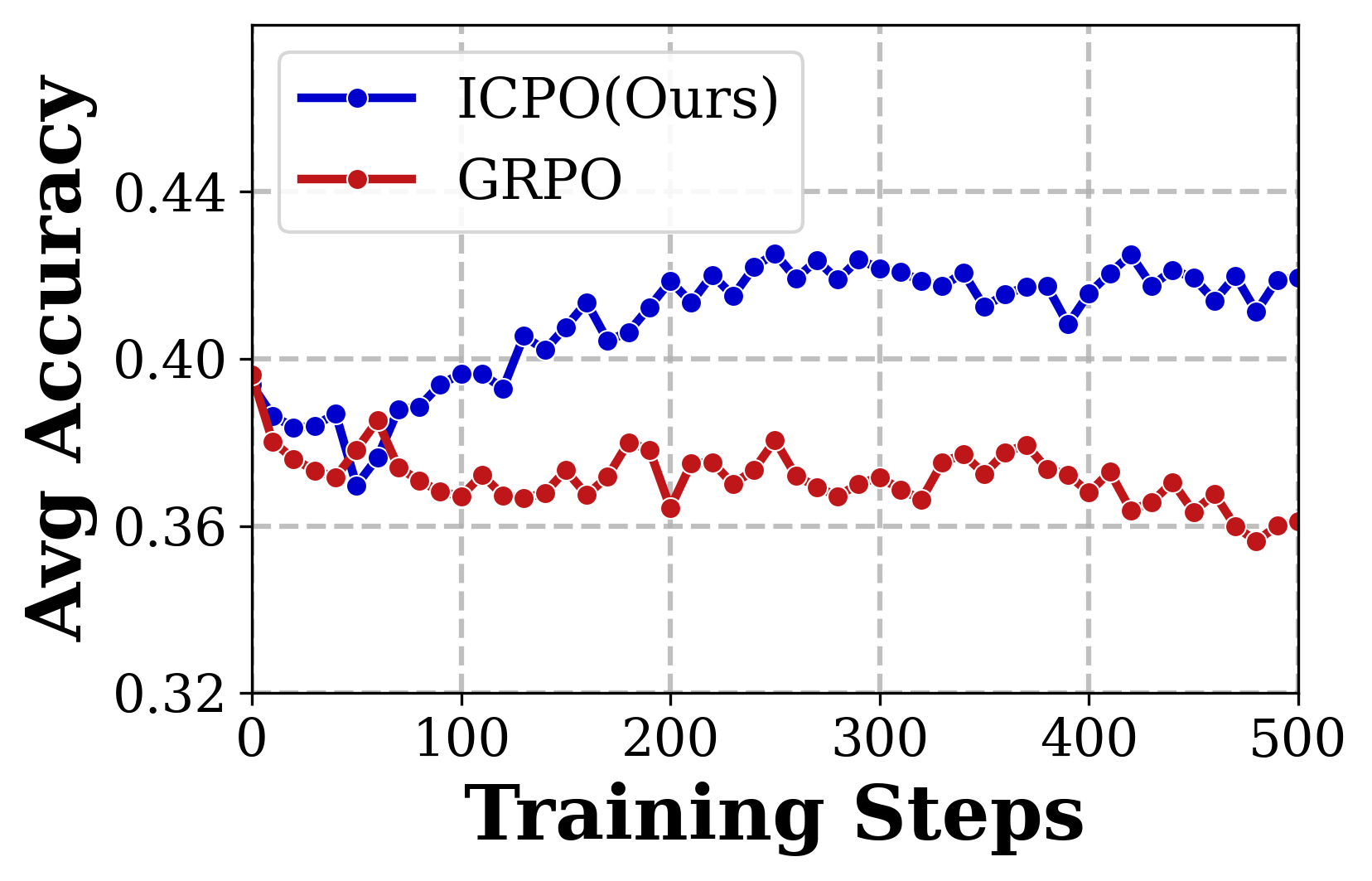}
    \subcaption{0.4 random reward noise}
    \label{fig3a}
  \end{minipage}
    \begin{minipage}{0.245\textwidth}
    \includegraphics[width=\textwidth]{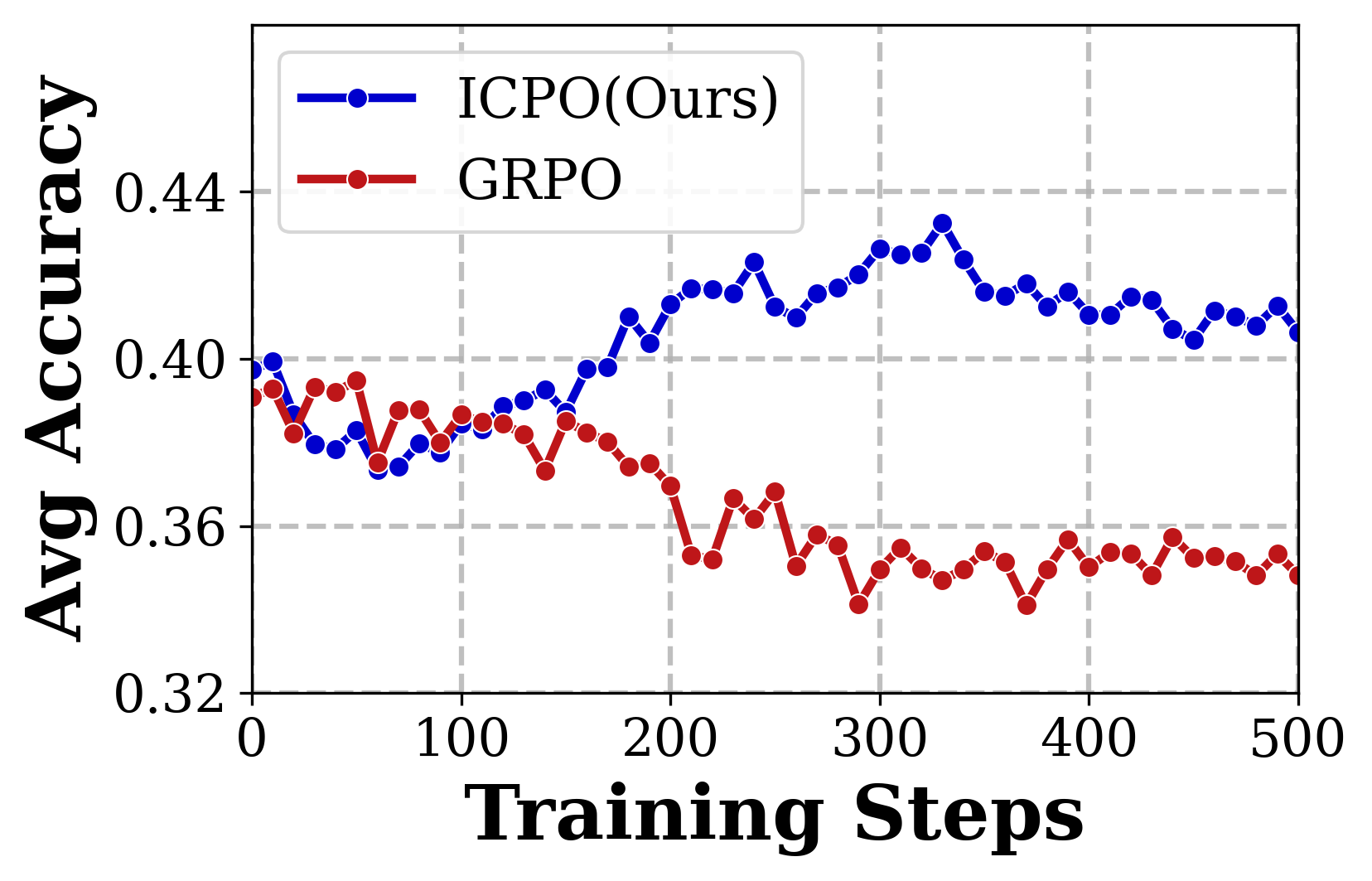}
    \subcaption{0.5 random reward noise}
    \label{fig3a}
  \end{minipage}
  \begin{minipage}{0.245\textwidth}
    \includegraphics[width=\textwidth]{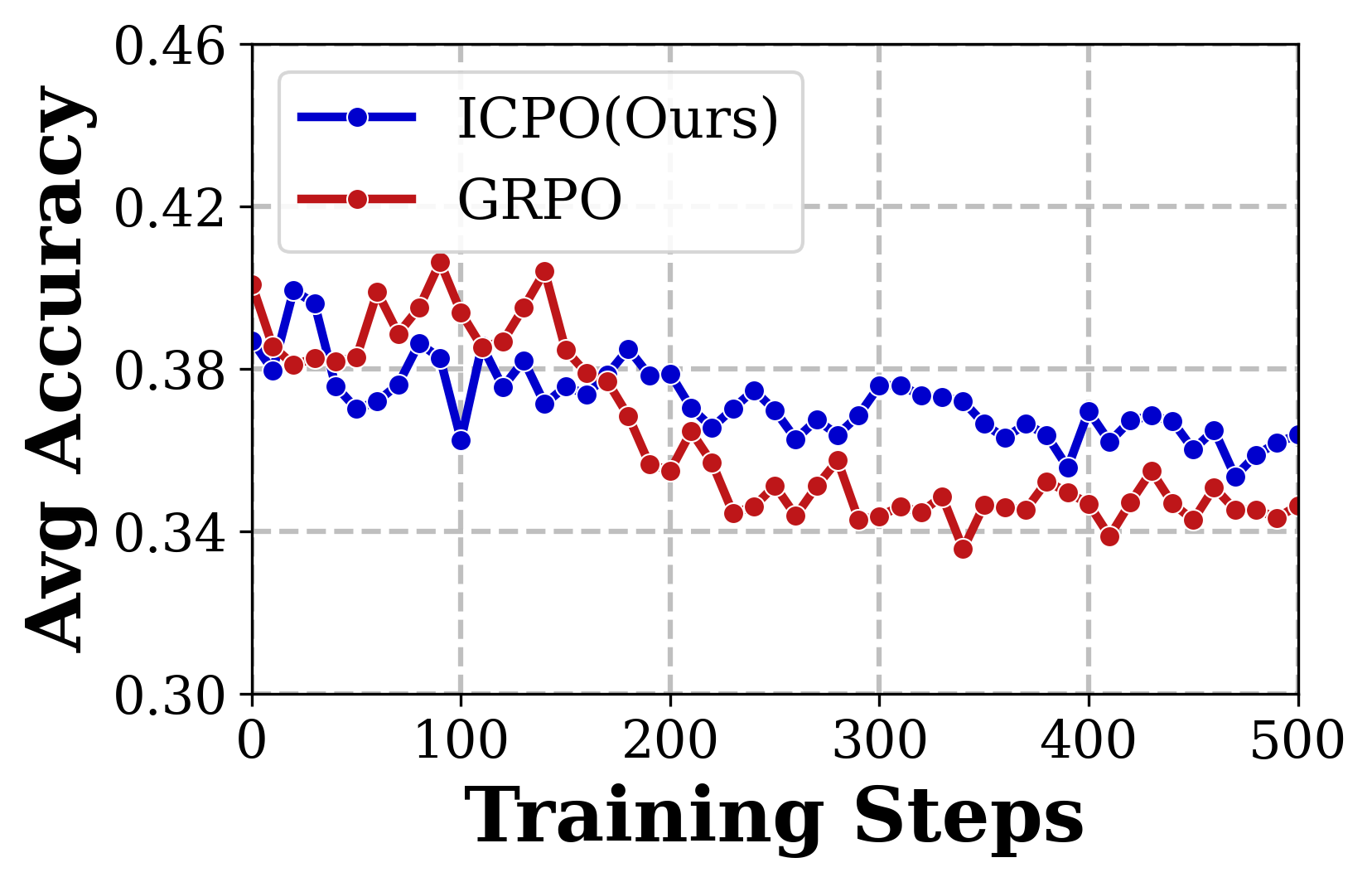}
    \subcaption{0.6 random reward noise}
    \label{fig3a}
  \end{minipage}
  \begin{minipage}{0.245\textwidth}
    \includegraphics[width=\textwidth]{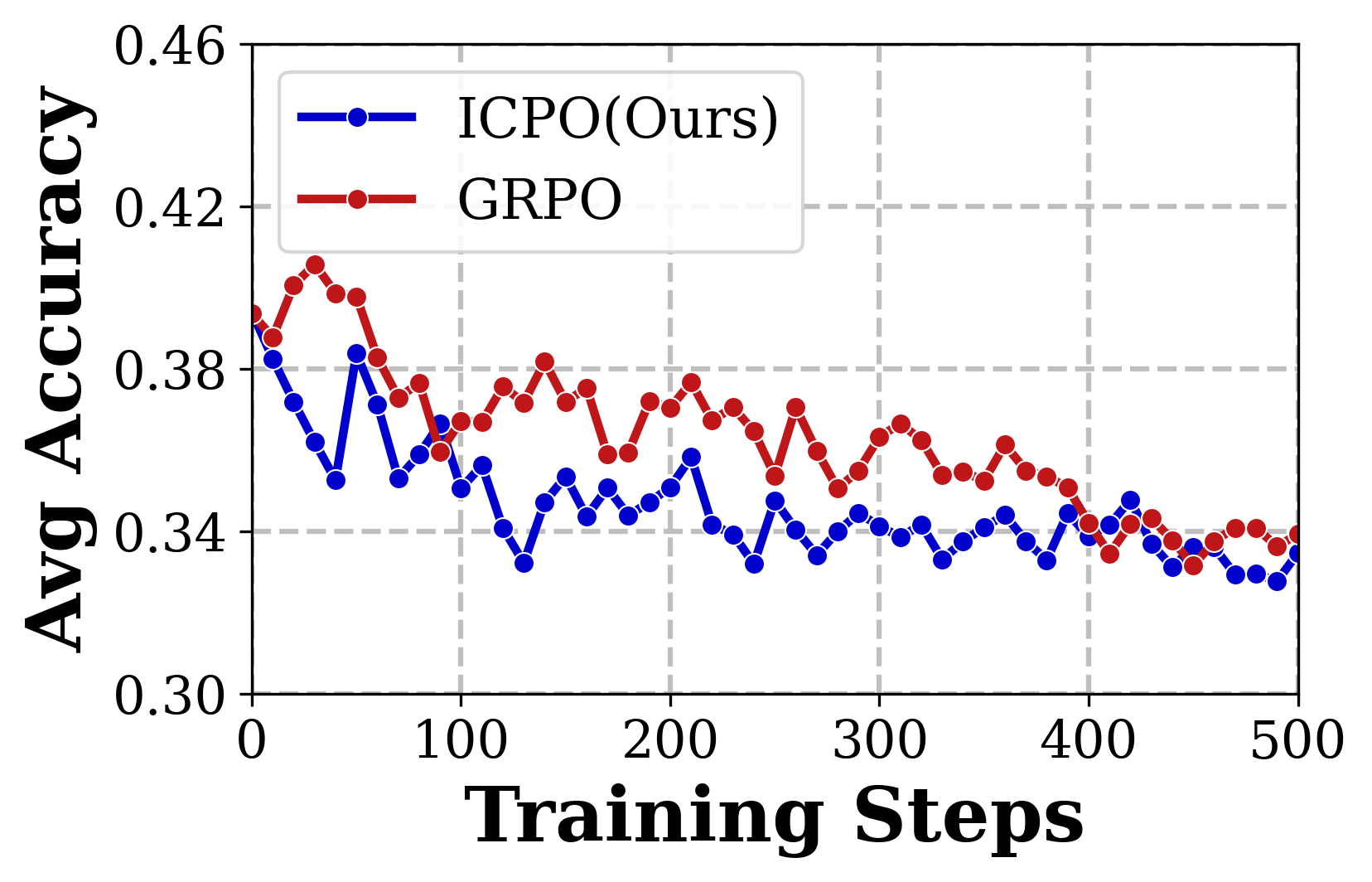}
    \subcaption{0.7 random reward noise}
    \label{fig3b}
  \end{minipage}
  \begin{minipage}{0.245\textwidth}
    \includegraphics[width=\textwidth]{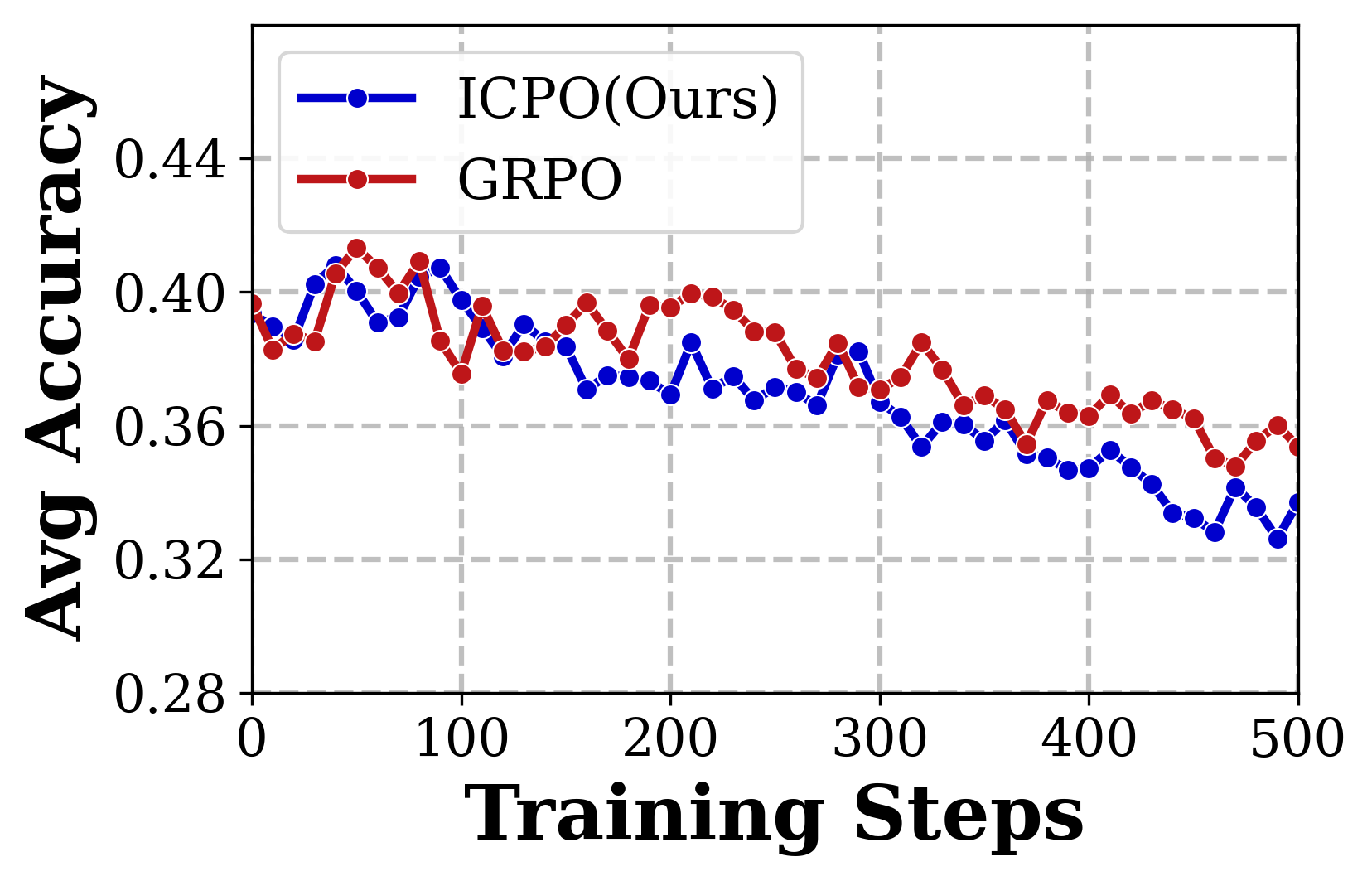}
    \subcaption{0.8 random reward noise}
    \label{fig3a}
  \end{minipage}
    \begin{minipage}{0.245\textwidth}
    \includegraphics[width=\textwidth]{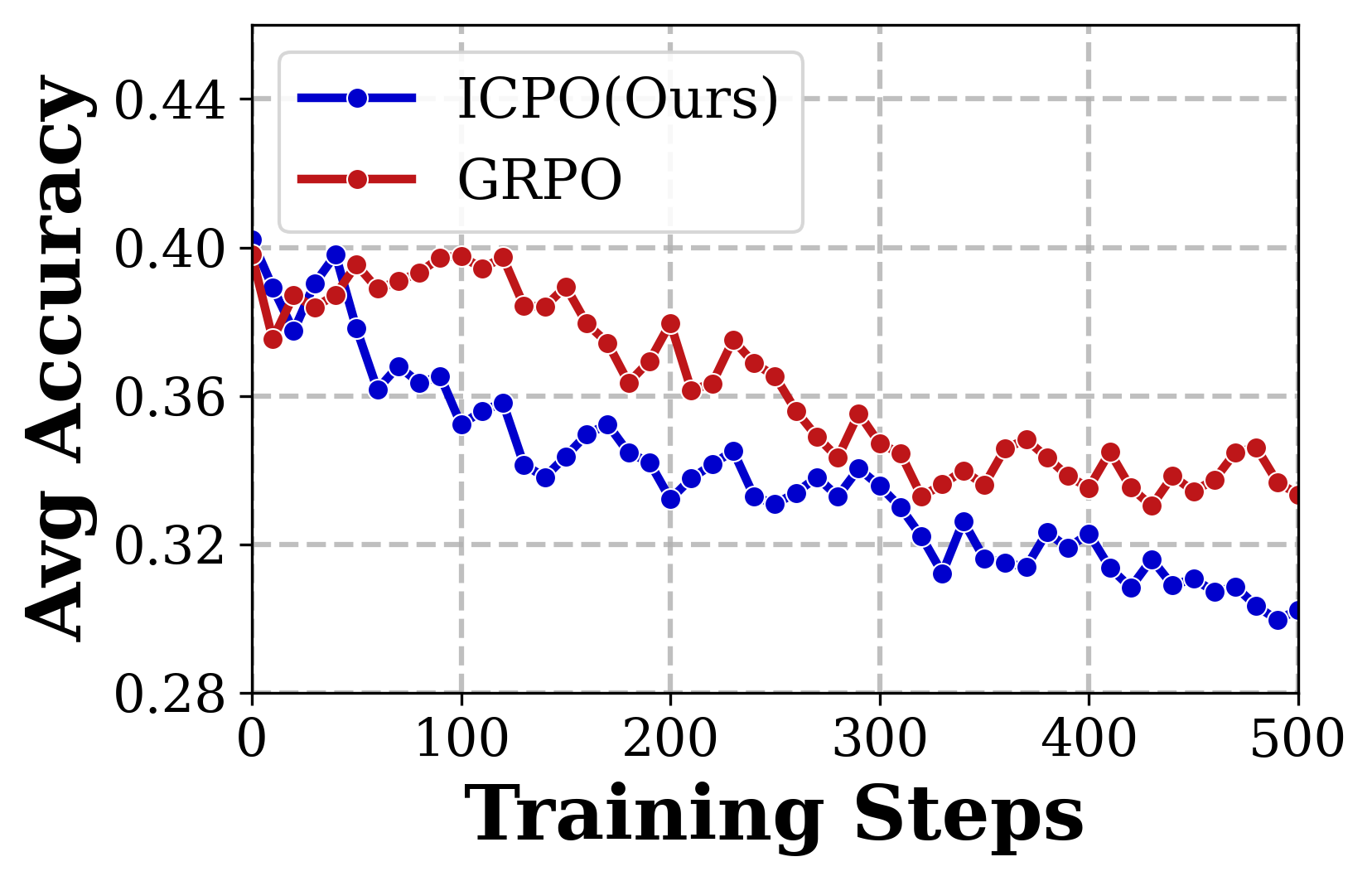}
    \subcaption{0.9 random reward noise}
    \label{fig3a}
  \end{minipage}
  \vspace{-0.5em}
  \caption{Comparison of ICPO and GRPO Performance under Different Intensities of Random Noise Injection during the Training Process Based on the Qwen2.5-7B-Base Model.}
  \label{fig10}
\end{figure*}

\begin{figure*}[t]
  \begin{minipage}{0.245\textwidth}
    \includegraphics[width=\textwidth]{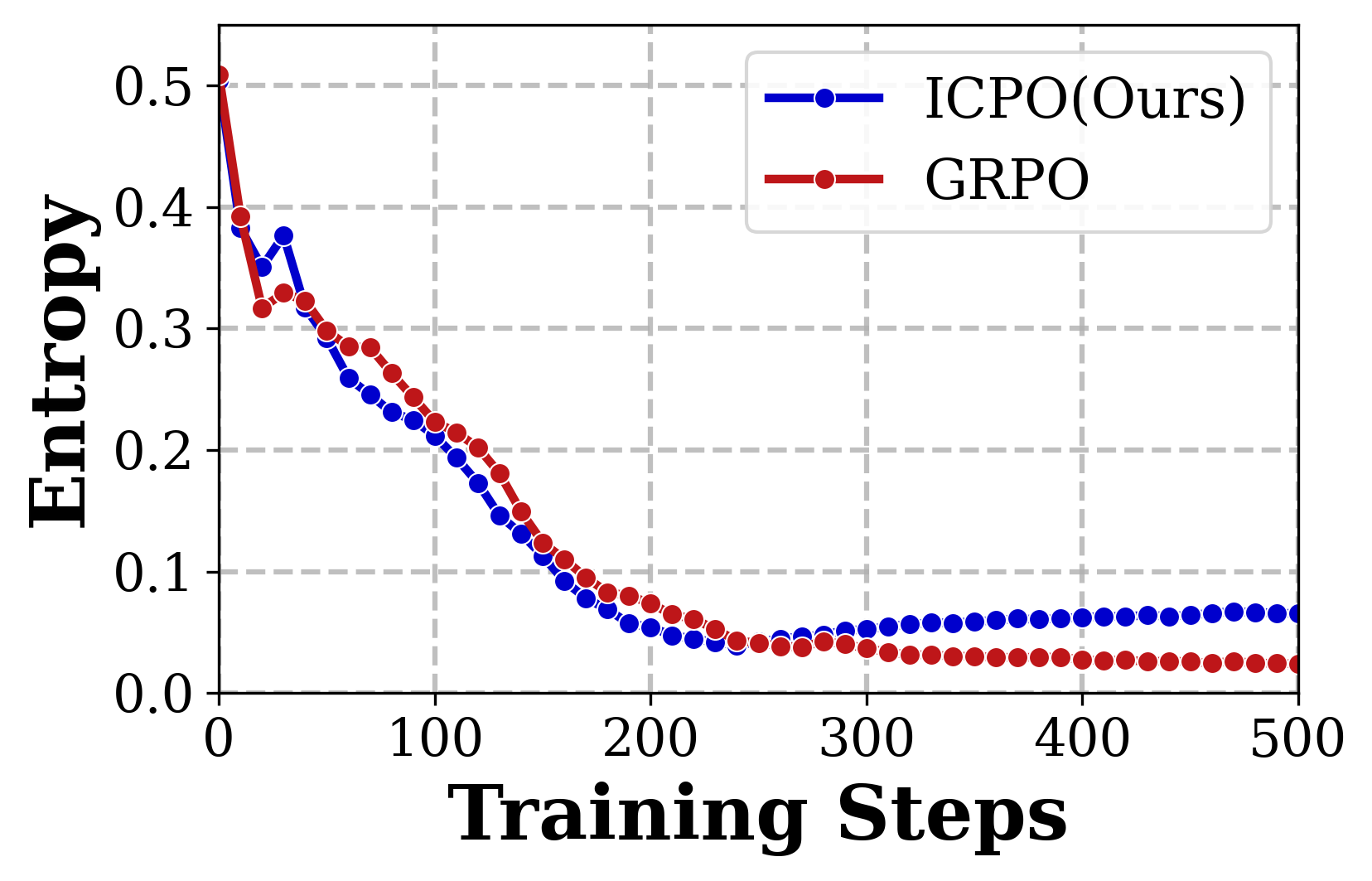}
    \subcaption{0.1 random reward noise}
    \label{fig3a}
  \end{minipage}
  \begin{minipage}{0.245\textwidth}
    \includegraphics[width=\textwidth]{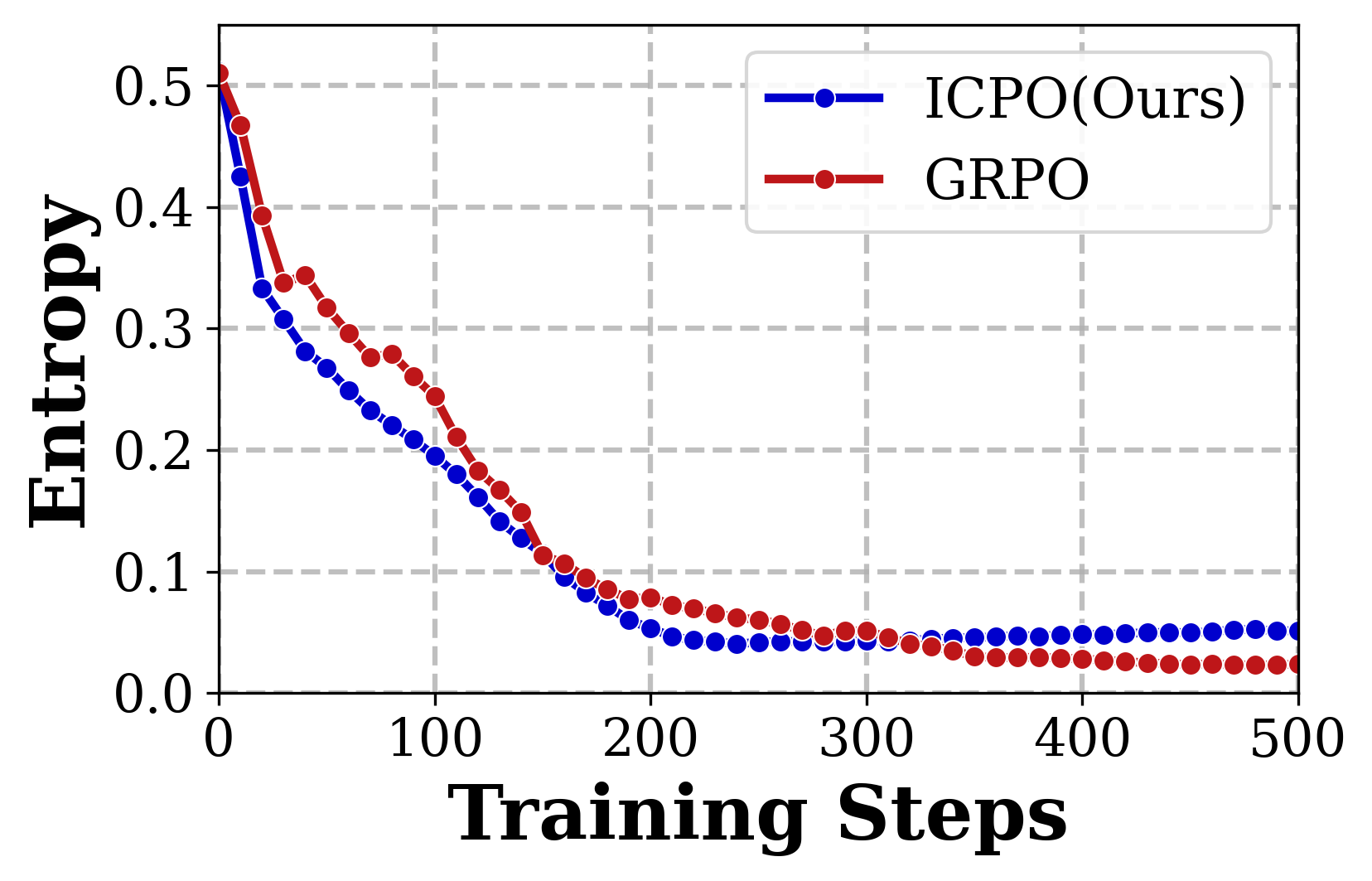}
    \subcaption{0.2 random reward noise}
    \label{fig3b}
  \end{minipage}
  \begin{minipage}{0.245\textwidth}
    \includegraphics[width=\textwidth]{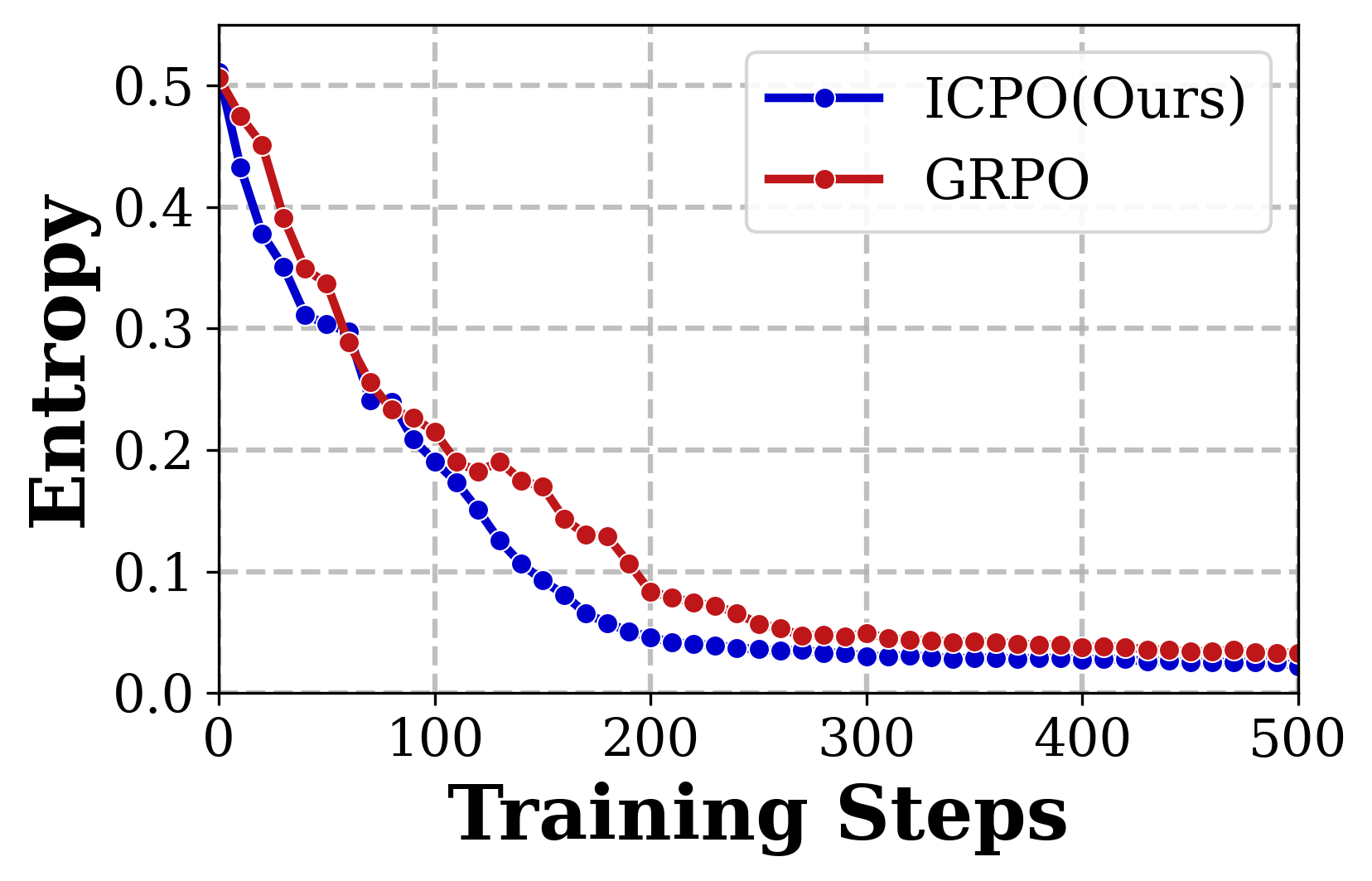}
    \subcaption{0.4 random reward noise}
    \label{fig3a}
  \end{minipage}
    \begin{minipage}{0.245\textwidth}
    \includegraphics[width=\textwidth]{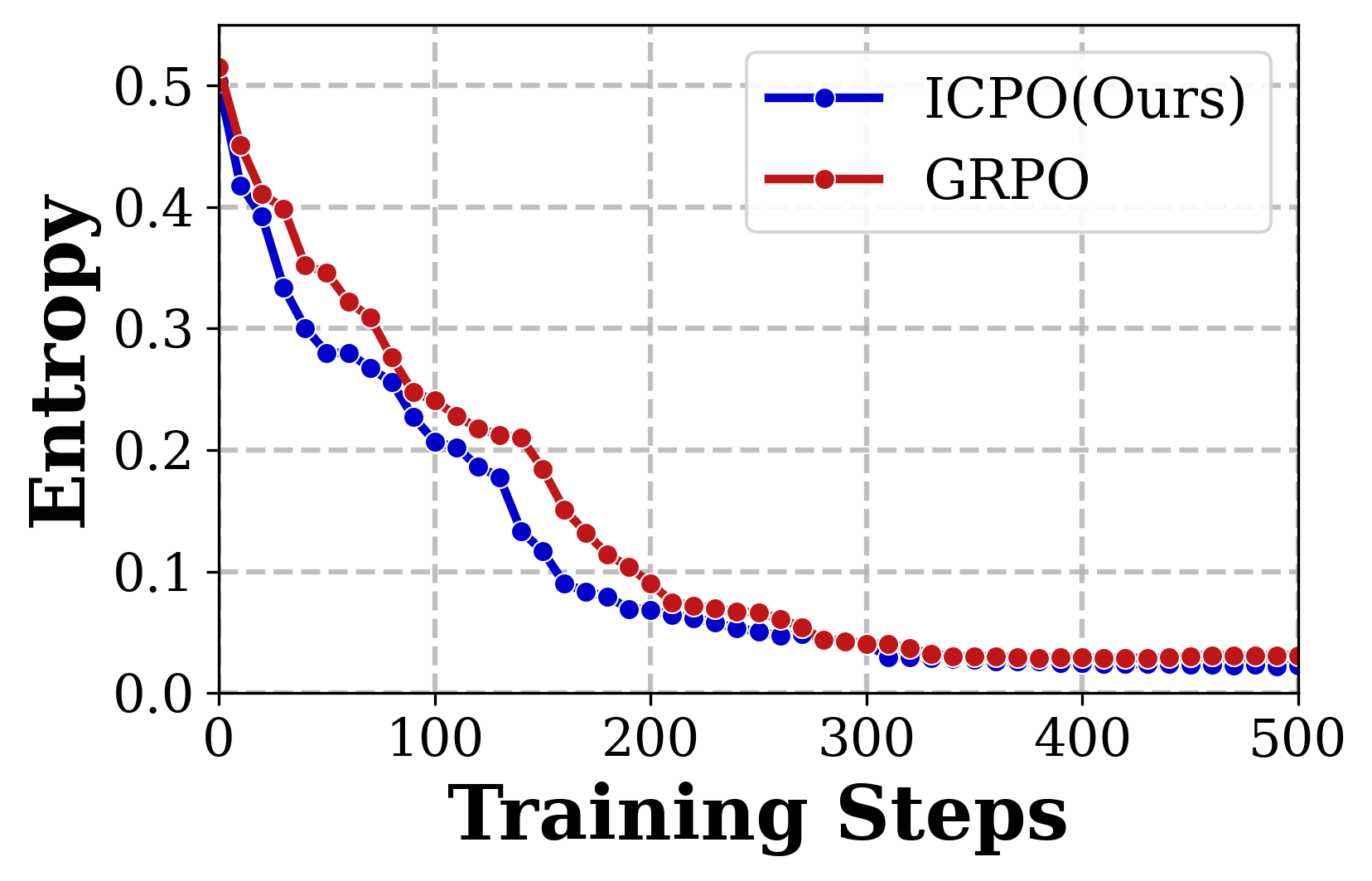}
    \subcaption{0.5 random reward noise}
    \label{fig3a}
  \end{minipage}
  \begin{minipage}{0.245\textwidth}
    \includegraphics[width=\textwidth]{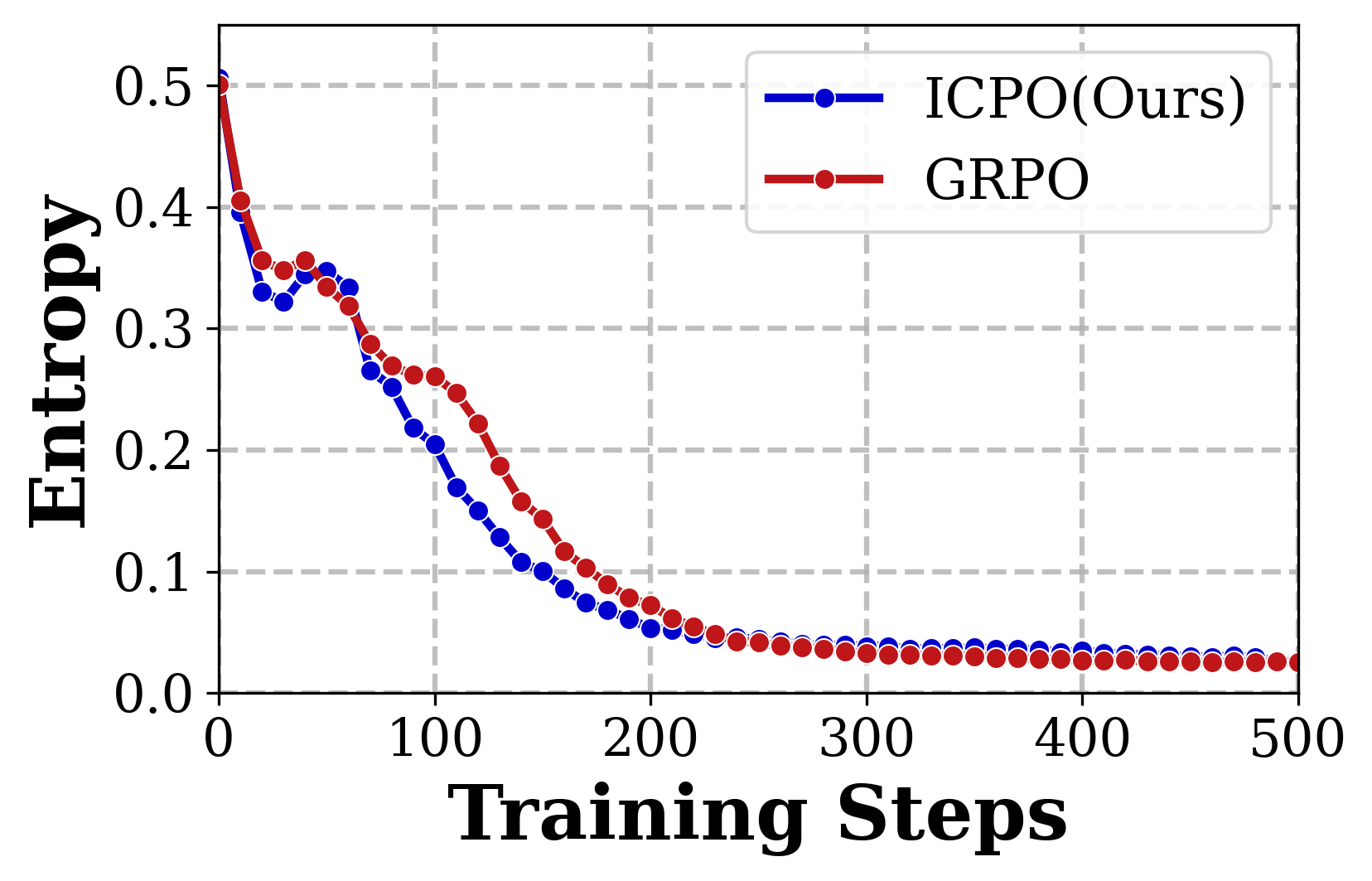}
    \subcaption{0.6 random reward noise}
    \label{fig3a}
  \end{minipage}
  \begin{minipage}{0.245\textwidth}
    \includegraphics[width=\textwidth]{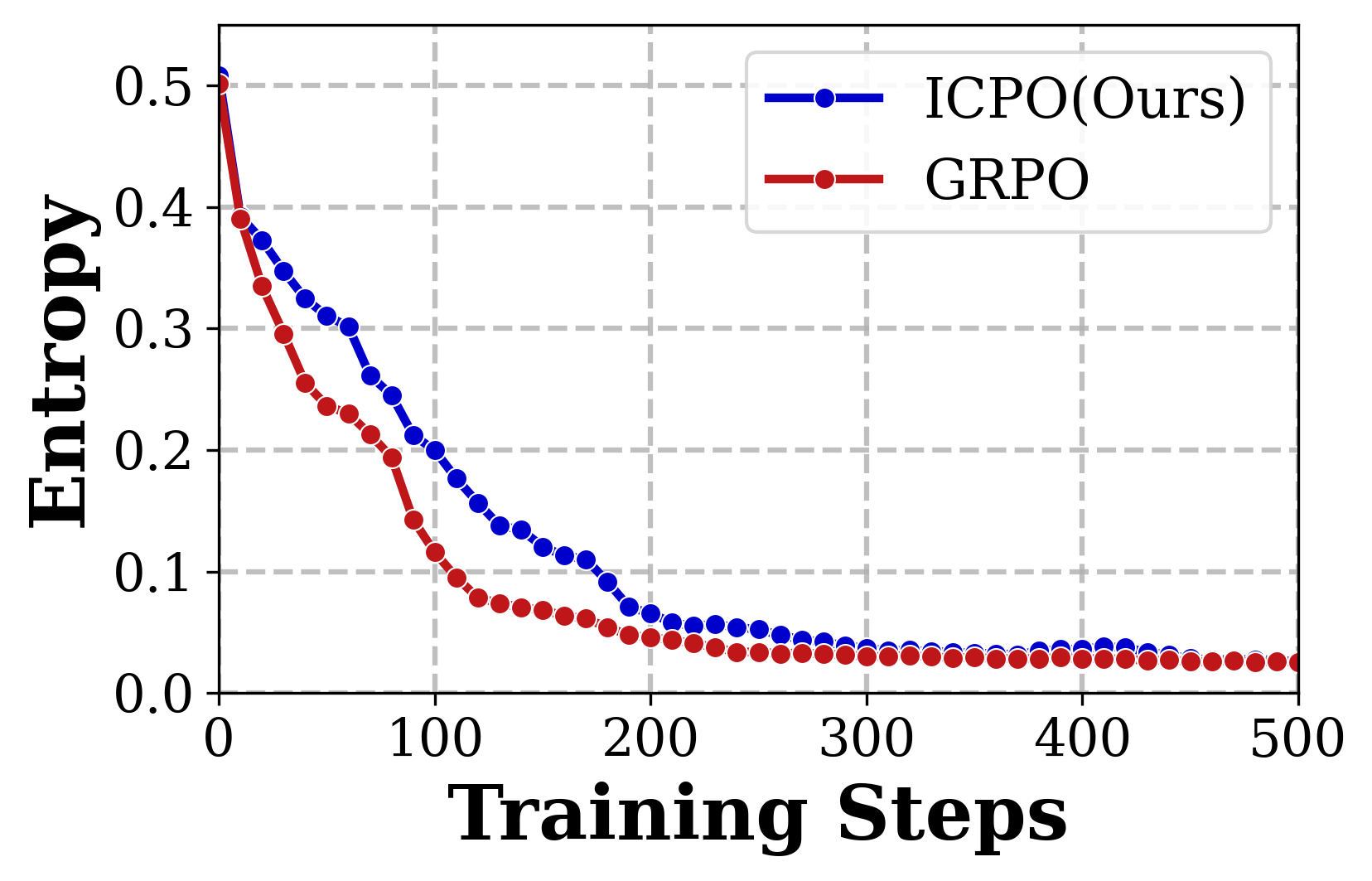}
    \subcaption{0.7 random reward noise}
    \label{fig3b}
  \end{minipage}
  \begin{minipage}{0.245\textwidth}
    \includegraphics[width=\textwidth]{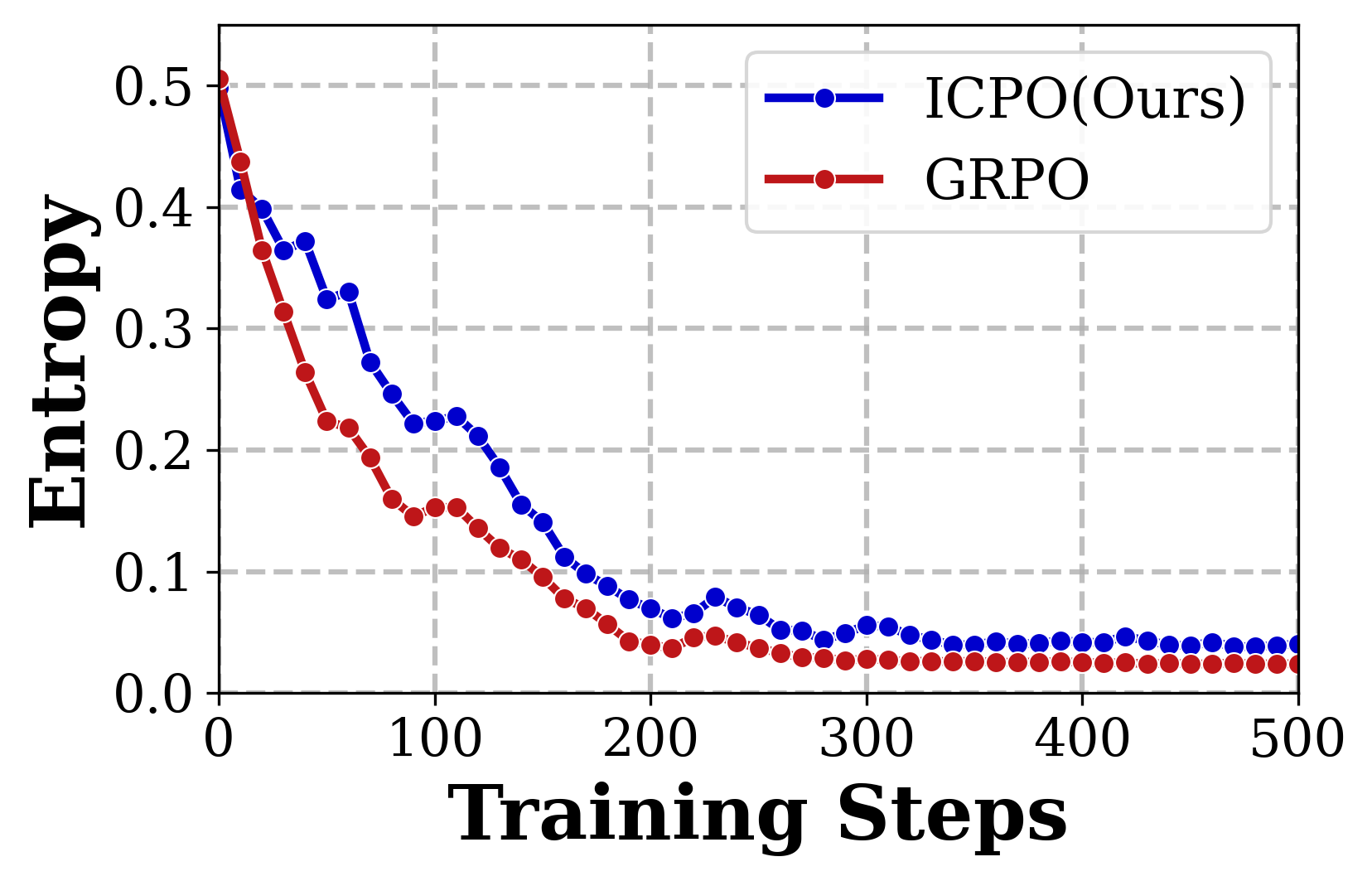}
    \subcaption{0.8 random reward noise}
    \label{fig3a}
  \end{minipage}
    \begin{minipage}{0.245\textwidth}
    \includegraphics[width=\textwidth]{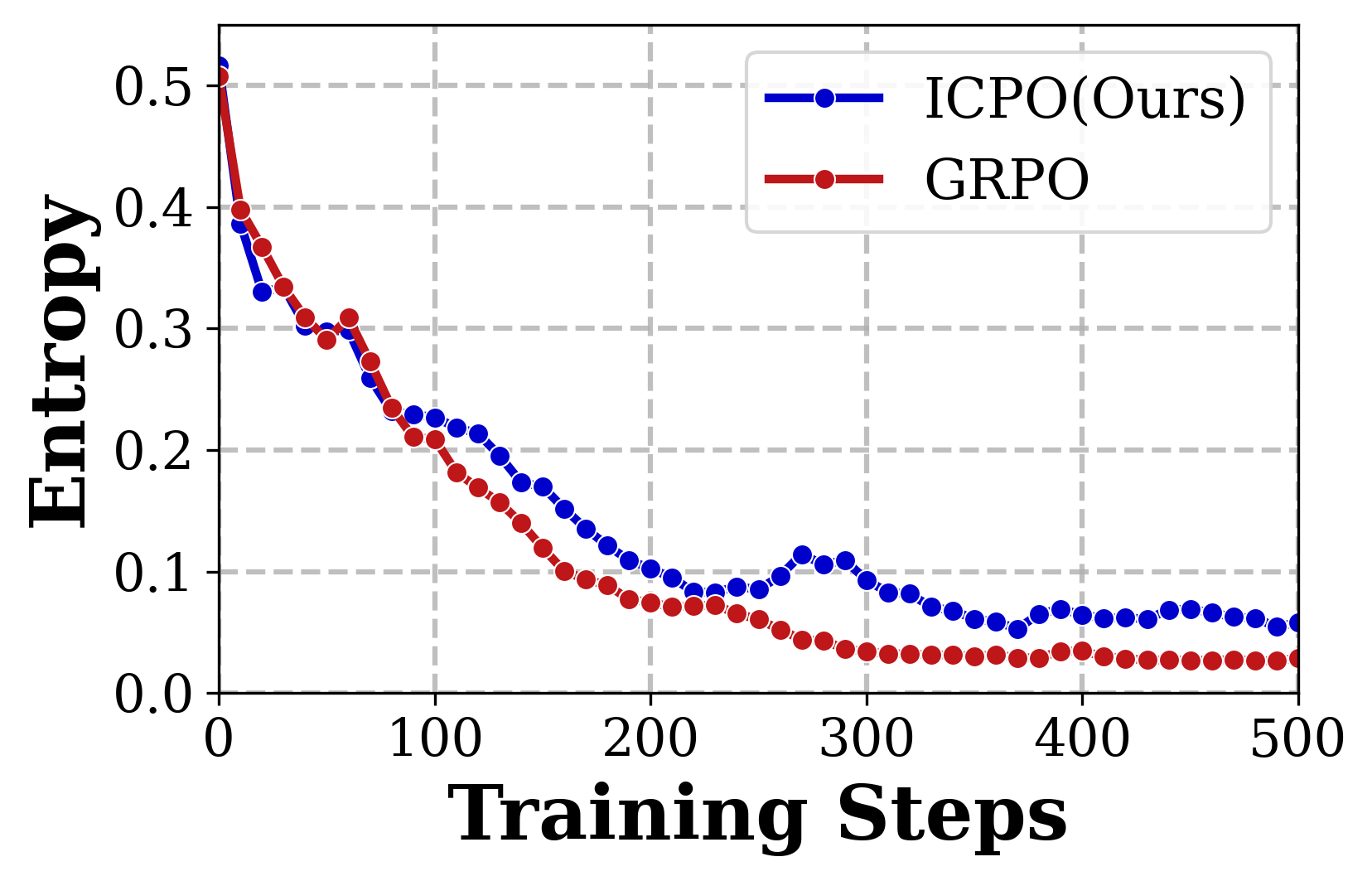}
    \subcaption{0.9 random reward noise}
    \label{fig3a}
  \end{minipage}
  \vspace{-0.5em}
  \caption{Comparison of ICPO and GRPO Entropy under Different Intensities of Random Noise Injection during the Training Process Based on the Qwen2.5-7B-Base Model.}
  \label{fig11}
  \vspace{-0.5em}
\end{figure*}

To validate our key insights, we plotted the probability distributions of generating correct and incorrect responses across different training periods throughout the entire training process, as shown in Figure~\ref{fig9}. It can be observed that: (1) In the very initial stage of training, there are extremely few correct yet low-probability responses, with the majority of low-probability responses being incorrect ones, making it impossible to effectively distinguish novel responses with learning value. (2) As training progresses to the 20\%-40\% stage, correct yet low-probability responses increase significantly, revealing a large number of novel samples with high learning value. (3) In the later stages of training, the number of correct yet low-probability samples returns to a relatively low level, with most low-probability samples now being incorrect ones (data noise or overly complex problems), whose learning value diminishes. This is highly consistent with our proposed viewpoints and also provides an intuitive explanation for why the warm-up decay weight adjustment scheme is effective.

\section{Ablation Study on Noisy Rewards}
To further investigate the performance of ICPO in scenarios with noisy rewards, we varied the extent of random noise injected into the rewards. In the noise reward experiment in Section~\ref{sec4.3}, we randomly selected 40\% of the data during each update iteration and randomly added or subtracted 0.3 from the rewards of this portion of data as noise. Here, we further explored the comparative performance of ICPO and GRPO when the random noise levels were set to 0.1, 0.2, 0.4, 0.5, 0.6, 0.7, 0.8, and 0.9. The training dynamics are illustrated in Figures 10 and 11. It can be observed that when the random noise ranges from 0.1 to 0.5, ICPO can still effectively learn through its intrinsic confidence-driven mechanism, maintaining positive reinforcement for correct yet low-probability responses and effectively suppressing incorrect yet high-probability responses when external rewards are distorted, namely the two scenarios mentioned in Appendix~\ref{appendix4.3}.

However, when the random noise in rewards exceeds 0.6, ICPO's noise resistance reaches a turning point, which aligns precisely with the viewpoint we mentioned in the section on limitations: when rewards completely deviate from the true objectives of the task, this "encouragement-suppression" mechanism undergoes a complete reverse shift. Originally genuine and effective low-probability behaviors are mistakenly labeled as incorrect samples by the erroneous rewards, while those genuinely ineffective or even harmful behaviors are mispackaged by the erroneous rewards as correct yet low-probability high-quality samples, becoming the primary focus of ICPO's optimization efforts and ultimately leading to policy collapse. This also represents a major direction for future optimization of ICPO.

\section{Theoretical Explanation}
\label{sec:appendix6}
We systematically explain the effectiveness of ICPO from three aspects: (1) why ICPO can calibrate policy updates; (2) why ICPO can address the sparse reward problem; and (3) why ICPO can mitigate the interference from noisy rewards.

\subsection{Aspect 1: Calibrate Policy Updates}
As depicted in the "Effectiveness" section of Figure~\ref{fig1}, we have discovered that the intrinsic confidence-driven reward mechanism inherently penalizes confident errors while encouraging novel and correct responses. Specifically, novel correct replies (with relatively lower confidence levels) receive greater positive rewards, whereas incorrect replies stemming from confident reactions (with higher confidence levels) incur greater penalties due to receiving smaller intrinsic rewards. Most notably, for correct responses with extremely low confidence, ICPO does not excessively encourage them, as such responses may lack any learning value due to reward misjudgment or data noise. The following theorem formalizes this intuition.

\begin{figure}[t]
    \includegraphics[width=0.99\linewidth]{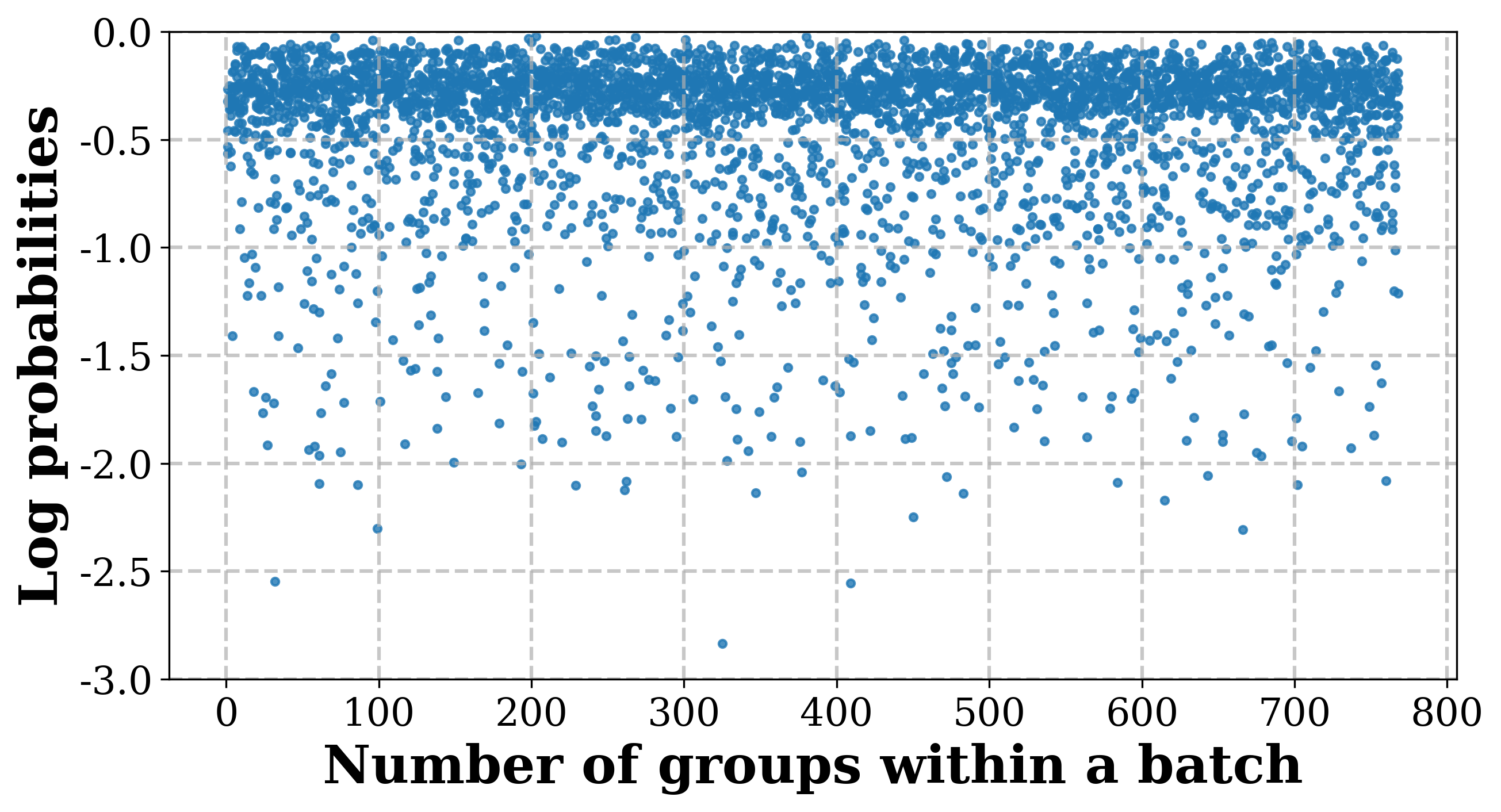}
    \vspace{-0.5em}
    \caption{Scatter plot of responses log-probability distributions within a batch, where each batch contains 768 groups and each group comprises 5 responses.}
  \label{fig12}
  \vspace{-1em}
\end{figure}

\textbf{Theorem 1.} \emph{Let $\pi_{\theta}^t$ denote the policy at training step $t$. Under the influence of the preference advantage score bonus term in Equation~\ref{eq9}, the calibration rules for updating the policy $\pi_{\theta}^{t+1}$ are specified as follows:}

\emph{(i) For correct responses, trajectories with relatively lower confidence (relatively lower generation probabilities) achieve a greater relative advantage enhancement;}

\emph{(ii) For incorrect responses, trajectories with higher confidence (higher generation probabilities) experience a greater relative advantage reduction;}

\emph{(iii) For trajectories with extremely low confidence (extremely low probabilities), their relative advantage enhancement is, conversely, smaller.}

We randomly selected a training batch during the overall training process and statistically analyzed the log-probability distributions of different responses within the group. Furthermore, we employed two sets of real in-group response examples to illustrate the aforementioned theorem. Figure~\ref{fig12} presents the log-probability distributions of responses within a training batch. It can be observed that the log-probability of most responses fluctuate between -2.0 and 0.0, while a small number of extreme responses exhibit log-probability reaching -3.0.

Table~\ref{tab9} illustrates the disparities in advantages of response examples within two real groups under the GRPO and ICPO methods. It can be observed that, compared to GRPO, ICPO tends to provide greater encouragement for correctly generated trajectories with lower confidence levels, thereby facilitating the model's acquisition of novel knowledge. Simultaneously, it reduces the weight assigned to erroneous trajectories with high confidence, thereby curbing the model's overconfidence. More importantly, ICPO's incentive mechanism is relatively moderate, avoiding excessive encouragement for samples that are correct but exhibit extremely low confidence levels.

\begin{table*}[h]
	\centering
    \normalsize
	\renewcommand\arraystretch{1.15}
		\begin{tabular}{m{6.0in}}
			\toprule
            \rowcolor{gray!30} \textbf{Log-probability: $P_1(-0.135)$, $P_2(-0.583)$, $P_3(-0.216)$, $P_4(-0.050)$, $P_5(-0.407)$} \\
            \rowcolor{gray!30} \textbf{Verifiable reward: $R^{verif.}_1(0.100)$, $R^{verif.}_2(0.100)$, $R^{verif.}_3(1.000)$, $R^{verif.}_4(0.100)$, $R^{verif.}_5(1.000)$} \\
			\midrule
            1. Responses within group are ranked as $O_2$, $O_5$, $O_3$, $O_1$, $O_4$ by Equation~\ref{eq6}. \\
            2. Based on Equation~\ref{eq8}, a set of preference pairs is constructed as: $\{(O2, O5), (O2, O3),$ $ (O2, O1), (O2, O4), (O5, O3), (O5, O1), (O5, O4), (O3, O1), (O3, O4), (O1, O4)\}$. \\
            3. According to Equation~\ref{eq9}, preference advantage scores are calculated for each response ($\delta=0.4$): \\
            - $S^p_2=0.4 \cdot (\frac{-0.407}{-0.583} + \frac{-0.216}{-0.583}+\frac{-0.135}{-0.583}+\frac{-0.050}{-0.583})-0.4 \cdot(-0.050)=0.5744$ \\
            - $S^p_5=0.4 \cdot (\frac{-0.216}{-0.407}+\frac{-0.135}{-0.407}+\frac{-0.050}{-0.407})-0.4 \cdot(-0.050)=0.4141$ \\
            - $S^p_3=0.4 \cdot (\frac{-0.135}{-0.216}+\frac{-0.050}{-0.216})-0.4 \cdot(-0.050)=0.3626$ \\
            - $S^p_1=0.4 \cdot (\frac{-0.050}{-0.135})-0.4 \cdot(-0.050)=0.1681$ \\
            - $S^p_4=-0.4 \cdot(-0.050)=0.020$ \\
            4. Reward for each response is calculated by equation~\ref{eq11} ($\tau=2.0, \omega=1.0$):\\
            - $R^{final}_2=0.1000+1.0 \cdot \operatorname{min}(0.5744, \frac{0.1000}{2.0})=0.15$ \\
            - $R^{final}_5=1.0000+1.0 \cdot \operatorname{min}(0.4141, \frac{1.0000}{2.0})=1.4141$ \\
            - $R^{final}_3=1.0000+1.0 \cdot \operatorname{min}(0.3626, \frac{1.0000}{2.0})=1.3626$ \\
            - $R^{final}_1=0.1000+1.0 \cdot \operatorname{min}(0.1681, \frac{0.1000}{2.0})=0.15$ \\
            - $R^{final}_4=0.1000+1.0 \cdot \operatorname{min}(0.020, \frac{0.1000}{2.0})=0.120$ \\
            5. Calculate the relative advantages of different responses in vanilla GRPO and ICPO: \\
            - \textbf{vanilla GRPO: $A_1(-0.816)$, $A_2(-0.816)$, $A_3(1.225)$, $A_4(-0.816)$, $A_5(1.225)$} \\
            - \textbf{ICPO: $A_1(-0.7996)$, $A_2(-0.7996)$, $A_3(1.1819)$, $A_4(-0.8986)$, $A_5(1.2660)$} \\
            \rowcolor{cyan!20} \textbf{Conclusion:} For the correct yet low-confidence response $O_5$, ICPO confers a greater advantage, whereas for the incorrect yet high-confidence response $O_4$, ICPO imposes a more substantial penalty.\\
            \midrule
            \rowcolor{gray!30} \textbf{Log-probability: $P_1(-0.286)$, $P_2(-0.287)$, $P_3(-0.144)$, $P_4(-0.066)$, $P_5(-2.356)$} \\
            \rowcolor{gray!30} \textbf{Verifiable reward: $R^{verif.}_1(0.100)$, $R^{verif.}_2(0.100)$, $R^{verif.}_3(1.000)$, $R^{verif.}_4(0.100)$, $R^{verif.}_5(1.000)$} \\
			\midrule
            Step 1.2 is analogous to the example presented above and will not be elaborated on excessively.\\
            3. According to Equation~\ref{eq9}, preference advantage scores are calculated for each response ($\delta=0.4$): \\
            - $S^p_5=0.4 \cdot (\frac{-0.287}{-2.356} + \frac{-0.286}{-2.356}+\frac{-0.144}{-2.356}+\frac{-0.066}{-2.356})-0.4 \cdot(-0.066)=0.1593$ \\
            - $S^p_2=0.4 \cdot (\frac{-0.286}{-0.287}+\frac{-0.144}{-0.287}+\frac{-0.066}{-0.287})-0.4 \cdot(-0.066)=0.7177$ \\
            - $S^p_1=0.4 \cdot (\frac{-0.144}{-0.286}+\frac{-0.066}{-0.286})-0.4 \cdot(-0.066)=0.3201$ \\
            - $S^p_3=0.4 \cdot (\frac{-0.066}{-0.144})-0.4 \cdot(-0.066)=0.2097$ \\
            - $S^p_4=-0.4 \cdot(-0.066)=0.0264$ \\
            4. Reward for each response is calculated by equation~\ref{eq11} ($\tau=2.0, \omega=1.0$):\\
            - $R^{final}_5=1.0000+1.0 \cdot \operatorname{min}(0.1593, \frac{1.0000}{2.0})=1.1593$ \\
            - $R^{final}_2=0.1000+1.0 \cdot \operatorname{min}(0.7177, \frac{0.1000}{2.0})=0.15$ \\
            - $R^{final}_1=0.1000+1.0 \cdot \operatorname{min}(0.3201, \frac{0.1000}{2.0})=0.15$ \\
            - $R^{final}_3=1.0000+1.0 \cdot \operatorname{min}(0.2097, \frac{1.0000}{2.0})=1.2097$ \\
            - $R^{final}_4=0.1000+1.0 \cdot \operatorname{min}(0.0264, \frac{0.1000}{2.0})=0.1264$ \\
            5. Calculate the relative advantages of different responses in vanilla GRPO and ICPO: \\
            - \textbf{vanilla GRPO: $A_1(-0.816)$, $A_2(-0.816)$, $A_3(1.225)$, $A_4(-0.816)$, $A_5(1.225)$} \\
            - \textbf{ICPO: $A_1(-0.8006)$, $A_2(-0.8006)$, $A_3(1.2733)$, $A_4(-0.8468)$, $A_5(1.1746)$} \\
            \rowcolor{cyan!20} \textbf{Conclusion:} For correct yet extremely low-confidence response $O_5$, ICPO does not over-encourage.\\
			\bottomrule
		\end{tabular}
    \caption{The advantage disparities of response examples in two real groups under GRPO and ICPO.\label{tab9}}
    \vspace{-0.5em}
\end{table*}

\begin{table*}[h]
	\centering
    \normalsize
	\renewcommand\arraystretch{1.15}
		\begin{tabular}{m{6.0in}}
			\toprule
            \rowcolor{gray!30} \textbf{Log-probability: $P_1(-0.135)$, $P_2(-0.583)$, $P_3(-0.216)$, $P_4(-0.050)$, $P_5(-0.407)$} \\
            \rowcolor{gray!30} \textbf{Verifiable reward: $R^{verif.}_1(0.100)$, $R^{verif.}_2(0.100)$, $R^{verif.}_3(1.000)$, $R^{verif.}_4(0.100)$, $R^{verif.}_5(1.000)$} \\
            \rowcolor{gray!30} \textbf{Noisy reward: $R^{Noisy}_1(0.400)$, $R^{Noisy}_2(0.100)$, $R^{Noisy}_3(1.000)$, $R^{Noisy}_4(0.400)$, $R^{Noisy}_5(0.700)$} \\
			\midrule
            1. Responses within group are ranked as $O_2$, $O_5$, $O_3$, $O_1$, $O_4$ by Equation~\ref{eq6}. \\
            2. Based on Equation~\ref{eq8}, a set of preference pairs is constructed as: $\{(O2, O5), (O2, O3),$ $ (O2, O1), (O2, O4), (O5, O3), (O5, O1), (O5, O4), (O3, O1), (O3, O4), (O1, O4)\}$. \\
            3. According to Equation~\ref{eq9}, preference advantage scores are calculated for each response ($\delta=0.4$): \\
            - $S^p_2=0.4 \cdot (\frac{-0.407}{-0.583} + \frac{-0.216}{-0.583}+\frac{-0.135}{-0.583}+\frac{-0.050}{-0.583})-0.4 \cdot(-0.050)=0.5744$ \\
            - $S^p_5=0.4 \cdot (\frac{-0.216}{-0.407}+\frac{-0.135}{-0.407}+\frac{-0.050}{-0.407})-0.4 \cdot(-0.050)=0.4141$ \\
            - $S^p_3=0.4 \cdot (\frac{-0.135}{-0.216}+\frac{-0.050}{-0.216})-0.4 \cdot(-0.050)=0.3626$ \\
            - $S^p_1=0.4 \cdot (\frac{-0.050}{-0.135})-0.4 \cdot(-0.050)=0.1681$ \\
            - $S^p_4=-0.4 \cdot(-0.050)=0.020$ \\
            4. Reward for each response is calculated by equation~\ref{eq11} ($\tau=2.0, \omega=1.0$):\\
            - $R^{final}_2=0.1000+1.0 \cdot \operatorname{min}(0.5744, \frac{0.1000}{2.0})=0.15$ \\
            - $R^{final}_5=0.7000+1.0 \cdot \operatorname{min}(0.4141, \frac{0.7000}{2.0})=1.05$ \\
            - $R^{final}_3=1.0000+1.0 \cdot \operatorname{min}(0.3626, \frac{1.0000}{2.0})=1.3626$ \\
            - $R^{final}_1=0.4000+1.0 \cdot \operatorname{min}(0.1681, \frac{0.4000}{2.0})=0.5681$ \\
            - $R^{final}_4=0.4000+1.0 \cdot \operatorname{min}(0.020, \frac{0.4000}{2.0})=0.420$ \\
            5. Calculate the relative advantages of different responses in vanilla GRPO and ICPO: \\
            - \textbf{vanilla GRPO $w$ $R^{verif.}$: $A_1(-0.816)$, $A_2(-0.816)$, $A_3(1.225)$, $A_4(-0.816)$, $A_5(1.225)$} \\
            - \textbf{vanilla GRPO $w$ $R^{Noisy}$: $A_1(-0.3371)$, $A_2(-1.3485)$, $A_3(1.6856)$, $A_4(-0.3371)$, $A_5(0.3371)$} \\
            - \textbf{ICPO $w$ $R^{Noisy}$: $A_1(-0.3243)$, $A_2(-1.2788)$, $A_3(1.4900)$, $A_4(-0.6624)$, $A_5(0.7759)$} \\
            \rowcolor{cyan!20} \textbf{Conclusion:} For responses that are inherently erroneous yet generated with high probability (such as $O_4$), when reward noise erroneously inflates their verifiable rewards, ICPO can effectively suppress such responses. Conversely, for responses that are inherently correct but generated with low probability (such as $O_5$), when reward noise erroneously reduces their verifiable rewards, ICPO can amplify their relative advantages, thereby providing effective encouragement.\\
			\bottomrule
		\end{tabular}
    \caption{The advantage disparities of response examples under GRPO and ICPO in noisy reward scenarios.\label{tab10}}
    \vspace{-0.5em}
\end{table*}

\subsection{Aspect 2: Tackle sparse Reward}
The sparse reward problem refers to the scenario where, under the same prompt, all generated responses receive identical rewards (typically stemming from the sparse design of the reward function). In such cases, GRPO is unable to compute the relative advantages among responses within the group, resulting in a lack of effective gradient signals for policy optimization. This, in turn, triggers policy collapse and a sharp decline in model performance.

In contrast, ICPO effectively addresses this issue by introducing preference advantage scores based on the response generation probabilities. Even when the external rewards for all responses are identical, the generation probabilities of different responses under the current policy may still vary. Consequently, ICPO can assign distinct preference advantage scores to each response. When these preference advantage scores are fused with externally verifiable rewards, the relative merits among responses within the group become distinguishable, thereby providing fine-grained and meaningful optimization signals for policy updates. This significantly enhances training stability and ultimate performance. The examples presented in Table~\ref{tab9} also effectively substantiate this point: when responses have identical rewards, the advantage values computed by GRPO are entirely uniform, failing to accurately differentiate between the quality of responses. However, ICPO successfully distinguishes the relative merits among responses with the same rewards, offering more fine-grained guidance for policy updates.

\subsection{Aspect 3: Mitigate Noisy Reward} \label{appendix4.3}
Through empirical analysis, we have discovered that the intrinsic confidence-driven reward mechanism can mitigate two types of interference caused by noisy rewards. For responses that are inherently erroneous yet generated with high probability, when reward noise erroneously inflates their verifiable rewards, ICPO can effectively suppress such responses. Conversely, for responses that are inherently correct but generated with low probability, when reward noise erroneously reduces their verifiable rewards, ICPO can enhance their relative advantages, thereby providing effective encouragement. The following theorem formalizes this intuition:

\textbf{Theorem 2.} \emph{Let $\pi_{\theta}^t$ denote the policy at training step $t$. Under the influence of the preference advantage score bonus term in Equation~\ref{eq9}, the update from $\pi_{\theta}^t$ to $\pi_{\theta}^{t+1}$ follows the following specific rule for alleviating noisy rewards:}

\emph{(i) For trajectories that are correct yet have relatively low confidence (i.e., relatively lower generation probability), when noise causes their rewards to be lower, ICPO, compared to GRPO, enhances their relative advantage;}

\emph{(ii) For trajectories that are erroneous yet exhibit higher confidence (i.e., higher generation probability), when noise causes their rewards to be higher, ICPO, compared to GRPO, suppresses their relative advantage.}

Table~\ref{tab10} elucidates the theorem through a concrete example. However, for the other two scenarios—namely, trajectories that are correct yet possess high confidence and those that are erroneous with low confidence—when rewards are subject to noise, neither ICPO nor GRPO can achieve accurate guidance. This also constitutes a key direction for future exploration and a significant challenge for the ICPO method.

\end{document}